\documentclass{article}

\PassOptionsToPackage{numbers, square}{natbib}

\usepackage[final]{neurips_2024}

\usepackage[utf8]{inputenc} 
\usepackage[T1]{fontenc}    
\usepackage{hyperref}       
\usepackage{url}            
\usepackage{booktabs}       
\usepackage{amsfonts}       
\usepackage{nicefrac}       
\usepackage{microtype}      
\usepackage{xcolor}         
\usepackage{graphicx}
\usepackage{subfigure}
\usepackage{subcaption}
\usepackage{wrapfig}
\usepackage{multirow}
\usepackage{amsmath}
\usepackage{amssymb}
\usepackage{mathtools}
\usepackage{amsthm}
\usepackage[textsize=tiny]{todonotes}
\makeatletter
\newcommand\figcaption{\def\@captype{figure}\caption}
\newcommand\tabcaption{\def\@captype{table}\caption}
\makeatother

\newcommand\mathrb[1]{\mathrm{\mathbf{#1}}}
\newcommand\nsqure{$\mathtt{\mathbf{N^2}}$}

\title{Towards Dynamic Message Passing on Graphs}

\author{%
  Junshu Sun$^{1,2}$\hfill
  Chenxue Yang$^3$\hfill
  Xiangyang Ji$^4$\hfill
  Qingming Huang$^{1,2,5}$\hfill
  Shuhui Wang$^{1,5}$\thanks{Corresponding author.}\\
  $^1$Institute of Computing Technology, CAS\hspace{1em}
  $^2$University of Chinese Academy of Sciences\\
  $^3$Agriculture Information Institute, CAAS\hspace{1em}
  $^4$Tsinghua University\hspace{1em}
  $^5$Peng Cheng Laboratory\\
  \texttt{\{sunjunshu21s,wangshuhui\}@ict.ac.cn}\\
  \texttt{yangchenxue@caas.cn
  xyji@tsinghua.edu.cn
  qmhuang@ucas.ac.cn}\\  
}

\begin{document}

\maketitle

\begin{abstract}
Message passing plays a vital role in graph neural networks (GNNs) for effective feature learning. However, the over-reliance on input topology diminishes the efficacy of message passing and restricts the ability of GNNs. Despite efforts to mitigate the reliance, existing study encounters message-passing bottlenecks or high computational expense problems, which invokes the demands for flexible message passing with low complexity. In this paper, we propose a novel dynamic message-passing mechanism for GNNs. It projects graph nodes and learnable pseudo nodes into a common space with measurable spatial relations between them. With nodes moving in the space, their evolving relations facilitate flexible pathway construction for a dynamic message-passing process. Associating pseudo nodes to input graphs with their measured relations, graph nodes can communicate with each other intermediately through pseudo nodes under linear complexity. We further develop a GNN model named \nsqure~based on our dynamic message-passing mechanism. \nsqure~employs a single recurrent layer to recursively generate the displacements of nodes and construct optimal dynamic pathways. Evaluation on eighteen benchmarks demonstrates the superior performance of \nsqure~over popular GNNs. \nsqure~successfully scales to large-scale benchmarks and requires significantly fewer parameters for graph classification with the shared recurrent layer.
\end{abstract}

\section{Introduction}
\begin{wrapfigure}{r}{0.38\linewidth}
\centering
\vskip -0.2in
\includegraphics[width=0.75\linewidth]{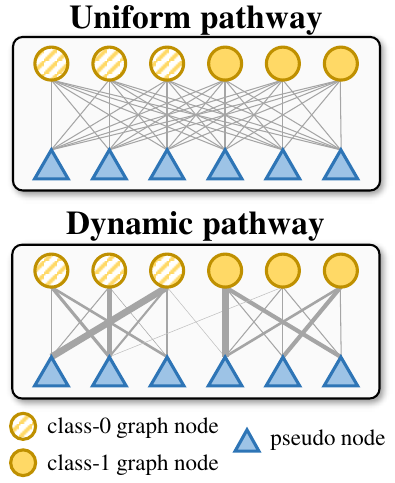}
\caption{\textbf{Comparison between different connection patterns for pseudo nodes and graph nodes.}}
\label{fig:teaser}
\vskip -0.2in
\end{wrapfigure}

The inherent irregular structure of graphs poses nontrivial challenges in graph learning~\cite{zhou_GraphNeuralNetworks_2020}. To enable effective learning on graphs, graph neural networks (GNNs)~\cite{kipf_SemiSupervisedClassificationGraph_2017, velickovic_GraphAttentionNetworks_2018} have been specifically designed for graph-structured data. Within GNN models, message passing serves as a crucial function in extracting informative graph features~\cite{gilmer_NeuralMessagePassing_2017, velickovic_MessagePassingAll_2022}. The vanilla message passing~\cite{kipf_SemiSupervisedClassificationGraph_2017, wu_SimplifyingGraphConvolutional_2019} applies node-centric aggregation constrained between the adjacent nodes. During this aggregation process, central nodes access their neighbors in an isotropic manner~\cite{huang_GoingDeeperPermutationSensitive_2022} and aggregate multi-hop features iteratively. In consequence, the vanilla message passing relies heavily on the input graph structure, leading to over-smoothing~\cite{oono_GraphNeuralNetworks_2021} or over-squashing~\cite{alon_BottleneckGraphNeural_2021, digiovanni_OverSquashingMessagePassing_2023, giovanni_HowDoesOversquashing_2023} issues on GNNs. 

To further improve the effectiveness of GNNs for graph representation learning, one straightforward solution is to decouple message pathways from the input graph structure. Following this direction, methods~\cite{franceschi_LearningDiscreteStructures_2019, deac_ExpanderGraphPropagation_2022} have been proposed to perform message passing beyond input structures for global information exchange. Some methods directly model the pairwise relation between nodes~\cite{ying_TransformersReallyPerform_2021, kreuzer_RethinkingGraphTransformers_2021, mialon_GraphiTEncodingGraph_2021}, but the dense relation causes high computational and space complexity. Other methods incorporate pseudo nodes connected with all the graph nodes to serve as message pathways~\cite{gilmer_NeuralMessagePassing_2017, hwang_RevisitingVirtualNodes_2021, liu_BoostingGraphStructure_2022}.
However, these models employ uniform pathways, \textit{i.e.}, each pseudo node is connected to all the graph nodes with equal weights, as illustrated at the top of Fig.~\ref{fig:teaser}. In consequence, an overwhelming number of node features are squashed equally into pseudo nodes, leading to information bottlenecks on pseudo nodes for message passing and less discriminative representations for downstream tasks~\cite{shirzad_ExphormerSparseTransformers_2023}.

The above limitations call for message passing with flexible pathways and low complexity. In this paper, we propose a novel dynamic message-passing mechanism. To construct flexible pathways, our method measures the specific spatial relations between nodes across time and gives rise to dynamic pathways, as illustrated at the bottom of Fig.~\ref{fig:teaser}. 
To reduce complexity, learnable pseudo nodes are introduced as message-passing proxies between pairs of graph nodes.

Specifically, both graph nodes and pseudo nodes are embedded in a common space with measurable spatial relations between them. By moving nodes in the space, their measured relations evolve accordingly, facilitating a dynamic message-passing process with flexibility. Regarding the measured relation as pseudo edges, the message passing on the input graphs can be extended to pseudo nodes. As a result, graph nodes can communicate with each other intermediately through pseudo nodes, free from dense relation modeling.

To achieve this dynamic process, we further develop a GNN model named \nsqure, based on our graph \textbf{N}odes and pseudo \textbf{N}odes mechanism for message passing. \nsqure~incorporates a recurrent layer to parameterize the displacements of graph nodes and pseudo nodes in the common space. With both types of nodes moving in the common space, \nsqure~measures the actively changing spatial relations and constructs evolving pathways for dynamic message passing.

Our contributions are summarized as follows.
First, we design a flexible and low-complexity message-passing mechanism from a new perspective, where dynamic message pathways are built upon evolving spatial relations between nodes.
Second, we develop a novel GNN model named \nsqure~to achieve dynamic message passing, which employs a recurrent layer to parameterize the evolutionary displacements of nodes.
Third, we demonstrate the advantages of \nsqure~on eighteen real-world benchmarks, where \nsqure~achieves superior performance.
Codes are available at \url{https://github.com/sunjss/N2}.

\section{Related Work}
\textbf{Flexible message passing on graphs.}
In pursuit of expressive operators for graph learning, methods propose to approximate diverse filters with parameterized polynomials~\cite{chien_AdaptiveUniversalGeneralized_2022,defferrard_ConvolutionalNeuralNetworks_2017,klicpera_DiffusionImprovesGraph_2019}. However, due to the prohibitive computational complexity of higher-order polynomials, these methods~\cite{kipf_SemiSupervisedClassificationGraph_2017, xu*_HowPowerfulAre_2019} are constrained with lower orders and only perform local aggregation on graphs. 
This constrained process couples message passing with input topology and contains inherent limitations, including over-smoothing~\cite{li_DeeperInsightsGraph_2018,oono_GraphNeuralNetworks_2021} and over-squashing~\cite{alon_BottleneckGraphNeural_2021,topping_UnderstandingOversquashingBottlenecks_2021}. To overcome these limitations, some works try to decouple message passing from input topology and introduce alternate pathways, including edge shortcuts~\cite{abu-el-haija_MixHopHigherOrderGraph_2019, topping_UnderstandingOversquashingBottlenecks_2021, deac_ExpanderGraphPropagation_2022, gutteridge_DRewDynamicallyRewired_2023}, pseudo nodes~\cite{liu_BoostingGraphStructure_2022, shirzad_ExphormerSparseTransformers_2023}, and graph pooling operations~\cite{gao_GraphUNets_2019, ranjan_ASAPAdaptiveStructure_2020, yuan_StructPoolStructuredGraph_2020}. 

By adding edge shortcuts, methods aim to improve the message-passing efficiency on graphs. These methods can relieve certain bottlenecks on the input graphs~\cite{deac_ExpanderGraphPropagation_2022, qian_ProbabilisticallyRewiredMessagePassing_2023,topping_UnderstandingOversquashingBottlenecks_2021} and aggregate multi-hop information during message passing~\cite{abu-el-haija_MixHopHigherOrderGraph_2019, gutteridge_DRewDynamicallyRewired_2023}. 
Notably, we categorize graph structure learning methods~\cite{zhu_DeepGraphStructure_2021, kazi_DifferentiableGraphModule_2023, wang_PROSEGraphStructure_2023, zhao_GraphGLOWUniversalGeneralizable_2023} into the edge shortcut paradigm, which constructs message pathways by modeling edge connectivity between nodes. 
While edge shortcuts refine local connections, pseudo nodes directly enable global message passing. However, the pseudo nodes in prior works employ uniform pathways to connect with graph nodes, which become bottlenecks in message passing~\cite{shirzad_ExphormerSparseTransformers_2023} and limit efficient global communication. Unlike these works, \nsqure~models dynamic interactions between graph nodes and pseudo nodes, with edge weights varying flexibly across them.

Another line of effort in decoupling message passing from input topology is hierarchical GNNs. These methods~\cite{ranjan_ASAPAdaptiveStructure_2020, yuan_StructPoolStructuredGraph_2020} learn multi-scale graph features through iterative graph pooling, \textit{i.e.}, node clustering or node drop.
Node clustering~\cite{ying_HierarchicalGraphRepresentation_2018, baek_AccurateLearningGraph_2022} learns soft assignment matrices to aggregate nodes into coarser levels. On the other hand, node drop~\cite{gao_GraphUNets_2019, lee_SelfAttentionGraphPooling_2019} ranks and selects salient subsets to prune less critical nodes. Different from graph pooling methods, \nsqure~performs both local and global message passing in each recursive step, avoiding information loss in coarser graphs~\cite{wu_StructuralEntropyGuided_2022}.

\textbf{Reducing complexity for global message passing.}
Self-attention~\cite{vaswani_AttentionAllYou_2017} that models pairwise relations between nodes can be seen as message passing on fully connected graphs. 
However, dense attention requires quadratic space and computational complexity, which is intractable for large-scale graphs. In order to scale attention-based global message passing to larger graphs, recent methods propose to approximate dense attention through expander graphs~\cite{shirzad_ExphormerSparseTransformers_2023}, kernel functions~\cite{wu_NodeFormerScalableGraph_2022}, and diffusion~\cite{wu_DIFFormerScalableGraph_2023}. One similar work~\cite{cai_ConnectionMPNNGraph_2023} to ours proves that message-passing layers with a single pseudo node can approximate dense attention. In this paper, we follow a contrary thread and develop \nsqure~with a single shared message passing layer and multiple pseudo nodes. Each pseudo node interacts dynamically with graph nodes, avoiding becoming message-passing bottlenecks as the uniform connected pattern~\cite{gilmer_NeuralMessagePassing_2017, liu_BoostingGraphStructure_2022}.

\textbf{Recurrent layer for graph learning}.
Scarselli et al.~\cite{scarselli_GraphNeuralNetwork_2009} first employ a recurrent layer to update node features recursively. According to Banach's fixed point theorem \cite{Banach1922}, implementing the recurrent layer as a contraction mapping guarantees the existence of a unique fixed point representation for any input graph, towards which the recursive updates converge. However, their contraction mapping formulation is topology-dependent, incurring over-smoothing as the number of recursive steps increases. In contrast, the recurrent layer in \nsqure~decouples message passing from input topology, empowering flexible communication between graph nodes.

\begin{figure*}[htb]
\centering
\includegraphics[width=\textwidth]{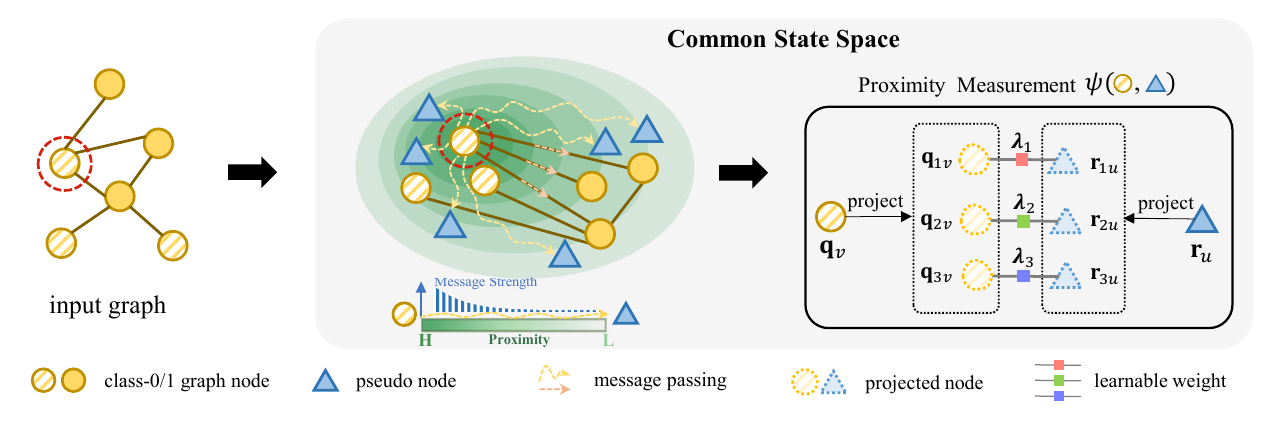}
\caption{\textbf{Dynamic message-passing pathway construction in common state space.} Graph nodes and pseudo nodes interact actively in the common state space, constructing dynamic message pathways through proximity measurement. In empirical model analysis, pseudo nodes tend to be attracted toward a distinct graph node cluster.}
\label{fig:main}
\end{figure*}

\section{Towards Dynamic Message Passing}
Our dynamic message-passing mechanism parameterizes message pathways with measurable relations between nodes in a common space. 
The displacements of nodes in the space give rise to evolving relations. As a result, the message-passing process also changes dynamically.
This section defines the common space tailored for the dynamic message-passing process, where pseudo nodes are employed to reduce the computational complexity. The next section demonstrates how \nsqure~learns the displacements of nodes towards the dynamic process.

\subsection{Preliminaries}\label{ssec:preliminaries}

\textbf{Notations.}~
Let $\mathcal{G}=(\mathcal{V}, \mathcal{E})$ be a graph, where $\mathcal{V}=\{v_1, \cdots, v_n\}$ denotes the node set of size $n$ and $\mathcal{E}=\{e_{v_i,v_j}|v_j \in \mathcal{N}(v_i)\}$ denotes the edge set of size $m$. $\mathcal{N}(\cdot)$ denotes the one-hop neighbor set of a given node. Each node $v\in \mathcal{V}$ corresponds to a feature vector $\mathrb{x}_v \in \mathbb{R}^{d}$ where $d$ is the number of features. Let $\mathrb{X} = (\mathrb{x}_{v_1}, \cdots, \mathrb{x}_{v_n})^\top \in \mathbb{R}^{n\times d}$ be the node feature matrix composed of feature vectors. Let $\mathrb{E}\in \mathbb{R}^{n\times n\times d_e}$ be the edge feature matrix with $d_e$ features. $\mathrb{E}_{i,j,\cdot}=\mathrb{e}_{v_i,v_j}\in\mathbb{R}^{d_e}$.

\textbf{Parameterize pathways as shared functions.}~
When introducing pseudo nodes for message passing, uniform pathways~\cite{gilmer_NeuralMessagePassing_2017, liu_BoostingGraphStructure_2022} directly parameterize the edge weights between pairs of graph nodes and pseudo nodes. Specifying different weights for each pair entails a parameter count scaling to the number of graph nodes.
To ensure a tractable number of parameters for various scales of input graphs, we gain inspiration from DyN~\cite{pei_DynamicsinspiredNeuromorphicVisual_2023}, a distinctive neural network model that directly models neurons, instead of learning connection weights as conventional approaches~\cite{krizhevsky_ImageNetClassificationDeep_2017, vaswani_AttentionAllYou_2017}. The connection weights in DyN are parameterized with a path integral function given the spatial coordinates of neurons. By sharing the integral function across neurons, DyN circumvents the need for heavy parameters.
Drawing an analogy between neurons and nodes, we can also parameterize relations between nodes as a shared function.


\subsection{Common State Space}\label{ssec:state-space}
In light of DyN, we propose to unify graph nodes and pseudo nodes in a common state space $\mathcal{S} \subseteq \mathbb{R}^{q}$ and define their spatial proximity with a shared measurement function. This function measures the relations between pairs of nodes, enabling the construction of dynamic message pathways. 
Here, we borrow the concept of \textit{state} from prior works, such as DyN and LSTM~\cite{hochreiter_LongShortTermMemory_1997}, to denote the learned descriptive embedding of nodes. Examples of the information encoded in the states include node features and local topology. The states of a node identify its spatial coordinate in the state space. For more discussion on "state space", please refer to  Appendix~\ref{sec:state-space}.

\textbf{Embedded node states.}
Given a pseudo node set $\mathcal{U}=\{u_1, \cdots, u_{n_p}\}$ of size $|\mathcal{U}|=n_p$, we embed pseudo nodes in the state space $\mathcal{S}$ as learnable parameters $\mathrb{R}=(\mathrb{r}_{u_1}, \cdots, \mathrb{r}_{u_{n_p}})^\top\in\mathbb{R}^{n_p\times q}$. These learnable elements are named "pseudo nodes" to align with "graph nodes" from inputs. Pseudo nodes can be associated with input graphs through pseudo edges and participate in the message passing between graph nodes. We will elaborate on this association after introducing the spatial proximity measurement in the state space. Given the graph node set $\mathcal{V}$ with feature matrix $\mathrb{X}$, we have graph node states $\mathrb{Q}=(\mathrb{q}_1, \cdots, \mathrb{q}_n)^\top=f(\mathrb{X})\in\mathbb{R}^{n\times q}$, where $f:\mathbb{R}^d\mapsto\mathcal{S}$ is a permutation equivariant function. Displacements of a node in the state space signify shifts in its states, \textit{i.e.,} the learned descriptive embeddings. The objective of a GNN is to model the true distribution of graph nodes in the state space, wherein nodes with similar features obtain proximal embeddings. In the following, we refer to both graph nodes and pseudo nodes collectively as embedded nodes.

\textbf{Proximity measurement.}~
To model complex relations, we assume that the embedded nodes have a non-linear relationship in the state space $\mathcal{S}$. However, modeling non-linearity via operations such as path integral can be computationally complex. To address this problem, we approximate non-linear relations with piece-wise weighted inner products. 
Taking proximity measurement between graph nodes as an example, each node is divided into $k$ pieces with their proximity formulated as
\begin{equation}\label{eq:w-innerproduct}
\psi \left( \mathbf{q}_v,\mathbf{q}_{v'} \right)
= \sum_{i=1}^{k}\lambda_i \mathbf{q}_{iv}^\top \mathbf{q}_{iv'}^{}, \quad
\mathbf{q}_{i\cdot} = \mathtt{NL}_i(\mathbf{q}_{\cdot}),
\end{equation}
where $\lambda_i$ is a learnable parameter, $\mathtt{NL}(\cdot)$ denotes a non-linear function with linear mapping followed by $\mathtt{LeakyReLU}(\cdot)$. $\psi(\cdot, \cdot)$ is termed as proximity instead of distance because it ranges in $\mathbb{R}$. The spatial proximity is weighted with different inner product similarity between $k$ pairs of pieces to approximate complex relations. We conduct ablation studies on the number of pieces $k$ in Appendix~\ref{ssec:nq}.

\subsection{Associating Pseudo Nodes to Input Graphs}\label{ssec:neuron-as-node}
\textbf{Pseudo edges.}
Based on the measurable proximity between the embedded nodes, we now introduce how to obtain pseudo edges and thus associate pseudo nodes to input graphs. For each pseudo node $u\in\mathcal{U}$ embedded at $\mathrb{r}_u$ in the state space, it has pseudo edges with any pseudo node $u'\in\mathcal{U}$ and graph node $v\in\mathcal{V}$, where pseudo edge weights $e_{u, u'}$ and $e_{u, v}$ can be formulated as the proximity
\begin{equation}
    e_{u, u'} = \psi(\mathrb{r}_u, \mathrb{r}_{u'}),\quad
    e_{u, v} = \psi(\mathrb{r}_u, \mathrb{q}_v).
\end{equation}

\textbf{Messages for embedded nodes.}
Following the common practice in GNNs~\cite{velickovic_MessagePassingAll_2022}, we interpret the interaction between the embedded nodes as message passing. To achieve this, both graph node $v$ and pseudo node $u$ learn their messages $\mathrb{m}_v, \mathrb{m}_u\in \mathbb{R}^d$ to be passed in the state space. As a result, graph node $v$ and pseudo node $u$ in the state space can be further described as
\begin{equation}
\begin{aligned}
    &u \triangleq \left(\mathrb{r}_u, \mathrb{m}_u, \{e_{u, w}, e_{w, u}|w\in\left(\mathcal{U}\cup\mathcal{V}\right)\}\right),\\
    &v \triangleq \left(\mathrb{q}_v, \mathrb{m}_v, \{e_{v, w}, e_{w, v}|w\in\left(\mathcal{N}(v)\cup\mathcal{U}\right)\}\right).
\end{aligned}
\end{equation} 
The messages of graph nodes are initialized with node features. Each graph edge $e_{v, v'}$ is characterized by its edge feature $\mathrb{e}_{v, v'}$.

\subsection{Dynamic Message Passing in the State Space}\label{ssec:dy-mp}
Given the measured proximity as message pathways and the messages to be passed, the embedded nodes can interact with each other through message passing. Specifically, the interactions between graph nodes are performed in both local and global scope.

\textbf{Global message passing.}~
In the state space, graph nodes perform global message passing based on their states, where pseudo nodes serve as proxies. Given graph node states $\mathrb{Q}\in\mathbb{R}^{n\times q}$, pseudo-node states $\mathrb{R}\in\mathbb{R}^{n_p\times q}$ and graph node messages $\mathrb{M}^{\mathtt{n}}=(\mathrb{m}_{v_1}, \cdots, \mathrb{m}_{v_n})^\top\in \mathbb{R}^{n\times d}$, the global message-passing process can be formulated as
\begin{align}
    &\texttt{(Diffuse)}
    &&\mathrb{G} =\mathrb{E}^\mathtt{np}\mathrb{M}^{\mathtt{n}}, \quad
    \mathrb{E}^\mathtt{np}_{ij} = \!e_{u_i, v_j}\! = \psi(\mathrb{R}_{i, \cdot}, \mathrb{Q}_{j, \cdot}), \label{eq:glob-mp-np}\\
    &\texttt{(Refine)}
    &&\mathrb{\hat{G}} = \mathrb{E}^\mathtt{pp}\mathrb{G}, \quad
    \mathrb{E}^\mathtt{pp}_{ij} = \!e_{u_i, u_j}\! = \psi(\mathrb{R}_{i, \cdot}, \mathrb{R}_{j, \cdot}),\label{eq:glob-mp-agg}\\
    &\
    &&\Delta\mathrb{R} = \mathtt{NL}(\mathrb{\hat{G}}), \quad
    \mathrb{M}^{\mathtt{p}} = \mathtt{NL}(\mathrb{\hat{G}}),  \nonumber\\
    &\texttt{(Collect)}
    &&\mathrb{M}^\mathtt{glob}\! =\! \mathrb{E}^\mathtt{pn}\mathrb{M}^{\mathtt{p}},~
    \mathrb{E}^\mathtt{pn}_{ij} \!= \!e_{v_i, u_j}\! =\! \psi(\mathrb{Q}_{i, \cdot}, [\mathrb{R} \!+\! \Delta\mathrb{R}]_{j, \cdot}), \label{eq:glob-mp-pn}
\end{align}
where $\mathrb{E}^\mathtt{np}\in\mathbb{R}^{n_p\times n}$ denotes the edge weight matrix from graph nodes to pseudo nodes, $\mathrb{E}^\mathtt{pp}$ and $\mathrb{E}^\mathtt{pn}$ follow the similar name rule. An example of $\mathrb{E}^\mathtt{np}_{ij}$ computation is illustrated in Fig.~\ref{fig:main}. Eq.~\ref{eq:glob-mp-np} formulates the process that graph nodes diffuse messages to pseudo nodes. Eq.~\ref{eq:glob-mp-agg} formulates the global feature refinement at the pseudo-node level, where $\Delta\mathrb{R}\in\mathbb{R}^{n_p\times q}$ denotes the learned displacements for pseudo nodes, $\mathrb{M}^{\mathtt{p}}\in\mathbb{R}^{n_p\times d}$ encodes the refined pseudo-node messages with global information. Eq.~\ref{eq:glob-mp-pn} formulates the message collection process from pseudo nodes to graph nodes.

For simplicity, we compile Eq.~\ref{eq:glob-mp-np}-\ref{eq:glob-mp-pn} as $\mathrb{M}^{\mathtt{glob}}, \Delta\mathrb{R}=\mathtt{GlobMP}(\mathrb{Q}, \mathrb{M}^{\mathtt{n}}, \mathrb{R})$. Note that the space complexity of our global message passing is $O(knn_p)$ with $k, n_p\ll n$, significantly lower than $O(n^2)$ in dense global message passing.

\textbf{Local message passing.}~
Topology-coupled message passing is employed to encode the local structure. The resulted local message-passing process for graph node $v$ can be formulated as
\begin{equation}\label{eq:local-mp}
    \mathrb{m}^\mathtt{local}_v \!=\! \frac{1}{|\mathcal{N}(v)|+1}
    \left[ \mathrb{m}_v \!+\! \!
    \sum_{v'\in\mathcal{N}(v)}\mathtt{NL}\!(\mathrb{m}_{v'}||\mathrb{e}_{v,v'}) \right],
\end{equation}
where $||$ denotes the concatenate operation. Through local message passing, graph nodes aggregate messages from their adjacent nodes. Eq.~\ref{eq:local-mp} can be compiled as $\mathrb{M}^{\mathtt{local}}=\mathtt{LocalMP}(\mathrb{M}, \mathrb{E})$ for all the graph nodes.

\section{Implementing Dynamic Message Passing with \nsqure}\label{sec:implementation}
We further develop a GNN model named \nsqure~to move the embedded nodes to their optimal positions. By feeding the states recursively into a single recurrent layer, \nsqure~learns the displacements of all the embedded nodes in the state space and updates their positions. The changes in position reshape the spatial relations between the embedded nodes and thus reshape the dynamic message pathways. These dynamically evolving pathways empower \nsqure~to adapt to the specific positions of the embedded nodes at each recursive step. In this section, we first outline the key process in the $l$-th recursive step, \textit{i.e.}, pseudo-node adaptation and dynamic message passing, then describe different output fashion for downstream tasks.

\subsection{Pseudo-node Adaptation}
Pseudo nodes are initialized randomly in the state space. To adapt to specific input graphs, \nsqure~first diffuses graph node messages to pseudo nodes, adjusting pseudo-node states and corresponding messages accordingly. Given graph node messages $\mathrb{M}^{\mathtt{n}(l-1)}$~($\mathrb{M}^{\mathtt{n}(0)}=\mathrb{X}$), graph node states $\mathrb{Q}^{(l-1)}$ and pseudo-node states $\mathrb{R}^{(l-1)}$ at the $l$-th recursive step, the adaptation process can be formulated as
\begin{equation}\label{eq:adaptation}
\begin{aligned}
    &\mathrb{\hat{M}}^{\mathtt{glob}(l)}, \Delta\mathrb{\hat{R}}^{(l)} \!=\! \mathtt{GlobMP}
    \left(\mathrb{Q}^{(l-1)}, \mathrb{M}^{\mathtt{n}(l-1)}, \mathrb{R}^{(l-1)}\right),\\
    &\mathrb{\hat{R}}^{(l)} = \mathrb{R}^{(l-1)} + \Delta\mathrb{\hat{R}}^{(l)},
\end{aligned}
\end{equation}
where $\mathrb{\hat{M}}^{\mathtt{glob}(l)}$ denotes pseudo-node messages that are collected by graph nodes, serving as query signals for different patterns on graphs. $\mathrb{\hat{R}}^{(l)}$ denotes the adjusted pseudo-node states.

\begin{table*}
\tabcaption{\textbf{Graph classification results on small-scale benchmarks (measured by accuracy: \%).}}
\label{tab:graph-small}
\centering
\begin{small}
\begin{sc}
\resizebox{0.62\linewidth}{!}{
\begin{tabular}{lccccc} 
\toprule
                 & PROTEIN                 & NCI1                    & IMDB-B                  & IMDB-M                  & COLLAB                   \\ 
\midrule
\#Graphs         & 1,113                   & 4,110                   & 1,000                   & 1,500                   & 5,000                    \\
\#Nodes          & 39.06                   & 29.87                   & 19.77                   & 13.00                   & 74.49                    \\
\#Edges          & 145.60                  & 64.60                   & 193.10                  & 131.87                  & 4,914.4                   \\
\#Node Features  & 3                       & 37                      & 0                       & 0                       & 0                        \\ 
\midrule
PATCHY-SAN~\cite{niepert_LearningConvolutionalNeural_2016}       & 75.00$_{\color{gray}\pm2.51}$          & 78.60$_{\color{gray}\pm1.90}$          & 71.00$_{\color{gray}\pm2.29}$          & 45.23$_{\color{gray}\pm2.84}$          & 72.60$_{\color{gray}\pm2.15}$           \\
GCN~\cite{kipf_SemiSupervisedClassificationGraph_2017}              & 73.24$_{\color{gray}\pm0.73}$          & 76.29$_{\color{gray}\pm1.79}$          & 73.26$_{\color{gray}\pm0.46}$          & 50.39$_{\color{gray}\pm0.41}$          & 80.59$_{\color{gray}\pm0.27}$           \\
PG~\cite{huang_GoingDeeperPermutationSensitive_2022}               & 76.80$_{\color{gray}\pm3.80}$          & 82.80$_{\color{gray}\pm1.30}$          & 76.80$_{\color{gray}\pm2.60}$          & 53.20$_{\color{gray}\pm3.60}$          & 80.90$_{\color{gray}\pm0.80}$           \\ 
CoCN~\cite{sun_AllinARow_2023}             & 76.86$_{\color{gray}\pm0.13}$          & 82.89$_{\color{gray}\pm0.19}$          & 77.26$_{\color{gray}\pm0.27}$          & 56.32$_{\color{gray}\pm0.18}$          & 86.15$_{\color{gray}\pm0.10}$           \\
\midrule
GIN~\cite{xu*_HowPowerfulAre_2019}              & 73.84$_{\color{gray}\pm4.46}$          & 76.62$_{\color{gray}\pm1.80}$          & 72.78$_{\color{gray}\pm0.86}$          & 48.13$_{\color{gray}\pm1.36}$          & 78.19$_{\color{gray}\pm0.63}$           \\
+pseudo node~\cite{liu_BoostingGraphStructure_2022}     & 74.11$_{\color{gray}\pm4.12}$          & 77.08$_{\color{gray}\pm1.49}$          & -                       & -                       & -                        \\
GraphSAGE~\cite{hamilton_InductiveRepresentationLearning_2017}        & 73.48$_{\color{gray}\pm5.66}$          & 73.82$_{\color{gray}\pm2.17}$          & 68.80$_{\color{gray}\pm4.50}$          & 47.60$_{\color{gray}\pm3.50}$          & 73.90$_{\color{gray}\pm1.70}$           \\
+pseudo node~\cite{liu_BoostingGraphStructure_2022}     & 73.93$_{\color{gray}\pm5.68}$          & 74.31$_{\color{gray}\pm2.27}$          & -                       & -                       & -                        \\
DiffPool~\cite{ying_HierarchicalGraphRepresentation_2018}         & 75.62$_{\color{gray}\pm5.17}$          & 76.62$_{\color{gray}\pm1.93}$          & 73.14$_{\color{gray}\pm0.70}$          & 51.31$_{\color{gray}\pm0.72}$          & 82.13$_{\color{gray}\pm0.43}$           \\
+pseudo node~\cite{liu_BoostingGraphStructure_2022}     & 75.98$_{\color{gray}\pm3.89}$          & 77.08$_{\color{gray}\pm1.33}$          & -                       & -                       & -                        \\ 
\midrule
TopKPool~\cite{gao_GraphUNets_2019}         & 70.48$_{\color{gray}\pm1.01}$          & 67.02$_{\color{gray}\pm2.25}$          & 71.58$_{\color{gray}\pm0.95}$          & 48.59$_{\color{gray}\pm0.72}$          & 77.58$_{\color{gray}\pm0.85}$           \\
SAGPool~\cite{lee_SelfAttentionGraphPooling_2019}          & 71.56$_{\color{gray}\pm1.49}$          & 67.45$_{\color{gray}\pm1.11}$          & 72.55$_{\color{gray}\pm1.28}$          & 50.23$_{\color{gray}\pm0.44}$          & 78.03$_{\color{gray}\pm0.31}$           \\
StructPool~\cite{yuan_StructPoolStructuredGraph_2020}       & 75.16$_{\color{gray}\pm0.86}$          & 78.64$_{\color{gray}\pm1.53}$          & 72.06$_{\color{gray}\pm0.64}$          & 50.23$_{\color{gray}\pm0.53}$          & 77.27$_{\color{gray}\pm0.51}$           \\
SEP~\cite{wu_StructuralEntropyGuided_2022}              & 76.42$_{\color{gray}\pm0.39}$          & 79.35$_{\color{gray}\pm0.33}$          & 74.12$_{\color{gray}\pm0.56}$          & 51.53$_{\color{gray}\pm0.65}$          & 81.28$_{\color{gray}\pm0.15}$           \\
GMT~\cite{baek_AccurateLearningGraph_2022}              & 75.09$_{\color{gray}\pm0.59}$          & 76.35$_{\color{gray}\pm2.62}$          & 73.48$_{\color{gray}\pm0.76}$          & 50.66$_{\color{gray}\pm0.82}$          & 80.74$_{\color{gray}\pm0.54}$           \\ 
\midrule
\nsqure~(Ours)   & \textbf{77.53$_{\color{gray}\pm1.78}$} & \textbf{83.52$_{\color{gray}\pm3.75}$} & \textbf{79.95$_{\color{gray}\pm2.46}$} & \textbf{57.31$_{\color{gray}\pm2.19}$} & \textbf{86.72$_{\color{gray}\pm1.62}$}  \\
\bottomrule
\end{tabular}}
\end{sc}
\end{small}
\end{table*}

\subsection{Dynamic Message Passing}
\nsqure~performs both local and global message passing. At the local level, graph nodes exchange their own messages $\mathrb{M}^{\mathtt{n}(l-1)}$, collected messages $\mathrb{\hat{M}}^{\mathtt{glob}(l)}$ and graph node states $\mathrb{Q}^{(l-1)}$:
\begin{equation}\label{eq:nsqure-local-mp}
\begin{aligned}
    &\mathrb{M}^{\mathtt{local}(l)}=\mathtt{LocalMP}
    \left[
    \left(\mathrb{M}^{\mathtt{n}(l-1)} \| \mathrb{\hat{M}}^{\mathtt{glob}(l)} \| \mathrb{Q}^{(l-1)}\right), \mathrb{E}\right], \\
    &\mathrb{\hat{Q}}^{(l)} = \mathrb{Q}^{(l-1)} + \mathtt{NL}(\mathrb{M}^{\mathtt{local}(l)}).
\end{aligned}
\end{equation}
Through local message passing, graph node messages $\mathrb{M}^{\mathtt{local}(l)}$ are generated in response to the query messages $\mathrb{\hat{M}}^{\mathtt{glob}(l)}$. \nsqure~then sends the updated messages to global message passing:
\begin{equation}\label{eq:nsqure-global-mp}
\begin{aligned}
    &\mathrb{M}^{\mathtt{glob}(l)}, \Delta\mathrb{R}^{(l)} \!=\! \mathtt{GlobMP}
    \left(\mathrb{\hat{Q}}^{(l)}, \mathrb{M}^{\mathtt{local}(l)}, \mathrb{\hat{R}}^{(l)}\right),\\
    &\mathrb{Q}^{(l)} = \mathrb{\hat{Q}}^{(l)} + \mathtt{NL}(\mathrb{M}^{\mathtt{glob}(l)}),\quad
    \mathrb{M}^{\mathtt{n}(l)} = \mathrb{M}^{\mathtt{n}(l-1)} + \mathrb{M}^{\mathtt{glob}(l)}, \\
    &\mathrb{R}^{(l)} = \mathrb{\hat{R}}^{(l)} + \Delta\mathrb{R}^{(l)}.
\end{aligned}
\end{equation}

\subsection{Output Module}
\nsqure~updates the states of the embedded nodes recursively with a single recurrent layer. Instead of employing different layers for each recursive step, the associated parameters are shared across steps. After $L$ recursive steps, the embedded nodes now reach their final states $\mathrb{Q}^{(L)}$ and $\mathrb{R}^{(L)}$. \nsqure~then takes graph node states and pseudo-node states as the learned representation for node-level and graph-level tasks, respectively. For graph classification, \nsqure~further applies $\mathtt{NL}(\cdot)$ to aggregate the pseudo-node states. To make class predictions, \nsqure~employs learnable parameter $\mathrb{C}\in\mathbb{R}^{n_c\times q}$ as the states of $n_c$ class nodes and outputs the proximity between the recursive output and the class nodes.

\section{Experiment}\label{sec:exp}
In this section, we provide empirical evaluation results of \nsqure~on real-world benchmarks. \nsqure~is implemented with PyTorch \cite{paszke_PyTorchImperativeStyle_2019} and PyTorch Geometric \cite{fey_FastGraphRepresentation_2019}, and trained on a single Nvidia Geforce RTX 4090. The detailed experimental settings are presented in Appendix~\ref{sec:app-exp-setup}.

\subsection{Graph Classification}
\textbf{Experimental setups.}~
We adopt six benchmarks including three biochemical datasets (OGB-molpcba~\cite{hu_OpenGraphBenchmark_2020}, PROTEINS~\cite{morris_TUDatasetCollectionBenchmark_2020}, NCI1~\cite{morris_TUDatasetCollectionBenchmark_2020}) and three social network datasets~\cite{morris_TUDatasetCollectionBenchmark_2020} (COLLAB, IMDB-BINARY and IMDB-MULTI). The benchmark statistics are summarized in Tab.~\ref{tab:graph-small} and Tab.~\ref{tab:graph-large}.
We choose convolutional GNNs, GNNs with a single pseudo node, hierarchical GNNs, and graph transformers as the baselines of graph classification. 
For more details, please refer to Appendix.~\ref{ssec:graph-setup}.

\textbf{Performance.}~
\nsqure~and baselines are evaluated on both small-scale and large-scale benchmarks. The evaluation results in Tab.~\ref{tab:graph-small} and Tab.~\ref{tab:graph-large} showcase the ability of \nsqure~to outperform various GNNs and graph transformers. Especially on the large-scale benchmark OGB-molpcba, \nsqure~surpasses baseline models with only 500K parameters, while Graphormer reaches 31.39\% average precision with 119.5M parameters. Compared to GNNs with a single pseudo node, \nsqure~gains significant improvements. This demonstrates the effectiveness of our common state space where pseudo nodes and graph nodes can interact dynamically with each other.

\begin{table*}
    \begin{minipage}{0.33\linewidth}
        \tabcaption{\textbf{Graph classification results on large-scale benchmark OGB-molpcba.}}
        \label{tab:graph-large}
        \centering
        \resizebox{\linewidth}{!}{
        \begin{small}
\begin{sc}
\begin{tabular}{lcr} 
\toprule
\multicolumn{1}{c}{\#Graphs}  & \#Nodes                                                          & \multicolumn{1}{c}{\#Edges}                                               \\
\multicolumn{1}{c}{437,929}   & 26                                                               & \multicolumn{1}{c}{28.1}                                                  \\
\midrule
Methods                       & \begin{tabular}[c]{@{}c@{}}Average \\Precision \\(\%)\end{tabular} & \multicolumn{1}{c}{\begin{tabular}[c]{@{}c@{}}\#Params\\(K)\end{tabular}}  \\ 
\midrule
GCN~\cite{kipf_SemiSupervisedClassificationGraph_2017}                           & 20.20$_{\color{gray}\pm0.24}$                                    & 565                                                                       \\
+pseudo node~\cite{hu_OpenGraphBenchmark_2020}                  & 24.24$_{\color{gray}\pm0.34}$                                    & 2,017                                                                     \\
GIN~\cite{xu*_HowPowerfulAre_2019}                           & 22.66$_{\color{gray}\pm0.28}$                                    & 1,923                                                                     \\
+pseudo node~\cite{hu_OpenGraphBenchmark_2020}                  & 27.03$_{\color{gray}\pm0.23}$                                    & 3,374                                                                     \\
MPNN                                                        & - & - \\
+pseudo node~\cite{cai_ConnectionMPNNGraph_2023}              & 28.48$_{\color{gray}\pm0.26}$                                    & -                                                                         \\ 
\midrule
GraphTrans~\cite{cai_ConnectionMPNNGraph_2023}                    & 27.61$_{\color{gray}\pm0.29}$                                    & -                                                                         \\
SAN~\cite{kreuzer_RethinkingGraphTransformers_2021}                           & 27.65$_{\color{gray}\pm0.42}$                                    & 5,885                                                                     \\
GraphGPS~\cite{rampasek_RecipeGeneralPowerful_2022}                      & 29.07$_{\color{gray}\pm0.28}$                                    & 9,745                                                                     \\
Graphormer~\cite{ying_TransformersReallyPerform_2021}                    & 31.39$_{\color{gray}\pm0.32}$                                    & 119,529                                                                   \\ 
\midrule
\nsqure~(Ours) & \textbf{33.90$_{\color{gray}\pm0.73}$}                           & 516                                                                       \\
\bottomrule
\end{tabular}
\end{sc}
\end{small}}
    \end{minipage}
    \hfill
    \begin{minipage}{0.65\linewidth}
        \tabcaption{\textbf{Node classification results on small-scale homophilic graphs (measured by accuracy: \%).}}
        \label{tab:homo-node-small}
        \centering
        \resizebox{\linewidth}{!}{
\begin{small}
\begin{sc}
\begin{tabular}{lcccc}
\toprule
             & CoauthorCS & CoauthorPhy     & AmzPhoto      & AmzComputers  \\
\midrule
\#Nodes      & 18,333     & 34,493          & 7,487          & 13,381           \\
\#Edges      & 81,894     & 247,962         & 119,043        & 245,778          \\ 
\midrule
{\footnotesize GCN~\cite{kipf_SemiSupervisedClassificationGraph_2017}}          & 92.92$_{\color{gray}\pm0.12}$ & 96.18$_{\color{gray}\pm0.07}$      & 92.70$_{\color{gray}\pm0.20}$     & 89.65$_{\color{gray}\pm0.52}$       \\
{\footnotesize GAT~\cite{velickovic_GraphAttentionNetworks_2018}}          & 93.61$_{\color{gray}\pm0.14}$ & 96.17$_{\color{gray}\pm0.08}$      & 93.87$_{\color{gray}\pm0.11}$     & 90.78$_{\color{gray}\pm0.17}$       \\
{\footnotesize GPRGNN~\cite{chien_AdaptiveUniversalGeneralized_2022}}       & 95.13$_{\color{gray}\pm0.09}$ & 96.85$_{\color{gray}\pm0.08}$      & 94.49$_{\color{gray}\pm0.14}$     & 89.32$_{\color{gray}\pm0.29}$       \\
{\footnotesize APPNP~\cite{gasteiger_PredictThenPropagate_2018}}        & 94.49$_{\color{gray}\pm0.07}$ & 96.54$_{\color{gray}\pm0.07}$      & 94.32$_{\color{gray}\pm0.14}$     & 90.18$_{\color{gray}\pm0.17}$       \\
\midrule
{\footnotesize GT~\cite{shi_MaskedLabelPrediction_2021}}           & 94.64$_{\color{gray}\pm0.13}$ & 97.05$_{\color{gray}\pm0.05}$      & 94.74$_{\color{gray}\pm0.13}$     & 91.18$_{\color{gray}\pm0.17}$       \\
{\footnotesize Graphormer~\cite{ying_TransformersReallyPerform_2021}}   & OOM        & OOM             & 92.74$_{\color{gray}\pm0.14}$     & OOM              \\
{\footnotesize SAN~\cite{kreuzer_RethinkingGraphTransformers_2021}}          & 94.51$_{\color{gray}\pm0.15}$ & OOM             & 94.86$_{\color{gray}\pm0.10}$     & 89.83$_{\color{gray}\pm0.16}$       \\
{\footnotesize GraphGPS~\cite{rampasek_RecipeGeneralPowerful_2022}}     & 93.93$_{\color{gray}\pm0.12}$ & OOM             & 95.06$_{\color{gray}\pm0.13}$     & OOM              \\
{\footnotesize NAGphormer~\cite{chen_NAGphormerTokenizedGraph_2023}}   & 95.75$_{\color{gray}\pm0.09}$ & 97.34$_{\color{gray}\pm0.03}$      & 95.49$_{\color{gray}\pm0.11}$     & 91.22$_{\color{gray}\pm0.14}$       \\
{\footnotesize Exphormer~\cite{shirzad_ExphormerSparseTransformers_2023}}    & \textbf{95.77$_{\color{gray}\pm0.15}$} & 97.16$_{\color{gray}\pm0.13}$      & 95.27$_{\color{gray}\pm0.42}$     & 91.59$_{\color{gray}\pm0.31}$       \\ 
\midrule
{\footnotesize \nsqure~(Ours)} & 94.44$_{\color{gray}\pm0.45}$      & \textbf{97.56$_{\color{gray}\pm0.28}$}  & \textbf{95.75$_{\color{gray}\pm0.34}$} & \textbf{92.51$_{\color{gray}\pm0.13}$}   \\
\bottomrule
\end{tabular}
\end{sc}
\end{small}
}
    \end{minipage}
\end{table*}

\begin{wrapfigure}{r}{0.65\linewidth}
\tabcaption{\textbf{Node classification results on small-scale heterophilic graphs (measured by ROC-AUC except accuracy for amazon-ratings: \%).} $\dagger$ denotes our reproduced results.}
\label{tab:hetero-node-small}
\centering
\begin{small}
\begin{sc}
\resizebox{0.98\linewidth}{!}{
\begin{tabular}{lcccc} 
\toprule
           & Questions      & Amazon-ratings & Tolokers       & Minesweeper     \\
\midrule
\#Nodes    & 48,921         & 24,492         & 11,758         & 10,000          \\
\#Edges    & 153,540        & 93,050         & 519,000        & 39,402          \\ 
\midrule
SGC~\cite{wu_SimplifyingGraphConvolutional_2019}                  & 75.91$_{\color{gray}\pm0.96}$     & 50.66$_{\color{gray}\pm0.48}$     & 80.70$_{\color{gray}\pm0.97}$     & 70.88$_{\color{gray}\pm0.90}$      \\
GCN~\cite{kipf_SemiSupervisedClassificationGraph_2017}                  & 76.09$_{\color{gray}\pm1.27}$     & 48.70$_{\color{gray}\pm0.63}$     & 83.64$_{\color{gray}\pm0.67}$     & 89.75$_{\color{gray}\pm0.52}$      \\
GAT~\cite{velickovic_GraphAttentionNetworks_2018}                  & 77.43$_{\color{gray}\pm1.20}$     & 49.09$_{\color{gray}\pm0.63}$     & 83.70$_{\color{gray}\pm0.47}$     & 92.01$_{\color{gray}\pm0.68}$      \\
GPRGNN~\cite{chien_AdaptiveUniversalGeneralized_2022}               & 55.48$_{\color{gray}\pm0.91}$     & 44.88$_{\color{gray}\pm0.34}$     & 72.94$_{\color{gray}\pm0.97}$     & 86.24$_{\color{gray}\pm0.61}$      \\
H$_2$GCN~\cite{zhu_HomophilyGraphNeural_2020}                & 63.59$_{\color{gray}\pm1.46}$     & 36.47$_{\color{gray}\pm0.23}$     & 73.35$_{\color{gray}\pm1.01}$     & 89.71$_{\color{gray}\pm0.31}$      \\
FAGCN~\cite{bo_LowfrequencyInformationGraph_2021}                & 77.24$_{\color{gray}\pm1.26}$     & 44.12$_{\color{gray}\pm0.30}$     & 77.75$_{\color{gray}\pm1.05}$     & 88.17$_{\color{gray}\pm0.73}$      \\
GloGNN~\cite{li_FindingGlobalHomophily_2022}               & 65.74$_{\color{gray}\pm1.19}$     & 36.89$_{\color{gray}\pm0.14}$     & 73.39$_{\color{gray}\pm1.17}$     & 51.08$_{\color{gray}\pm1.23}$      \\
\midrule          
GT~\cite{shi_MaskedLabelPrediction_2021}                   & 77.95$_{\color{gray}\pm0.68}$     & \textbf{51.17$_{\color{gray}\pm0.66}$}     & 83.23$_{\color{gray}\pm0.64}$     & 91.85$_{\color{gray}\pm0.76}$      \\
Graphormer~\cite{ying_TransformersReallyPerform_2021}           & OOM                & OOM            & OOM            & OOM             \\
GraphGPS~\cite{rampasek_RecipeGeneralPowerful_2022}             & OOM                & OOM            & 84.70$_{\color{gray}\pm0.56}$     & 92.29$_{\color{gray}\pm0.61}$      \\
Exphormer~\cite{shirzad_ExphormerSparseTransformers_2023} $\dagger$  & 73.86$_{\color{gray}\pm0.58}$     & 49.36$_{\color{gray}\pm0.36}$  & 84.20$_{\color{gray}\pm0.22}$   & 90.42$_{\color{gray}\pm0.10}$           \\ 
\midrule
\nsqure~(Ours)    & \textbf{78.07$_{\color{gray}\pm0.63}$} & 50.25$_{\color{gray}\pm0.53}$     & \textbf{86.25$_{\color{gray}\pm0.41}$} & \textbf{93.97$_{\color{gray}\pm0.27}$}  \\
\bottomrule
\end{tabular}
}
\end{sc}
\end{small}
\end{wrapfigure}
\subsection{Node Classification}
\textbf{Experimental setups.}~
For node classification, we conduct experiments on (\textbf{1}) six small-scale benchmarks: homophilic graphs (AmazonPhoto, AmazonComputers, CoauthorCS, and CoauthorPhysics)~\cite{shchur_PitfallsGraphNeural_2019}, heterophilic graphs (questions, amazon-ratings, tolokers, and minesweeper)~\cite{platonov_CriticalLookEvaluation_2023}; and (\textbf{2}) four large-scale benchmarks: homophilic graphs (OGB-arXiv, OGB-proteins)~\cite{hu_OpenGraphBenchmark_2020}, heterophilic graphs (arXiv-year, genius)~\cite{lim_LargeScaleLearning_2021}. The statistics of node classification benchmarks are summarized in Tab.~\ref{tab:homo-node-small}-\ref{tab:node-large}.
We choose convolutional GNNs and graph transformers as the baselines. 
For detailed baseline setups, please refer to Appendix.~\ref{ssec:node-setup}.

\textbf{Performance.}~
The evaluation results are presented in Tab.~\ref{tab:homo-node-small}-\ref{tab:node-large}. \nsqure~reaches superior or comparable performance against strong GNN baselines on both small-scale and large-scale benchmarks. Compared to graph transformers based on dense attention, including Graphormer, SAN, and GraphGPS, \nsqure~surpasses the baselines on small-scale benchmarks and successfully scales to large-scale benchmarks. Note that Exphormer based on sparse attention also suffers from the out-of-memory problem in our experimental environment. For other sparse graph transformers, \nsqure~can achieve comparable results and perform consistently between heterophilic and homophilic benchmarks.

\begin{table*}
    \begin{minipage}{0.65\linewidth}
        \tabcaption{\textbf{Node classification results on large-scale benchmarks (measured by ROC-AUC/accuracy: \%).}~$\dagger$ denotes our reproduced results.}
        \label{tab:node-large}
        \centering
        \resizebox{\linewidth}{!}{
\begin{small}
\begin{sc}
\resizebox{0.98\linewidth}{!}{
\begin{tabular}{lcccc} 
\toprule
              & Genius     & arXiv-year & OGB-arXiv        & OGB-proteins      \\
Metrics      & ROC-AUC     & accuracy   & accuracy        & ROC-AUC                 \\
\midrule
\#Nodes       & 421,961    & 169,343    & 169,343          & 132,534           \\
\#Edges       & 984,979    & 1,166,243  & 1,166,243        & 39,561,252        \\ 
\midrule
SGC~\cite{wu_SimplifyingGraphConvolutional_2019}           & 82.36$_{\color{gray}\pm0.37}$ & 32.83$_{\color{gray}\pm0.13}$ & 67.79$_{\color{gray}\pm0.27}$       & 70.31$_{\color{gray}\pm0.23}$        \\
GCN~\cite{kipf_SemiSupervisedClassificationGraph_2017}           & 87.42$_{\color{gray}\pm0.37}$ & 46.02$_{\color{gray}\pm0.26}$ & 71.74$_{\color{gray}\pm0.29}$       & 72.51$_{\color{gray}\pm0.35}$        \\
GAT~\cite{velickovic_GraphAttentionNetworks_2018}           & 55.80$_{\color{gray}\pm0.87}$ & 46.05$_{\color{gray}\pm0.51}$ & 67.63$_{\color{gray}\pm0.23}$       & 74.63$_{\color{gray}\pm1.24}$    \\
APPNP~\cite{gasteiger_PredictThenPropagate_2018}         & 85.36$_{\color{gray}\pm0.62}$ & 38.15$_{\color{gray}\pm0.26}$ & - & -  \\
H$_2$GCN~\cite{zhu_HomophilyGraphNeural_2020}         & OOM        & 49.09$_{\color{gray}\pm0.10}$ & OOM              & OOM               \\
LINKX~\cite{lim_LargeScaleLearning_2021}         & \textbf{90.77$_{\color{gray}\pm0.27}$} & 56.00$_{\color{gray}\pm1.34}$ & - & -  \\
\midrule
Graphormer~\cite{ying_TransformersReallyPerform_2021}    & \multicolumn{4}{c}{}                       \\
SAN~\cite{kreuzer_RethinkingGraphTransformers_2021}           & \multicolumn{4}{c}{OOM}                    \\
GraphGPS~\cite{rampasek_RecipeGeneralPowerful_2022}      & \multicolumn{4}{c}{}                                           \\
Exphormer~\cite{shirzad_ExphormerSparseTransformers_2023}     & OOM        & OOM        & 72.44$_{\color{gray}\pm0.28}$   & OOM               \\
Nodeformer~\cite{wu_NodeFormerScalableGraph_2022}    & 88.62$_{\color{gray}\pm0.27}$ ($\dagger$)  & 37.68$_{\color{gray}\pm0.30}$ ($\dagger$)  & 59.90$_{\color{gray}\pm0.42}$       & 77.45$_{\color{gray}\pm1.15}$        \\
SGFormer~\cite{wu_SGFormerSimplifyingEmpowering_2023}      & 83.91$_{\color{gray}\pm2.60}$ ($\dagger$)  & 44.34$_{\color{gray}\pm0.07}$ ($\dagger$)  & \textbf{72.63$_{\color{gray}\pm0.13}$}       & 79.53$_{\color{gray}\pm0.38}$        \\ 
\midrule
\nsqure~(Ours) & 89.32$_{\color{gray}\pm0.26}$     & \textbf{58.69$_{\color{gray}\pm0.42}$}      & 70.01$_{\color{gray}\pm0.65}$            & \textbf{79.55$_{\color{gray}\pm0.79}$}             \\
\bottomrule
\end{tabular}
}
\end{sc}
\end{small}
}
    \end{minipage}
    \begin{minipage}{0.31\linewidth}
        \centering
        \includegraphics[width=0.99\textwidth]{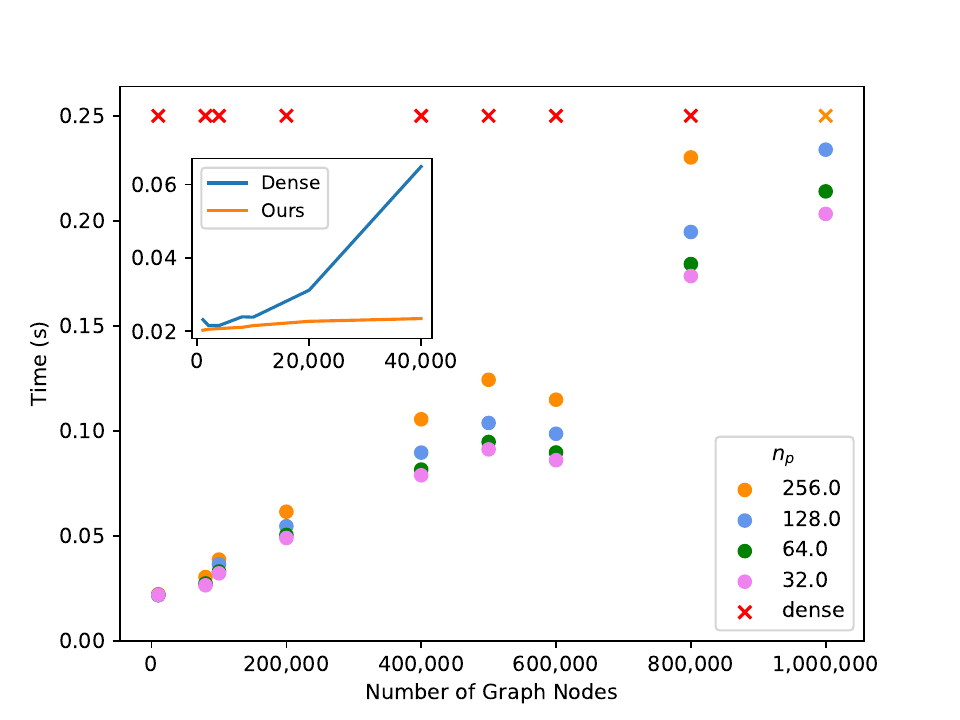}
        \figcaption{\textbf{Complexity analysis.} The results under different numbers of graph nodes are compared. Marker \texttt{X} denotes out-of-memory. \texttt{Dense} denotes dense pairwise relation modeling.}
        \label{fig:time}
    \end{minipage}
\end{table*}

\subsection{Model Analysis}
\subsubsection{Complexity Analysis}
Our proposed dynamic message passing enables graph nodes to access each other without dense pairwise modeling. 
As described in Sec.~\ref{ssec:dy-mp}, the space complexity of our global message passing is $O(knn_p)$. We further conduct an empirical analysis of \nsqure~on computational complexity. As presented in Fig.~\ref{fig:time}, the computational time exhibits linear scalability for the number of graph nodes $n$, while being insensitive to the number of pseudo nodes $n_p$. 
In comparison to dense pairwise modeling, we substitute the global message passing in \nsqure~with a dense attention scheme. The resulting growth rate of computational time for \nsqure~is significantly lower than that of the dense method. Moreover, the dense method fails to extend to large-scale graphs, whereas \nsqure~only encounters out-of-memory under the extreme situation of $n_p=256$ and $n=1\mathtt{M}$.

\begin{wrapfigure}{r}{0.36\linewidth}
\centering
\vskip -0.3in
\subfigure[Over-squashing]{\label{fig:squashing}
\begin{minipage}{\linewidth}
    \centering
    \includegraphics[width=\textwidth]{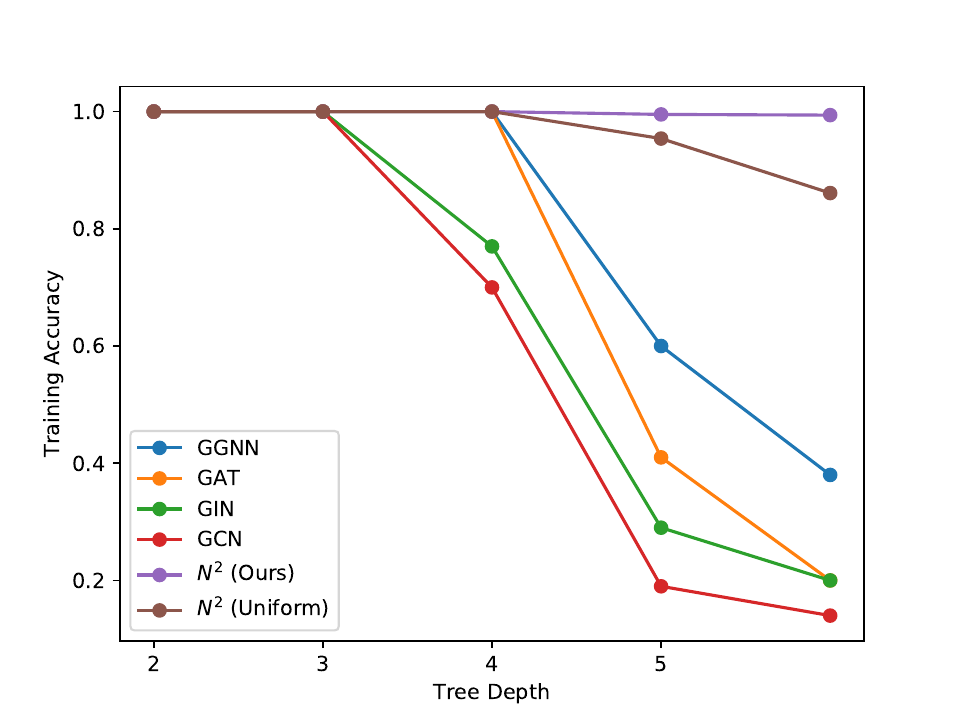}
\end{minipage}}
\subfigure[Over-smoothing]{\label{fig:smoothing}
\begin{minipage}{\linewidth}
    \centering
    \includegraphics[width=\textwidth]{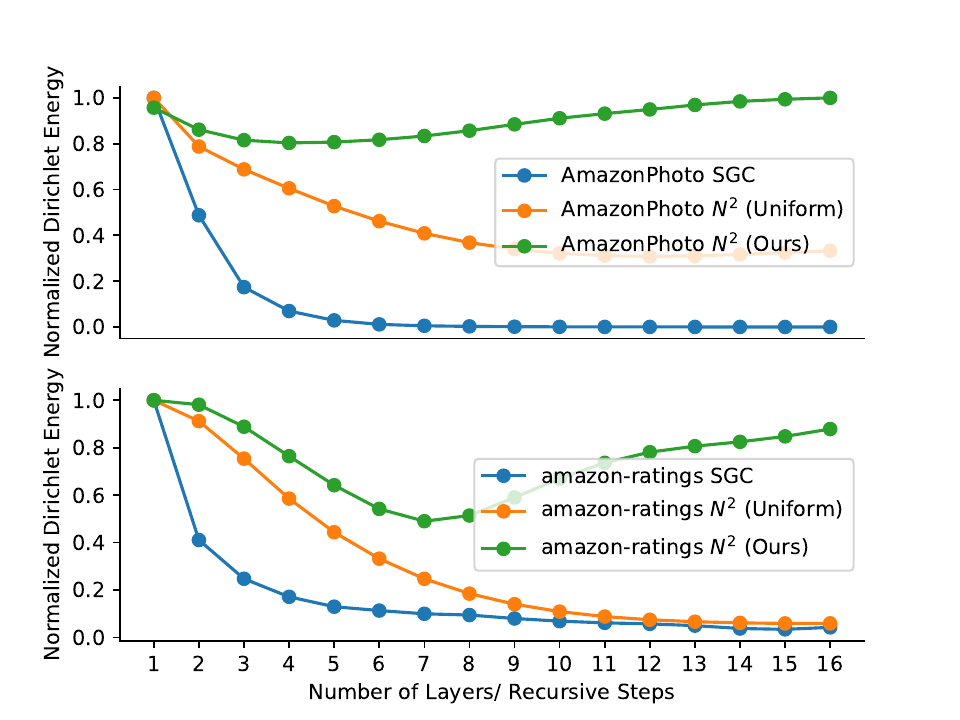}
\end{minipage}}
\figcaption{\textbf{Effectiveness study.}}
\vskip -0.1in
\end{wrapfigure}
\subsubsection{Effectiveness of \nsqure}
\textbf{Tackling over-squashing.}~
\nsqure~employs pseudo nodes in the common state space to achieve global message passing with linear complexity. These pseudo nodes also benefit \nsqure~in tackling over-squashing from two aspects, \textit{i.e.}, avoiding forming new bottlenecks on pseudo nodes and detouring from the bottlenecks on input graphs. 
Specifically, our pseudo nodes avoid becoming new bottlenecks by employing dynamic connections. Compared to the uniform connection, dynamic connection learns the specific edge weights for each pseudo node to different graph nodes. During global message passing, these weights can eliminate messages from certain graph nodes while preserving the others. As a result, the messages will not be squashed equally into the pseudo nodes, preventing the nodes from becoming new information bottlenecks.
Regarding the bottlenecks on input graphs, the pseudo nodes produce two-hop message highways for message passing, detouring messages from these bottlenecks.
To evaluate \nsqure~in tackling over-squashing, we conduct experiments on Trees Match~\cite{alon_BottleneckGraphNeural_2021} with training accuracy results presented in Fig.~\ref{fig:squashing}. \nsqure~maintains the fitting ability across different tree depths, demonstrating its ability to counteract over-squashing. Fig.~\ref{fig:squashing} also verifies the superiority of the dynamic connection in tackling over-squashing, where \nsqure~with the uniform connection shows a decline in accuracy as the depth increases.

\textbf{Tackling over-smoothing.}~
The other issue encountered when message passing relies heavily on input topology is over-smoothing~\cite{oono_GraphNeuralNetworks_2021}. 
In \nsqure, pseudo nodes are adopted as message pathway alternates, decoupling the message-passing process from input topology. During global message passing, our dynamic connection empowers the connected graph nodes on the input graphs to receive different outputs and avoid becoming too similar to each other. For local message passing, learning node displacements in \nsqure~allows the combination of local message-passing outputs with layer inputs and global message-passing outputs. Although stacking multi-layer local message passing leads to over-smoothing, the layer inputs and global outputs can maintain the high-frequency signals. To evaluate whether \nsqure~can alleviate the over-smoothing issue, we explore the changes in the Dirichlet energy~\cite{cai_NoteOverSmoothingGraph_2020} as the number of recursive steps increases. Fig.~\ref{fig:smoothing} compares among \nsqure, \nsqure~with the uniform connection, and SGC~\cite{wu_SimplifyingGraphConvolutional_2019} on AmazonPhoto and amazon-ratings. The reported values are normalized to [0, 1] by dividing the maximum energy values. Results show that \nsqure~with uniform connections encounters over-smoothing as the number of recursive steps increases, while \nsqure~with dynamic connections can maintain the Dirichlet energy and alleviate over-smoothing.

\begin{wrapfigure}{r}{0.56\linewidth}
\centering
\vspace{-0.6in}
\subfigure[Epoch=20]{
      \begin{minipage}[t]{0.49\linewidth}
      \centering
      \includegraphics[width=\textwidth]{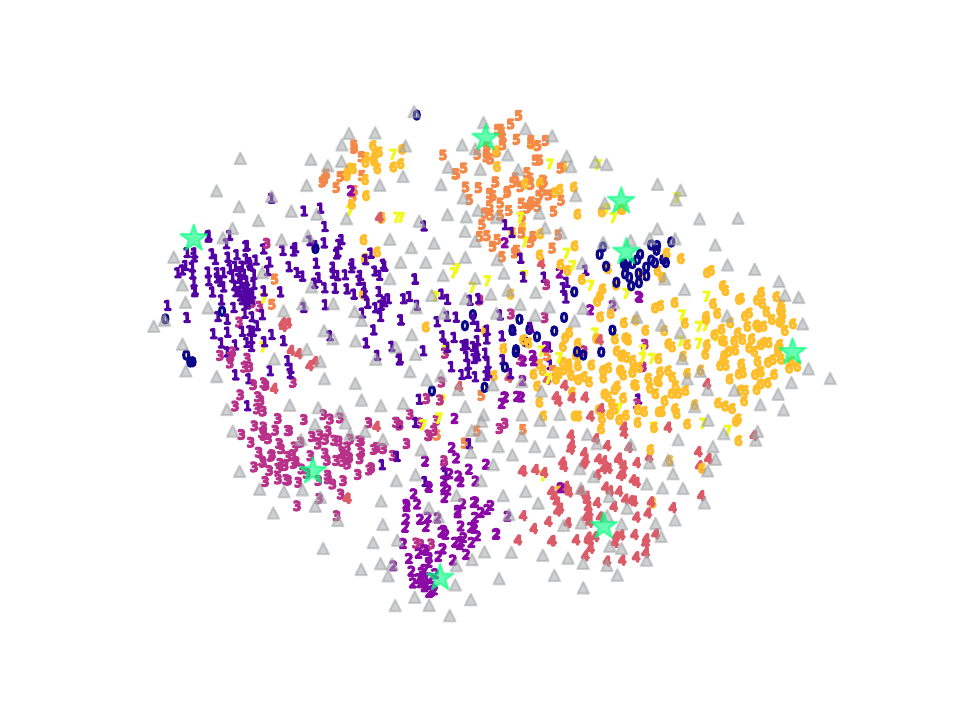}
      \end{minipage}}
\hspace{-0.2in}
 \subfigure[Epoch=500]{
      \begin{minipage}[t]{0.49\linewidth}
      \centering
      \includegraphics[width=\textwidth]{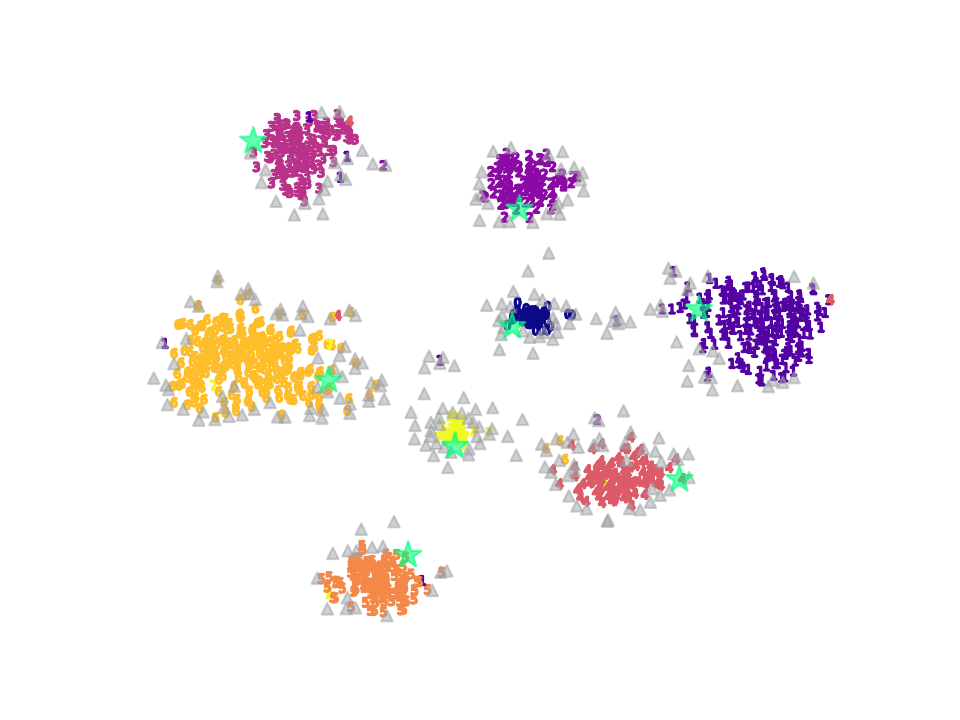}
      \end{minipage}}
\caption{\textbf{Distribution of embedded nodes.} The t-sne~\cite{t-SNE} results under different training epochs are compared. \textbf{\textcolor[RGB]{22,17,139}{0}, \textcolor[RGB]{83,2,163}{1}, \textcolor[RGB]{139,10,165}{2}, \textcolor[RGB]{184,50,137}{3}, \textcolor[RGB]{219,92,104}{4}, \textcolor[RGB]{244,136,74}{5}} and \textbf{ \textcolor[RGB]{254,189,42}{6}} denote graph nodes with different labels. \textbf{\textcolor[RGB]{197,198,199}{$\bigtriangleup$}} denotes pseudo nodes. \textbf{\textcolor[RGB]{76,255,164}{$\star$}} denotes class nodes.}
\label{fig:distribution}
\vspace{-0.1in}
\end{wrapfigure}



\subsubsection{Distribution of Embedded Nodes}
We visualize the distribution of all the embedded nodes during training. Results on AmazonPhoto at the first recursive step are depicted in Fig.~\ref{fig:distribution}. For results through recursion, please refer to Appendix~\ref{ssec:app-distribution}. We can see that as the graph nodes are clustered with different class nodes, pseudo nodes are also split into several groups. Each pseudo-node group is attracted toward a distinct graph node cluster. This indicates that pseudo nodes will attend a particular graph node group, serving as the global message pathways to other groups. As a result, each pseudo node assumes a balanced fraction of the global message load, mitigating the risk of becoming bottlenecks for message passing.

\begin{figure*}
    \begin{minipage}{0.54\linewidth}
        \centering
        \subfigure[$\mathrb{E}^{\mathtt{np}}$]{
              \begin{minipage}[t]{0.48\linewidth}
              \centering
              \includegraphics[width=\textwidth]{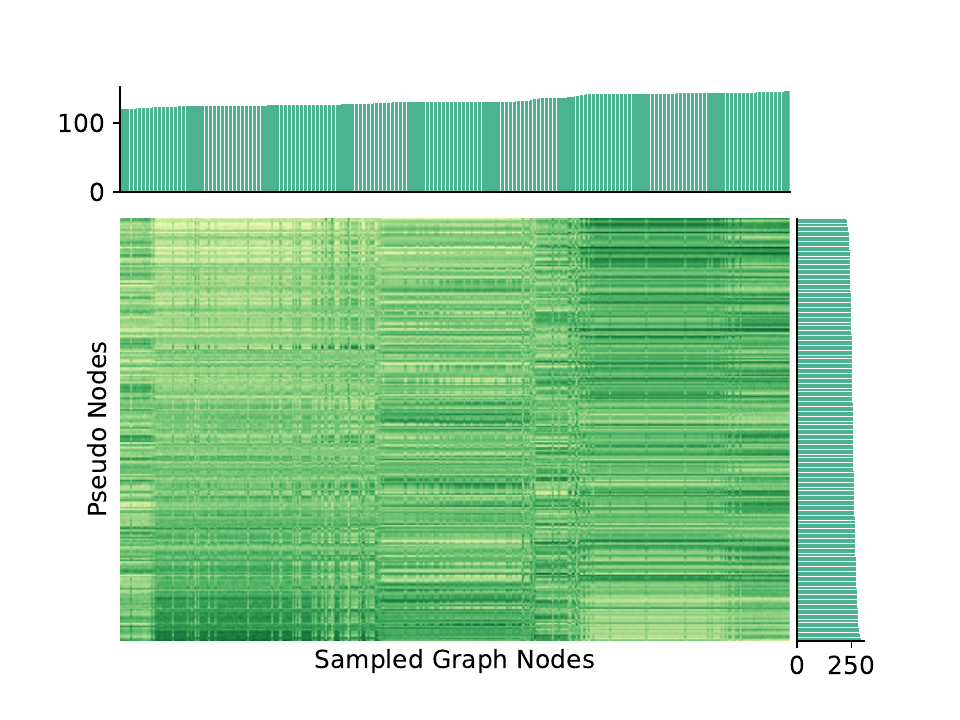}
              \end{minipage}}%
        \hfill
        \subfigure[$\mathrb{E}^{\mathtt{pn}\top}$]{
              \begin{minipage}[t]{0.48\linewidth}
              \centering
              \includegraphics[width=\textwidth]{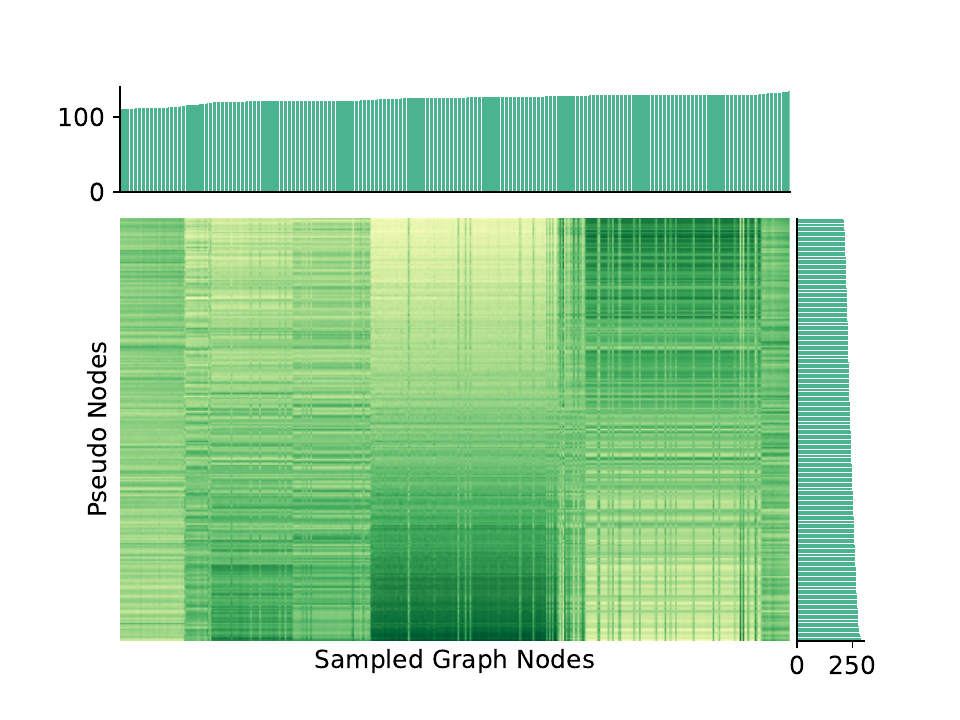}
              \end{minipage}}
        \vspace{-0.1in}
        \figcaption{\textbf{Message passing analysis.} The proximities between sampled graph nodes and pseudo nodes are depicted in the center of each sub-figure, where \textcolor[RGB]{0,96,51}{darker green} indicates higher proximity and \textcolor[RGB]{81,179,101}{brighter green} indicates lower proximity. The distribution on the top/right of each sub-figure denotes the sum of proximity for each graph/pseudo node. Graph/pseudo nodes are ranked based on the sum of proximity.}
        \label{fig:mp}
    \end{minipage}
    \hfill
    \begin{minipage}{0.45\linewidth}
\caption{\textbf{Ablation on the modules of \nsqure~(measured by accuracy: \%).} PA., L., and G. denote pseudo-node adaptation, local message passing, and global message passing. Att. and Prx. denote the attention and proximity measurement.}
\label{tab:ablation}
\begin{center}
\begin{small}
\resizebox{0.98\linewidth}{!}{
\begin{tabular}{llccc} 
\toprule
                      &       & \begin{tabular}[c]{@{}c@{}}Amz\\-ratings\end{tabular} & \begin{tabular}[c]{@{}c@{}}Amz\\Photo\end{tabular} & \begin{tabular}[c]{@{}c@{}}RPO\\-TEINS\end{tabular}  \\  
\midrule
\multirow{3}{*}{w/o.} & PA.   & 48.20$_{\color{gray}\pm0.28}$        & 95.10$_{\color{gray}\pm0.72}$     & 75.76$_{\color{gray}\pm2.33}$  \\
                      & L. & 48.69$_{\color{gray}\pm0.67}$        & 94.84$_{\color{gray}\pm0.59}$     & 75.77$_{\color{gray}\pm2.17}$  \\
                      & G. & 50.16$_{\color{gray}\pm0.54}$        & 94.58$_{\color{gray}\pm0.40}$     & 75.54$_{\color{gray}\pm1.63}$  \\
\multicolumn{2}{c}{Full with Att.}      & 49.18$_{\color{gray}\pm0.47}$                                       & 95.02$_{\color{gray}\pm0.20}$                                     & 73.68$_{\color{gray}\pm1.87}$                                       \\
\multicolumn{2}{c}{Full with Prx.}      & \textbf{50.25}$_{\color{gray}\pm0.53}$                                       & \textbf{95.75}$_{\color{gray}\pm0.34}$                                     & \textbf{77.53}$_{\color{gray}\pm1.78}$                                       \\
\bottomrule
\end{tabular}
}
\end{small}
\end{center}
    \end{minipage}
\vskip -0.1in
\end{figure*}

To further analyze the message passing between graph nodes and pseudo nodes, we visualize the proximity matrix $\mathrb{E}^{\mathtt{pn}}$ and $\mathrb{E}^{\mathtt{np}}$ for pseudo-node adaptation~(Eq.~\ref{eq:adaptation}) on AmazonPhoto in Fig.~\ref{fig:mp}. For more proximity visualizations, please refer to Appendix~\ref{ssec:app-mp}. $1000$ graph nodes are sampled randomly and ranked based on their intro-/outre-proximity summation. The intro-/outre-proximity summation indicates the message load a graph node collects or diffuses during the message passing. As depicted in Fig.~\ref{fig:mp}, both graph nodes and pseudo nodes assume a balanced message load. Each pseudo node has various proximity values toward individual graph nodes, different from the uniform pathways. All these properties empower efficient global message passing on graphs.

\subsubsection{Ablation Study}

\begin{wrapfigure}{r}{0.35\linewidth}
\centering
\vskip -0.3in
\begin{minipage}{\linewidth}
    \centering
    \includegraphics[width=\linewidth]{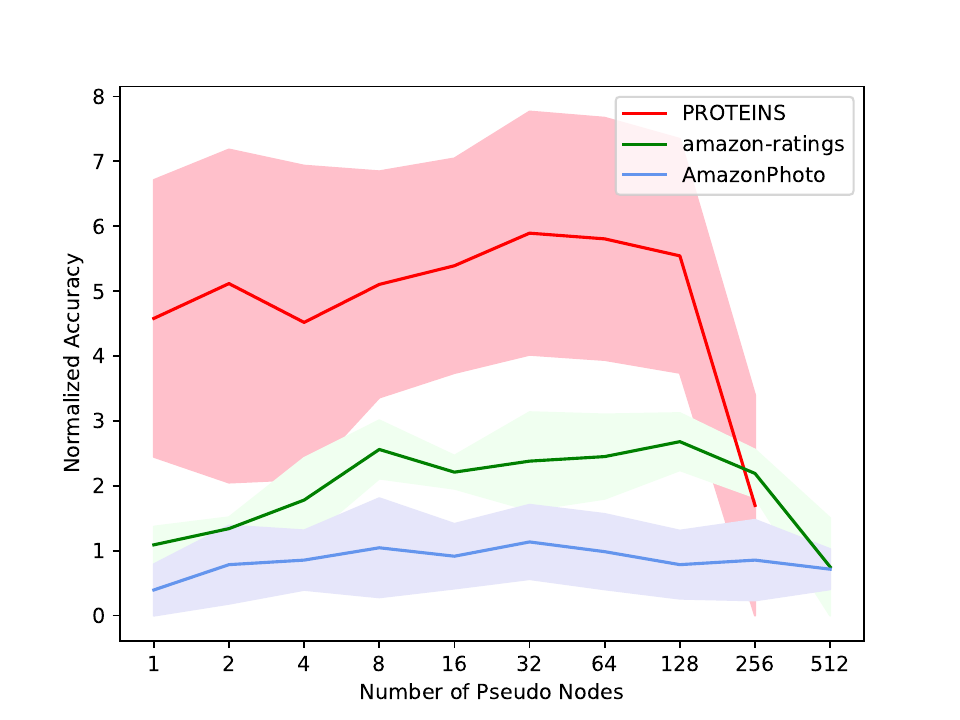}
    \figcaption{\textbf{Ablation on the number of pseudo nodes ($n_p$).}}
    \label{fig:num-np}
\end{minipage}

\vskip 0.2in

\begin{minipage}{\linewidth}
    \centering
    \includegraphics[width=\linewidth]{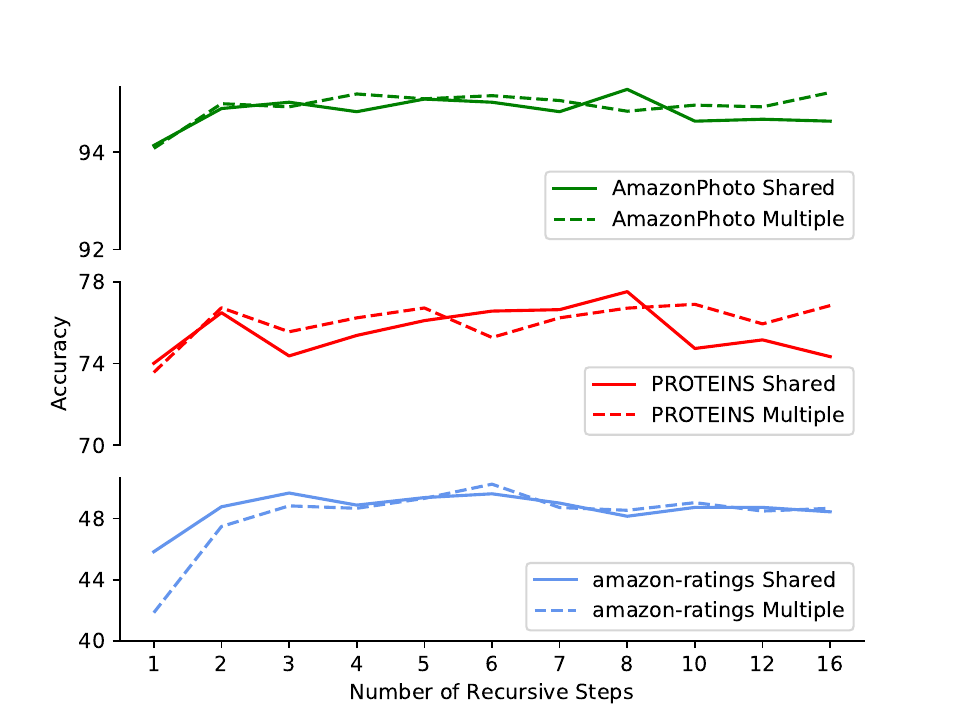}
    \figcaption{\textbf{Comparison between single shared recurrent layer and multiple recurrent layers.}}
    \label{fig:iter-deep}
\end{minipage}
\vskip -0.1in
\end{wrapfigure}
\textbf{Pseudo node.}~
We conduct ablation studies on the number of pseudo nodes $n_p$ in Fig.~\ref{fig:num-np}. To present the results in a comparable form within the same figure, the accuracy values have been normalized by subtracting the minimum value. As $n_p$ increases, \nsqure~gets improvement in accuracy for all three datasets. However, the accuracy exhibits degradation on PROTEINS when $n_p$ reaches 256. This is because $n_p$ exceeds the optimization capacity of \nsqure. A similar degradation also happens on amazon-ratings. In practice, the optimal value of $n_p$ is around 16 to 32 for graph classification while reaching 128 to 300 for node classification. We also conduct ablation studies on the engagement of pseudo nodes in Appendix~\ref{ssec:app-pseudo}, where \nsqure~with pseudo nodes outperforms dense message passing.

\textbf{Recurrent layer.}~
By employing the same recurrent layer through recursive steps, \nsqure~attains comparable performance to baseline models with substantially fewer parameters. To analyze the efficacy of parameter sharing, we conduct ablation studies on the number of recurrent layers, comparing \nsqure~with shared parameters against \nsqure~with multiple recurrent layers in Fig.~\ref{fig:iter-deep}. For simplicity, we denote the number of recurrent layers as $L_p$, where \nsqure~with multiple recurrent layers has $L_p$ equals to the number of recursive steps $L$ and \nsqure~with shared parameters has $L_p=1$. Given the same $L$, \nsqure~with $L_p=1$ achieves matching performance with $L_p=L$. This indicates that shared parameters are sufficient in modeling convergent dynamics of the embedded nodes in the state space (Fig.~\ref{fig:distribution}). However, \nsqure~achieves better performance with $L_p=L$ when $L$ surpasses $8$ on amazon-ratings and AmazonPhoto. 
In further empirical analysis, we find that the embedded nodes in \nsqure~with $L_p=1$ tend to maintain current dynamics through recursive steps and thus become less effective in precise position optimization. Please refer to Appendix~\ref{ssec:app-disp} for detailed analysis. A step-dependent parameter may further improve the performance of \nsqure~with $L_p=1$. We will keep exploring it in future work.

\textbf{Relation measurement.}~
We compare our proximity measurement with attention~\cite{vaswani_AttentionAllYou_2017} in Tab.~\ref{tab:ablation}. Results show that \nsqure~with proximity achieves better performance on all three benchmarks. We ascribe this to the flexible range of the proximity values. In contrast, normalized attention that ranges in $[0, 1]$ can yield equally small weights and attenuate the messages, especially when the optimal assignment involves a large number of graph nodes to the same pseudo node.

\textbf{\nsqure~modules.}~
In Tab.~\ref{tab:ablation}, \nsqure~with all three modules achieves superior performance across all the benchmarks. In comparison among the ablated \nsqure, removing global message passing leads to a larger degradation in graph classification accuracy. Conversely, node classification exhibits greater sensitivity to the removal of local message passing. Moreover, pseudo-node adaptation is required by \nsqure~on all three benchmarks. This indicates that adapting the randomly initialized distribution of pseudo nodes enables better interactions with graph nodes.

\section{Conclusion}
In this paper, we presented a dynamic message-passing method on graphs. 
Both graph nodes and pseudo nodes are embedded in a common state space with measurable relations between them. The measured relations serve as dynamic pathways between the embedded nodes, empowering flexible message passing. Associating pseudo nodes to input graphs with measured relations, graph nodes can communicate with each other intermediately through pseudo nodes under linear complexity. Based on the proposed dynamic message passing, we further developed a GNN model named \nsqure~for graph and node classification tasks. Experimental results show that \nsqure~achieves superior performance over competitive baseline models. For limitations discussion, please refer to Appendix~\ref{sec:app-limitation}.

\begin{ack}
This work was supported in part by the National Key R$\&$D Program of China under Grant 2023YFC2508704, in part by the National Natural Science Foundation of China: 62236008, U21B2038, and 61931008, and in part by the Fundamental Research Funds for the Central Universities. The authors would like to thank Wanxing Chang and the anonymous reviewers for their helpful comments and suggestions that improved this manuscript.
\end{ack}

{\small
\bibliographystyle{plain}
\bibliography{bibliography}

\begin{thebibliography}{10}

\bibitem{abu-el-haija_MixHopHigherOrderGraph_2019}
Sami {Abu-El-Haija}, Bryan Perozzi, Amol Kapoor, Nazanin Alipourfard, Kristina Lerman, Hrayr Harutyunyan, Greg~Ver Steeg, and Aram Galstyan.
\newblock {{MixHop}}: {{Higher-Order Graph Convolutional Architectures}} via {{Sparsified Neighborhood Mixing}}.
\newblock In {\em International {{Conference}} on {{Machine Learning}}}, pages 21--29, {Long Beach, USA}, 2019. {PMLR}.

\bibitem{alon_BottleneckGraphNeural_2021}
Uri Alon and Eran Yahav.
\newblock On the {{Bottleneck}} of {{Graph Neural Networks}} and its {{Practical Implications}}.
\newblock In {\em International {{Conference}} on {{Learning Representations}}}, {Virtual Only}, 2021.

\bibitem{baek_AccurateLearningGraph_2022}
Jinheon Baek, Minki Kang, and Sung~Ju Hwang.
\newblock Accurate {{Learning}} of {{Graph Representations}} with {{Graph Multiset Pooling}}.
\newblock In {\em International {{Conference}} on {{Learning Representations}}}, {virtual}, 2022.

\bibitem{Banach1922}
Stefan Banach.
\newblock Sur les opérations dans les ensembles abstraits et leur application aux équations intégrales.
\newblock {\em Fundamenta Mathematicae}, 3(1):133--181, 1922.

\bibitem{bo_LowfrequencyInformationGraph_2021}
Deyu Bo, Xiao Wang, Chuan Shi, and Huawei Shen.
\newblock Beyond {{Low-frequency Information}} in {{Graph Convolutional Networks}}.
\newblock In {\em {{AAAI Conference}} on {{Artificial Intelligence}}}, volume~35, pages 3950--3957, {virtual}, 2021.

\bibitem{bresson_ResidualGatedGraph_2018}
Xavier Bresson and Thomas Laurent.
\newblock Residual {{Gated Graph ConvNets}}, 2018.

\bibitem{cai_ConnectionMPNNGraph_2023}
Chen Cai, Truong~Son Hy, Rose Yu, and Yusu Wang.
\newblock On the {{Connection Between MPNN}} and {{Graph Transformer}}.
\newblock In {\em Proceedings of the 40th {{International Conference}} on {{Machine Learning}}}, pages 3408--3430. {PMLR}, 2023.

\bibitem{cai_NoteOverSmoothingGraph_2020}
Chen Cai and Yusu Wang.
\newblock A {{Note}} on {{Over-Smoothing}} for {{Graph Neural Networks}}.
\newblock {\em arXiv:2006.13318 [cs, stat]}, 2020.

\bibitem{chen_NAGphormerTokenizedGraph_2023}
Jinsong Chen, Kaiyuan Gao, Gaichao Li, and Kun He.
\newblock {{NAGphormer}}: {{A Tokenized Graph Transformer}} for {{Node Classification}} in {{Large Graphs}}.
\newblock In {\em International {{Conference}} for {{Learning Representations}}}, 2023.

\bibitem{chen_SimpleDeepGraph_2020}
Ming Chen, Zhewei Wei, Zengfeng Huang, Bolin Ding, and Yaliang Li.
\newblock Simple and {{Deep Graph Convolutional Networks}}.
\newblock In {\em International {{Conference}} on {{Machine Learning}}}, pages 1725--1735, {virtual}, 2020. {PMLR}.

\bibitem{chien_AdaptiveUniversalGeneralized_2022}
Eli Chien, Jianhao Peng, Pan Li, and Olgica Milenkovic.
\newblock Adaptive {{Universal Generalized PageRank Graph Neural Network}}.
\newblock In {\em International {{Conference}} on {{Learning Representations}}}, {virtual}, 2022.

\bibitem{deac_ExpanderGraphPropagation_2022}
Andreea Deac, Marc Lackenby, and Petar Veli{\v c}kovi{\'c}.
\newblock Expander {{Graph Propagation}}.
\newblock In {\em {{Learning}} on {{Graphs Conference}}}, volume 198, page~38. {PMLR}, 2022.

\bibitem{defferrard_ConvolutionalNeuralNetworks_2017}
Micha{\"e}l Defferrard, Xavier Bresson, and Pierre Vandergheynst.
\newblock Convolutional {{Neural Networks}} on {{Graphs}} with {{Fast Localized Spectral Filtering}}.
\newblock {\em arXiv:1606.09375 [cs, stat]}, 2017.

\bibitem{digiovanni_OverSquashingMessagePassing_2023}
Francesco Di~Giovanni, Lorenzo Giusti, Federico Barbero, Giulia Luise, Pietro Lio', and Michael Bronstein.
\newblock On {{Over-Squashing}} in {{Message Passing Neural Networks}}: {{The Impact}} of {{Width}}, {{Depth}}, and {{Topology}}.
\newblock In {\em International Conference on Machine Learning}, 2023.

\bibitem{giovanni_HowDoesOversquashing_2023}
Francesco Di~Giovanni, T.~Konstantin Rusch, Michael Bronstein, Andreea Deac, Marc Lackenby, Siddhartha Mishra, and Petar Veli{\v c}kovi{\'c}.
\newblock How does over-squashing affect the power of {{GNNs}}?
\newblock {\em Transactions on Machine Learning Research}, 2023.

\bibitem{diao_RelationalAttentionGeneralizing_2023a}
Cameron Diao and Ricky Loynd.
\newblock Relational {{Attention}}: {{Generalizing Transformers}} for {{Graph-Structured Tasks}}.
\newblock In {\em {{International Conference}} on {{Learning Representations}}}, 2023.

\bibitem{dwivedi_LongRangeGraph_2022}
Vijay~Prakash Dwivedi, Ladislav Ramp{\'a}{\v s}ek, Michael Galkin, Ali Parviz, Guy Wolf, Anh~Tuan Luu, and Dominique Beaini.
\newblock Long {{Range Graph Benchmark}}.
\newblock In {\em Advances in {{Neural Information Processing Systems}}}, volume~35, pages 22326--22340, 2022.

\bibitem{fey_FastGraphRepresentation_2019}
Matthias Fey and Jan~Eric Lenssen.
\newblock Fast {{Graph Representation Learning}} with {{PyTorch Geometric}}.
\newblock In {\em International {{Conference}} on {{Learning Representations Workshop}} on {{Graphs}} and {{Manifolds}}}, 2019.

\bibitem{franceschi_LearningDiscreteStructures_2019}
Luca Franceschi, Mathias Niepert, Massimiliano Pontil, and Xiao He.
\newblock Learning {{Discrete Structures}} for {{Graph Neural Networks}}.
\newblock In {\em International {{Conference}} on {{Machine Learning}}}, pages 1972--1982. {PMLR}, 2019.

\bibitem{gao_GraphUNets_2019}
Hongyang Gao and Shuiwang Ji.
\newblock Graph {{U-Nets}}.
\newblock In {\em International {{Conference}} on {{Machine Learning}}}, pages 2083--2092, {Long Beach, California, USA}, 2019. {PMLR}.

\bibitem{gasteiger_PredictThenPropagate_2018}
Johannes Gasteiger, Aleksandar Bojchevski, and Stephan G{\"u}nnemann.
\newblock Predict then {{Propagate}}: {{Graph Neural Networks}} meet {{Personalized PageRank}}.
\newblock In {\em International {{Conference}} on {{Learning Representations}}}, {Addis Ababa, Ethiopia}, 2018.

\bibitem{gilmer_NeuralMessagePassing_2017}
Justin Gilmer, Samuel~S. Schoenholz, Patrick~F. Riley, Oriol Vinyals, and George~E. Dahl.
\newblock Neural {{Message Passing}} for {{Quantum Chemistry}}.
\newblock In {\em International {{Conference}} on {{Machine Learning}}}, volume~70, pages 1263--1272, {Sydney, Australia}, 2017. {PMLR}.

\bibitem{gu_EfficientlyModelingLong_2021}
Albert Gu, Karan Goel, and Christopher Re.
\newblock Efficiently {{Modeling Long Sequences}} with {{Structured State Spaces}}.
\newblock In {\em International {{Conference}} on {{Learning Representations}}}, 2021.

\bibitem{gutteridge_DRewDynamicallyRewired_2023}
Benjamin Gutteridge, Xiaowen Dong, Michael~M. Bronstein, and Francesco~Di Giovanni.
\newblock {{DRew}}: {{Dynamically Rewired Message Passing}} with {{Delay}}.
\newblock In {\em International {{Conference}} on {{Machine Learning}}}, pages 12252--12267. {PMLR}, 2023.

\bibitem{hamilton_InductiveRepresentationLearning_2017}
William~L. Hamilton, Rex Ying, and Jure Leskovec.
\newblock Inductive representation learning on large graphs.
\newblock In {\em International {{Conference}} on {{Neural Information Processing Systems}}}, pages 1025--1035, {Red Hook, USA}, 2017. {Curran Associates Inc.}

\bibitem{hochreiter_LongShortTermMemory_1997}
Sepp Hochreiter and J{\"u}rgen Schmidhuber.
\newblock Long {{Short-Term Memory}}.
\newblock {\em Neural Computation}, 9(8):1735--1780, 1997.

\bibitem{hu_OpenGraphBenchmark_2020}
Weihua Hu, Matthias Fey, Marinka Zitnik, Yuxiao Dong, Hongyu Ren, Bowen Liu, Michele Catasta, and Jure Leskovec.
\newblock Open {{Graph Benchmark}}: {{Datasets}} for {{Machine Learning}} on {{Graphs}}.
\newblock In {\em Advances in {{Neural Information Processing Systems}}}, volume~33, pages 22118--22133. {Curran Associates, Inc.}, 2020.

\bibitem{hu*_StrategiesPretrainingGraph_2019}
Weihua Hu*, Bowen Liu*, Joseph Gomes, Marinka Zitnik, Percy Liang, Vijay Pande, and Jure Leskovec.
\newblock Strategies for {{Pre-training Graph Neural Networks}}.
\newblock In {\em International {{Conference}} on {{Learning Representations}}}, 2019.

\bibitem{huang_GoingDeeperPermutationSensitive_2022}
Zhongyu Huang, Yingheng Wang, Chaozhuo Li, and Huiguang He.
\newblock Going {{Deeper}} into {{Permutation-Sensitive Graph Neural Networks}}.
\newblock In {\em International {{Conference}} on {{Machine Learning}}}, pages 9377--9409, {Baltimore, Maryland, USA}, 2022. {PMLR}.

\bibitem{hwang_RevisitingVirtualNodes_2021}
EunJeong Hwang, Veronika Thost, Shib~Sankar Dasgupta, and Tengfei Ma.
\newblock Revisiting {{Virtual Nodes}} in {{Graph Neural Networks}} for {{Link Prediction}}.
\newblock 2021.

\bibitem{kazi_DifferentiableGraphModule_2023}
Anees Kazi, Luca Cosmo, Seyed-Ahmad Ahmadi, Nassir Navab, and Michael~M. Bronstein.
\newblock Differentiable {{Graph Module}} ({{DGM}}) for {{Graph Convolutional Networks}}.
\newblock {\em IEEE Transactions on Pattern Analysis and Machine Intelligence}, 45(2):1606--1617, 2023.

\bibitem{kingma_AdamMethodStochastic_2015}
Diederik~P. Kingma and Jimmy Ba.
\newblock Adam: {{A Method}} for {{Stochastic Optimization}}.
\newblock In {\em International {{Conference}} for {{Learning Representations}}}, {San Diego, USA}, 2015.

\bibitem{kipf_SemiSupervisedClassificationGraph_2017}
Thomas~N. Kipf and Max Welling.
\newblock Semi-{{Supervised Classification}} with {{Graph Convolutional Networks}}.
\newblock In {\em International {{Conference}} on {{Learning Representations}}}, {Toulon, France}, 2017.

\bibitem{klicpera_DiffusionImprovesGraph_2019}
Johannes Klicpera, Stefan {Wei{\ss} enberger}, and Stephan G{\"u}nnemann.
\newblock Diffusion {{Improves Graph Learning}}.
\newblock In {\em International {{Conference}} on {{Advances}} in {{Neural Information Processing Systems}}}, volume~32, pages 13333--13345, {Vancouver, Canada}, 2019. {Curran Associates, Inc.}

\bibitem{kreuzer_RethinkingGraphTransformers_2021}
Devin Kreuzer, Dominique Beaini, Will Hamilton, Vincent L{\'e}tourneau, and Prudencio Tossou.
\newblock Rethinking {{Graph Transformers}} with {{Spectral Attention}}.
\newblock In {\em Advances in {{Neural Information Processing Systems}}}, volume~34, pages 21618--21629. {Curran Associates, Inc.}, 2021.

\bibitem{krizhevsky_ImageNetClassificationDeep_2017}
Alex Krizhevsky, Ilya Sutskever, and Geoffrey~E. Hinton.
\newblock {{ImageNet}} classification with deep convolutional neural networks.
\newblock {\em Communications of the ACM}, 60(6):84--90, 2017.

\bibitem{lee_SelfAttentionGraphPooling_2019}
Junhyun Lee, Inyeop Lee, and Jaewoo Kang.
\newblock Self-{{Attention Graph Pooling}}.
\newblock In {\em International {{Conference}} on {{Machine Learning}}}, pages 3734--3743, {Long Beach, California, USA}, 2019. {PMLR}.

\bibitem{snapnets}
Jure Leskovec and Andrej Krevl.
\newblock {SNAP Datasets}: {Stanford} large network dataset collection.
\newblock \url{http://snap.stanford.edu/data}, 2014.

\bibitem{li_DeeperInsightsGraph_2018}
Qimai Li, Zhichao Han, and Xiao-Ming Wu.
\newblock Deeper {{Insights}} into {{Graph Convolutional Networks}} for {{Semi-Supervised Learning}}.
\newblock In {\em {{AAAI Conference}} on {{Artificial Intelligence}}}, {New Orleans, USA}, 2018.

\bibitem{li_FindingGlobalHomophily_2022}
Xiang Li, Renyu Zhu, Yao Cheng, Caihua Shan, Siqiang Luo, Dongsheng Li, and Weining Qian.
\newblock Finding {{Global Homophily}} in {{Graph Neural Networks When Meeting Heterophily}}.
\newblock In {\em International {{Conference}} on {{Machine Learning}}}, volume 162, pages 13242--13256, {Baltimore, Maryland, USA}, 2022. {PMLR}.

\bibitem{lim_LargeScaleLearning_2021}
Derek Lim, Felix Hohne, Xiuyu Li, Sijia~Linda Huang, Vaishnavi Gupta, Omkar Bhalerao, and Ser~Nam Lim.
\newblock Large {{Scale Learning}} on {{Non-Homophilous Graphs}}: {{New Benchmarks}} and {{Strong Simple Methods}}.
\newblock In {\em Advances in {{Neural Information Processing Systems}}}, volume~34, pages 20887--20902. {Curran Associates, Inc.}, 2021.

\bibitem{liu_BoostingGraphStructure_2022}
Xin Liu, Jiayang Cheng, Yangqiu Song, and Xin Jiang.
\newblock Boosting {{Graph Structure Learning}} with {{Dummy Nodes}}.
\newblock In {\em {{International Conference}} on {{Machine Learning}}}, pages 13704--13716. {PMLR}, 2022.

\bibitem{mialon_GraphiTEncodingGraph_2021}
Gr{\'e}goire Mialon, Dexiong Chen, Margot Selosse, and Julien Mairal.
\newblock {{GraphiT}}: {{Encoding Graph Structure}} in {{Transformers}}.
\newblock {\em arXiv preprint}.

\bibitem{morris_TUDatasetCollectionBenchmark_2020}
Christopher Morris, Nils~M. Kriege, Franka Bause, Kristian Kersting, Petra Mutzel, and Marion Neumann.
\newblock {{TUDataset}}: {{A}} collection of benchmark datasets for learning with graphs.
\newblock In {\em International {{Conference}} on {{Machine Learning Workshop}} on {{Graph Representation Learning}} and {{Beyond}}}, {Virtual Only}, 2020. {arXiv}.

\bibitem{niepert_LearningConvolutionalNeural_2016}
Mathias Niepert, Mohamed Ahmed, and Konstantin Kutzkov.
\newblock Learning convolutional neural networks for graphs.
\newblock In {\em International {{Conference}} on {{Machine Learning}}}, volume~48, pages 2014--2023, {New York, NY, USA}, 2016. {PMLR}.

\bibitem{oono_GraphNeuralNetworks_2021}
Kenta Oono and Taiji Suzuki.
\newblock Graph {{Neural Networks Exponentially Lose Expressive Power}} for {{Node Classification}}.
\newblock In {\em International {{Conference}} for {{Learning Representations}}}, 2021.

\bibitem{paszke_PyTorchImperativeStyle_2019}
Adam Paszke, Sam Gross, Francisco Massa, Adam Lerer, James Bradbury, Gregory Chanan, Trevor Killeen, Zeming Lin, Natalia Gimelshein, Luca Antiga, Alban Desmaison, Andreas K{\"o}pf, Edward Yang, Zach DeVito, Martin Raison, Alykhan Tejani, Sasank Chilamkurthy, Benoit Steiner, Lu~Fang, Junjie Bai, and Soumith Chintala.
\newblock {{PyTorch}}: An imperative style, high-performance deep learning library.
\newblock In {\em International {{Conference}} on {{Neural Information Processing Systems}}}, pages 8026--8037, {Red Hook, NY, USA}, 2019. {Curran Associates Inc.}

\bibitem{pei_DynamicsinspiredNeuromorphicVisual_2023}
Zhengqi Pei and Shuhui Wang.
\newblock Dynamics-inspired {{Neuromorphic Visual Representation Learning}}.
\newblock In {\em International {{Conference}} on {{Machine Learning}}}, pages 27521--27541. {PMLR}, 2023.

\bibitem{platonov_CriticalLookEvaluation_2023}
Oleg Platonov, Denis Kuznedelev, Michael Diskin, Artem Babenko, and Liudmila Prokhorenkova.
\newblock A critical look at the evaluation of {{GNNs}} under heterophily: {{Are}} we really making progress?
\newblock In {\em The {{Eleventh International Conference}} on {{Learning Representations}}}, {Kigali, Rwanda}, 2023.

\bibitem{qian_ProbabilisticallyRewiredMessagePassing_2023}
Chendi Qian, Andrei Manolache, Kareem Ahmed, Zhe Zeng, Guy~Van den Broeck, Mathias Niepert, and Christopher Morris.
\newblock Probabilistically {{Rewired Message-Passing Neural Networks}}.
\newblock In {\em {{International Conference}} on {{Learning Representations}}}, 2023.

\bibitem{rahmani_GraphNeuralNetworks_2023}
Saeed Rahmani, Asiye Baghbani, Nizar Bouguila, and Zachary Patterson.
\newblock Graph {{Neural Networks}} for {{Intelligent Transportation Systems}}: {{A Survey}}.
\newblock {\em IEEE Transactions on Intelligent Transportation Systems}, 24(8):8846--8885, 2023.

\bibitem{rampasek_RecipeGeneralPowerful_2022}
Ladislav Ramp{\'a}{\v s}ek, Michael Galkin, Vijay~Prakash Dwivedi, Anh~Tuan Luu, Guy Wolf, and Dominique Beaini.
\newblock Recipe for a {{General}}, {{Powerful}}, {{Scalable Graph Transformer}}.
\newblock In {\em Advances in {{Neural Information Processing Systems}}}, volume~35, pages 14501--14515, {New Orleans, USA}, 2022.

\bibitem{ranjan_ASAPAdaptiveStructure_2020}
Ekagra Ranjan, Soumya Sanyal, and Partha Talukdar.
\newblock {{ASAP}}: {{Adaptive Structure Aware Pooling}} for {{Learning Hierarchical Graph Representations}}.
\newblock In {\em {{AAAI Conference}} on {{Artificial Intelligence}}}, volume~34, pages 5470--5477, {New York, NY, USA}, 2020.

\bibitem{scarselli_GraphNeuralNetwork_2009}
Franco Scarselli, Marco Gori, Ah~Chung Tsoi, Markus Hagenbuchner, and Gabriele Monfardini.
\newblock The {{Graph Neural Network Model}}.
\newblock {\em IEEE Transactions on Neural Networks}, 20(1):61--80, 2009.

\bibitem{shchur_PitfallsGraphNeural_2019}
Oleksandr Shchur, Maximilian Mumme, Aleksandar Bojchevski, and Stephan G{\"u}nnemann.
\newblock Pitfalls of {{Graph Neural Network Evaluation}}.
\newblock {\em arXiv:1811.05868 [cs, stat]}, 2019.

\bibitem{shi_MaskedLabelPrediction_2021}
Yunsheng Shi, Zhengjie Huang, Shikun Feng, Hui Zhong, Wenjing Wang, and Yu~Sun.
\newblock Masked {{Label Prediction}}: {{Unified Message Passing Model}} for {{Semi-Supervised Classification}}.
\newblock In {\em {{International Joint Conference}} on {{Artificial Intelligence}}}, volume~2, pages 1548--1554, {Virtual Event / Montreal, Canada}, 2021. {ijcai.org}.

\bibitem{shirzad_ExphormerSparseTransformers_2023}
Hamed Shirzad, Ameya Velingker, Balaji Venkatachalam, Danica~J. Sutherland, and Ali~Kemal Sinop.
\newblock Exphormer: {{Sparse Transformers}} for {{Graphs}}.
\newblock In {\em International {{Conference}} on {{Machine Learning}}}, volume 202, {Honolulu, USA}, 2023. {PMLR}.

\bibitem{sun_AllinARow_2023}
Junshu Sun, Shuhui Wang, Xinzhe Han, Zhe Xue, and Qingming Huang.
\newblock All in a row: {{Compressed}} convolution networks for graphs.
\newblock In {\em International {{Conference}} on {{Machine Learning}}}, volume 202, pages 33061--33076, {Honolulu, USA}, 2023. {PMLR}.

\bibitem{sun_HomophilyStructureawarePath_2022}
Yifei Sun, Haoran Deng, Yang Yang, Chunping Wang, Jiarong Xu, Renhong Huang, Linfeng Cao, Yang Wang, and Lei Chen.
\newblock Beyond {{Homophily}}: {{Structure-aware Path Aggregation Graph Neural Network}}.
\newblock In {\em {{International Joint Conference}} on {{Artificial Intelligence}}}, volume~3, pages 2233--2240, 2022.

\bibitem{topping_UnderstandingOversquashingBottlenecks_2021}
Jake Topping, Francesco Di~Giovanni, Benjamin~Paul Chamberlain, Xiaowen Dong, and Michael~M. Bronstein.
\newblock Understanding over-squashing and bottlenecks on graphs via curvature.
\newblock In {\em International {{Conference}} for {{Learning Representations}}}, 2021.

\bibitem{t-SNE}
Laurens Van~der Maaten and Geoffrey Hinton.
\newblock Visualizing data using t-sne.
\newblock {\em Journal of machine learning research}, 9(11), 2008.

\bibitem{vaswani_AttentionAllYou_2017}
Ashish Vaswani, Noam Shazeer, Niki Parmar, Jakob Uszkoreit, Llion Jones, Aidan~N. Gomez, {\L}ukasz Kaiser, and Illia Polosukhin.
\newblock Attention is all you need.
\newblock In {\em International {{Conference}} on {{Neural Information Processing Systems}}}, pages 6000--6010, {Red Hook, NY, USA}, 2017. {Curran Associates Inc.}

\bibitem{velickovic_MessagePassingAll_2022}
Petar Veli{\v c}kovi{\'c}.
\newblock Message passing all the way up.
\newblock In {\em International Conference on Learning Representation Workshop on Geometrical and Topological Representation Learning}, 2022.

\bibitem{velickovic_GraphAttentionNetworks_2018}
Petar Veli{\v c}kovi{\'c}, Guillem Cucurull, Arantxa Casanova, Adriana Romero, Pietro Li{\`o}, and Yoshua Bengio.
\newblock Graph {{Attention Networks}}.
\newblock In {\em International {{Conference}} on {{Learning Representations}}}, {Vancouver, Canada}, 2018.

\bibitem{wang_PROSEGraphStructure_2023}
Huizhao Wang, Yao Fu, Tao Yu, Linghui Hu, Weihao Jiang, and Shiliang Pu.
\newblock {{PROSE}}: {{Graph Structure Learning}} via {{Progressive Strategy}}.
\newblock In {\em {{ACM SIGKDD Conference}} on {{Knowledge Discovery}} and {{Data Mining}}}, pages 2337--2348, New York, NY, USA, 2023. Association for Computing Machinery.

\bibitem{wu_SimplifyingGraphConvolutional_2019}
Felix Wu, Amauri Souza, Tianyi Zhang, Christopher Fifty, Tao Yu, and Kilian Weinberger.
\newblock Simplifying {{Graph Convolutional Networks}}.
\newblock In {\em International {{Conference}} on {{Machine Learning}}}, pages 6861--6871, {Long Beach, USA}, 2019. {PMLR}.

\bibitem{wu_StructuralEntropyGuided_2022}
Junran Wu, Xueyuan Chen, Shangzhe Li, and Ke~Xu.
\newblock Structural entropy guided graph hierarchical pooling.
\newblock In {\em International Conference on Machine Learning}, volume 162, pages 24017--24030, {Baltimore, Maryland, USA}, 2022. {PMLR}.

\bibitem{wu_DIFFormerScalableGraph_2023}
Qitian Wu, Chenxiao Yang, Wentao Zhao, Yixuan He, David Wipf, and Junchi Yan.
\newblock {{DIFFormer}}: {{Scalable}} ({{Graph}}) {{Transformers Induced}} by {{Energy Constrained Diffusion}}.
\newblock In {\em The {{Eleventh International Conference}} on {{Learning Representations}}}, 2023.

\bibitem{wu_NodeFormerScalableGraph_2022}
Qitian Wu, Wentao Zhao, Zenan Li, David Wipf, and Junchi Yan.
\newblock {{NodeFormer}}: {{A Scalable Graph Structure Learning Transformer}} for {{Node Classification}}.
\newblock In {\em Advances in {{Neural Information Processing Systems}}}, 2022.

\bibitem{wu_SGFormerSimplifyingEmpowering_2023}
Qitian Wu, Wentao Zhao, Chenxiao Yang, Hengrui Zhang, Fan Nie, Haitian Jiang, Yatao Bian, and Junchi Yan.
\newblock {{SGFormer}}: {{Simplifying}} and {{Empowering Transformers}} for {{Large-Graph Representations}}.
\newblock In {\em {{Conference}} on {{Neural Information Processing Systems}}}, 2023.

\bibitem{xu*_HowPowerfulAre_2019}
Keyulu Xu, Weihua Hu, Jure Leskovec, and Stefanie Jegelka.
\newblock How {{Powerful}} are {{Graph Neural Networks}}?
\newblock In {\em International {{Conference}} on {{Learning Representations}}}, {New Orleans, LA, USA}, 2019.

\bibitem{ying_TransformersReallyPerform_2021}
Chengxuan Ying, Tianle Cai, Shengjie Luo, Shuxin Zheng, Guolin Ke, Di~He, Yanming Shen, and Tie-Yan Liu.
\newblock Do {{Transformers Really Perform Bad}} for {{Graph Representation}}?
\newblock In {\em Advances in {{Neural Information Processing Systems}}}, volume~34, pages 28877--28888. {Curran Associates, Inc.}, 2021.

\bibitem{ying_HierarchicalGraphRepresentation_2018}
Zhitao Ying, Jiaxuan You, Christopher Morris, Xiang Ren, Will Hamilton, and Jure Leskovec.
\newblock Hierarchical {{Graph Representation Learning}} with {{Differentiable Pooling}}.
\newblock In {\em Advances in {{Neural Information Processing Systems}}}, volume~31, pages 4805--4815, {Montr\'eal, Canada}, 2018. {Curran Associates, Inc.}

\bibitem{yuan_StructPoolStructuredGraph_2020}
Hao Yuan and Shuiwang Ji.
\newblock {{StructPool}}: {{Structured Graph Pooling}} via {{Conditional Random Fields}}.
\newblock In {\em International {{Conference}} on {{Learning Representations}}}, {Addis Ababa, Ethiopia}, 2020.

\bibitem{zaidi_PretrainingDenoisingMolecular_2023}
Sheheryar Zaidi, Michael Schaarschmidt, James Martens, Hyunjik Kim, Yee~Whye Teh, Alvaro {Sanchez-Gonzalez}, Peter Battaglia, Razvan Pascanu, and Jonathan Godwin.
\newblock Pre-training via {{Denoising}} for {{Molecular Property Prediction}}.
\newblock In {\em The {{Eleventh International Conference}} on {{Learning Representations}}}, 2023.

\bibitem{zhao_GraphGLOWUniversalGeneralizable_2023}
Wentao Zhao, Qitian Wu, Chenxiao Yang, and Junchi Yan.
\newblock {{GraphGLOW}}: {{Universal}} and {{Generalizable Structure Learning}} for {{Graph Neural Networks}}.
\newblock In {\em Proceedings of the 29th {{ACM SIGKDD Conference}} on {{Knowledge Discovery}} and {{Data Mining}}}, pages 3525--3536, New York, NY, USA, 2023. Association for Computing Machinery.

\bibitem{zhou_GraphNeuralNetworks_2020}
Jie Zhou, Ganqu Cui, Shengding Hu, Zhengyan Zhang, Cheng Yang, Zhiyuan Liu, Lifeng Wang, Changcheng Li, and Maosong Sun.
\newblock Graph neural networks: {{A}} review of methods and applications.
\newblock {\em AI Open}, 1:57--81, 2020.

\bibitem{zhu_HomophilyGraphNeural_2020}
Jiong Zhu, Yujun Yan, Lingxiao Zhao, Mark Heimann, Leman Akoglu, and Danai Koutra.
\newblock Beyond homophily in graph neural networks: Current limitations and effective designs.
\newblock In {\em {{International Conference}} on {{Neural Information Processing Systems}}}, pages 7793--7804, {Red Hook, NY, USA}, 2020. {Curran Associates Inc.}

\bibitem{zhu_DeepGraphStructure_2021}
Yanqiao Zhu, Weizhi Xu, Jinghao Zhang, Qiang Liu, Shu Wu, and Liang Wang.
\newblock Deep {{Graph Structure Learning}} for {{Robust Representations}}: {{A Survey}}.
\newblock {\em arXiv:2103.03036 [cs]}, 2021.

\end{thebibliography}
}


\newpage
\appendix
\renewcommand{\thefigure}{S\arabic{figure}}
\renewcommand{\thetable}{S\arabic{table}}
\renewcommand{\theequation}{S\arabic{equation}}

\section{State Space}\label{sec:state-space}
\nsqure~unifies graph nodes and pseudo nodes in a common state space. The concept of state space has been widely adopted in various machine learning methods. In reinforcement learning, the state space is typically modeled as a discrete space, encompassing all possible states of an agent. By taking discrete actions, agents make transitions between states. Through interactions with the environment, agents learn optimal policies that determine their actions at certain states, maximizing the cumulative reward over time.

In contrast to the discrete paradigm, recurrent models such as long short-term memory~(LSTM)~\cite{hochreiter_LongShortTermMemory_1997}, and state space models~(SSMs)~\cite{gu_EfficientlyModelingLong_2021} utilize continuous state representations to model sequential data. In this setting, input sequences are tokenized and consumed by the model in a successive manner. As a result, recurrent models learn to update the state embeddings recursively.

Building upon the continuous paradigm, DyN~\cite{pei_DynamicsinspiredNeuromorphicVisual_2023} further applies measurements to the continuous state space, giving rise to measurable spatial relations between state embeddings. All the input tokens can now be modeled simultaneously in the state space.

However, modeling dense pairwise relations among a large number of graph nodes becomes intractable in the graph representation learning setting, due to the quadratic complexity. To reduce the complexity, we introduce pseudo nodes to serve as proxies for pairwise relations between graph nodes. Consequently, the relations between graph nodes are decomposed into two components, {\it i.e.}, relations between source graph nodes and pseudo nodes, and relations between pseudo nodes and target graph nodes. Since the number of pseudo nodes is substantially smaller than the number of graph nodes, this decomposition effectively reduces the overall complexity.

\section{Pseudo Nodes and Pooling Nodes}
In hierarchical GNNs, graph nodes are compressed into higher-level nodes through iterative graph pooling. These higher-level nodes, referred to as pooling nodes, are different from the pseudo nodes in \nsqure. First, pseudo nodes and pooling nodes are learned for different objectives. Pseudo nodes optimize communication efficiency between graph nodes during global message passing. Pooling nodes capture the hierarchy in the graph structures, ensuring better structure compression. Second, pooling nodes and pseudo nodes have different relations with input graph nodes. Pooling nodes are higher-level abstractions of graph nodes and are not physically connected with them. In contrast, pseudo nodes can be regarded as learnable graph nodes. These nodes are physically connected to graph nodes as part of the graphs and directly participate in the message passing between graph nodes. Third, pseudo nodes and pooling nodes employ different communication pathways. Pooling nodes employ the coarse adjacency matrix for message passing. Pseudo nodes are free from structure-preserving constraints and learn fully connected pathways to communicate with each other.

\section{Details on Experiments}\label{sec:app-exp-setup}
We have performed grid search for the hyper-parameters in Tab.~\ref{tab:app-hyp} and the reproduced models based on validation loss. All the learnable parameters in \nsqure~are optimized during training, including the weights in the linear transformations and the proximity measurement, the pseudo/class-node states. The cross-entropy loss is adopted for classification and the L1 loss for regression. No position or structural encoding methods are employed for \nsqure.
 
\subsection{Graph Classification}
\subsubsection{Benchmark Descriptions}
We adopt six benchmarks for graph classification, including three biochemical datasets (OGB-molpcba~\cite{hu_OpenGraphBenchmark_2020}, PROTEINS~\cite{morris_TUDatasetCollectionBenchmark_2020}, NCI1~\cite{morris_TUDatasetCollectionBenchmark_2020}) and three social network datasets~\cite{morris_TUDatasetCollectionBenchmark_2020} (COLLAB, IMDB-BINARY, and IMDB-MULTI). 

\textbf{OGB-molpcba} is a large-scale molecular property prediction benchmark. Each graph is a discrete molecule, wherein nodes denote individual atoms and edges encode chemical bonds between atoms.

\textbf{PROTEINS} comprises 1,113 protein graphs with amino acids constituting the nodes. The associated binary classification task involves predicting protein category labels, specifically discriminating between enzymes versus non-enzymes.

\textbf{NCI1} constitutes 4,110 graphs of chemical compounds assembled by the National Cancer Institute (NCI). Graph labels categorize compounds as exhibiting either positive or negative efficacy against cell lung cancer.

\textbf{IMDB-BINARY} and \textbf{IMDB-MULTI} constitute collaboration network graphs of movie actors and actresses, with graph labels indicating movie genres.
 
\textbf{COLLAB} constitutes scientific collaboration networks wherein each graph is an ego network for a researcher, encompassing their co-authors. Graph labels indicate the specific scientific interest corresponding to each researcher.

\subsubsection{Experimental Setups}\label{ssec:graph-setup}
We choose (\textbf{1}) Convolutional GNNs: GCN~\cite{kipf_SemiSupervisedClassificationGraph_2017}, PATCHY-SAN~\cite{niepert_LearningConvolutionalNeural_2016},  GraphSAGE~\cite{hamilton_InductiveRepresentationLearning_2017}, GIN~\cite{xu*_HowPowerfulAre_2019}, PG~\cite{huang_GoingDeeperPermutationSensitive_2022}, and CoCN~\cite{sun_AllinARow_2023}; (\textbf{2}) GNNs with a single pseudo node from~\cite{liu_BoostingGraphStructure_2022,hu_OpenGraphBenchmark_2020,cai_ConnectionMPNNGraph_2023}; (\textbf{3}) Hierarchical GNNs: DiffPool~\cite{ying_HierarchicalGraphRepresentation_2018}, TopKPool~\cite{gao_GraphUNets_2019}, SAGPool~\cite{lee_SelfAttentionGraphPooling_2019}, 
StructPool~\cite{yuan_StructPoolStructuredGraph_2020}, SEP~\cite{wu_StructuralEntropyGuided_2022}, and GMT~\cite{baek_AccurateLearningGraph_2022}; (\textbf{4}) Graph transformers: Graphormer~\cite{ying_TransformersReallyPerform_2021}, SAN~\cite{kreuzer_RethinkingGraphTransformers_2021}, GraphGPS~\cite{rampasek_RecipeGeneralPowerful_2022}, and GraphTrans~\cite{cai_ConnectionMPNNGraph_2023} as the baselines of graph classification.

Except for OGB-molpcba, we perform 10-fold cross-validation with LIB-SVM following~\cite{xu*_HowPowerfulAre_2019} and report average performance. Average precision is reported for OGB-molpcba while accuracy is for the others. Since COLLAB, IMDB-BINARY, and IMDB-MULTI have no graph node features, we use the one-hot encoding of node degrees as node features following~\cite{xu*_HowPowerfulAre_2019}. 

For experiments on all benchmarks, the learning rate is set to $1\times 10^{-3}$. We adopt Adam \cite{kingma_AdamMethodStochastic_2015} as optimizer and set weight decay as $1\times 10^{-6}$. Early stopping regularization is employed, where we stop the training if there is no further reduction in the validation loss during 300 epochs. The maximum epoch number is set to 1,000. The batch size is set to 1,024 on OGB-molpcba, 256 on PROTEINS, NCI1, IMDB-BINARY, IMDB-MULTI, and COLLAB. The detailed hyper-parameter settings on all benchmarks are reported in Tab.~\ref{tab:app-hyp}.

\subsection{Node Classification}

\subsubsection{Benchmark Descriptions}
For node classification, we conduct experiments on (\textbf{1}) six middle-scale benchmarks: homophilic graphs~\cite{shchur_PitfallsGraphNeural_2019} (AmazonPhoto, AmazonComputers, CoauthorCS, and CoauthorPhysics), heterophilic graphs~\cite{platonov_CriticalLookEvaluation_2023} (questions, amazon-ratings, tolokers, and minesweeper); (\textbf{2}) four large-scale benchmarks: homophilic graphs~\cite{hu_OpenGraphBenchmark_2020} (OGB-arXiv, OGB-proteins), heterophilic graphs~\cite{lim_LargeScaleLearning_2021} (arXiv-year, genius). 
 
\textbf{CoauthorCS and CoauthorPhysics} originate from the Microsoft Academic Graph, comprising co-authorship graphs wherein nodes denote researchers and edges denote co-authorships between pairs. Node features encapsulate keyword frequencies extracted from an author's publications. Graph labels categorize the dominant research interest in computer science or physics for each author.

\textbf{AmazonComputers and AmazonPhoto} constitute Amazon co-purchase graphs. Nodes denote products that are available for purchase and edges denote co-purchase relation between pairs of items. Node features are encoded customer review texts corresponding to each product. Graph labels categorize the products.

\textbf{Questions} originates from the Yandex Q question-answering website, comprising user activity graphs over a one-year interval (September 2021 to August 2022). Nodes represent users with interest in "medicine". Edges denote one user answered another user's posted question. The associated binary graph classification task involves predicting which users remained active on the website without account deletion or blocking. Node features are derived by averaging FastText word embedding vectors corresponding to user profile descriptions.

\textbf{Amazon-ratings} utilizes the Amazon product co-purchasing network metadata sourced from the SNAP Datasets~\cite{snapnets}. Nodes are products with edges encoding frequent co-purchase relations between item pairs. The associated task involves predicting the average reviewer rating for each product, which is grouped into five ordinal rating classes. Node features are derived by averaging FastText embedding representations corresponding to each product's description text.

\textbf{Tolokers} encapsulates crowdsourcing participation data sourced from the Toloka platform. Nodes denote contributors, referred to as "tolokers", involved in at least one out of 13 selected projects. Edges link toloker pairs that have completed the same tasks. The associated binary classification is to predict which tolokers have been banned from projects. Node features are profile attributes and task performance statistics of each toloker.

\textbf{Minesweeper} is a synthetic 100x100 grid network. Nodes denote grid cells. 20\% of nodes are randomly designated as mines. The associated prediction task is to classify which nodes are mine or not. For all nodes, input features are initialized as one-hot vectors encoding counts of neighboring mines. The initialized features of 50\% of randomly selected nodes are then reset to unknown values. These nodes are indicated by an additional binary indicator.

\textbf{OGB-arXiv} is a citation network of Computer Science papers on arXiv. Nodes represent individual articles whereas directed edges denote one paper citing another. Node features are derived by averaging 128-dimensional word embeddings corresponding to the title and abstract of each publication. The associated multi-class classification is to predict the primary category for arXiv articles across 40 classes.

\textbf{OGB-proteins} is an undirected, weighted graph of proteins. Nodes denote proteins while edges denote biologically meaningful associations between proteins. Edge features represent confidence scores for each association type. Node features are one-hot vectors denoting the species of each protein. The associated multi-label classification is to predict the presence of 112 potential protein functions, formulated as a binary prediction task for each label.

\textbf{Genius} is an online social network. Each node represents a user. The associated binary classification task is to predict whether the accounts are marked or not.

\textbf{arXiv-year} is a citation network from arXiv. Each node represents a research publication. The associated classification task is to predict the publication time of each node.

\begin{table}[t]
\caption{\textbf{Hyper-parameter setups for \nsqure.}}
\label{tab:app-hyp}
\begin{center}
\begin{small}
\begin{sc}
\begin{tabular}{lcccccc} 
\toprule
                & \begin{tabular}[c]{@{}c@{}}\#Recursive\\Steps \\($L$)\end{tabular} & \begin{tabular}[c]{@{}c@{}}Hidden\\~Dim.\end{tabular} & \begin{tabular}[c]{@{}c@{}}State\\~Space\\~Dim.\end{tabular} & \begin{tabular}[c]{@{}c@{}}\#Units\\~($k$)\end{tabular} & \begin{tabular}[c]{@{}c@{}}\#Pseudo\\~Nodes\\~($n_p$)\end{tabular} & Dropout  \\ 
\midrule
Genius          & 3                                                                  & 128                                                   & 64                                                         & 8                                                       & 256                                                              & 0.1      \\
arXiv-year      & 6                                                                  & 128                                                   & 64                                                         & 8                                                       & 300                                                              & 0.1      \\
OGB-arXiv       & 5                                                                  & 128                                                   & 64                                                         & 8                                                       & 320                                                              & 0.3      \\
OGB-proteins    & 3                                                                  & 128                                                   & 64                                                         & 8                                                       & 256                                                              & 0.3      \\
\midrule
questions       & 6                                                                  & 128                                                   & 64                                                         & 8                                                       & 256                                                              & 0.1      \\
amazon-ratings  & 8                                                                  & 64                                                    & 64                                                         & 8                                                       & 256                                                              & 0.3      \\
tolokers        & 6                                                                  & 128                                                   & 64                                                         & 8                                                       & 256                                                              & 0.1      \\
minesweeper     & 7                                                                  & 64                                                    & 64                                                         & 8                                                       & 128                                                              & 0.3      \\
\midrule
CoauthorCS      & 3                                                                  & 128                                                   & 64                                                         & 8                                                       & 256                                                              & 0.4      \\
CoauthorPhysics & 3                                                                  & 128                                                   & 64                                                         & 8                                                       & 256                                                              & 0.5      \\
AmazonPhoto    & 5                                                                  & 128                                                   & 64                                                         & 8                                                       & 256                                                              & 0.1      \\
AmazonComputers & 3                                                                  & 128                                                   & 64                                                         & 8                                                       & 256                                                              & 0.3      \\
\midrule
PROTEINS        & 8                                                                  & 128                                                   & 64                                                         & 8                                                       & 32                                                               & 0.1      \\
NCI1            & 6                                                                  & 128                                                   & 128                                                        & 8                                                       & 32                                                               & 0.1      \\
COLLAB          & 6                                                                  & 128                                                   & 64                                                         & 8                                                       & 32                                                               & 0.1      \\
IMDB-BINARY     & 6                                                                  & 128                                                   & 64                                                         & 8                                                       & 32                                                               & 0.1      \\
IMDB-MULTI      & 6                                                                  & 128                                                   & 64                                                         & 8                                                       & 8                                                                & 0.2      \\
OGB-molpcba     & 6                                                                  & 128                                                   & 64                                                         & 8                                                       & 32                                                               & 0.3      \\
\bottomrule
\end{tabular}
\end{sc}
\end{small}
\end{center}
\end{table}

\subsubsection{Experimental Setups}\label{ssec:node-setup}
For baseline models, we consider (\textbf{1}) Convolutional GNNs: GCN~\cite{kipf_SemiSupervisedClassificationGraph_2017}, GAT~\cite{velickovic_GraphAttentionNetworks_2018}, APPNP~\cite{gasteiger_PredictThenPropagate_2018}, 
SGC~\cite{wu_SimplifyingGraphConvolutional_2019}, H2GCN~\cite{zhu_HomophilyGraphNeural_2020}, FAGCN~\cite{bo_LowfrequencyInformationGraph_2021}, LINKX~\cite{lim_LargeScaleLearning_2021}, GPRGNN~\cite{chien_AdaptiveUniversalGeneralized_2022}, and GloGNN~\cite{li_FindingGlobalHomophily_2022}; (\textbf{2}) Graph transformers: GT~\cite{shi_MaskedLabelPrediction_2021}, Graphormer~\cite{ying_TransformersReallyPerform_2021}, SAN~\cite{kreuzer_RethinkingGraphTransformers_2021}, GraphGPS~\cite{rampasek_RecipeGeneralPowerful_2022}, Nodeformer~\cite{wu_NodeFormerScalableGraph_2022}, NAGphormer~\cite{chen_NAGphormerTokenizedGraph_2023}, SGFormer~\cite{wu_SGFormerSimplifyingEmpowering_2023} and Exphormer~\cite{shirzad_ExphormerSparseTransformers_2023} based on pseudo node.

ROC-AUC is reported for questions, tolokers, and minesweeper while accuracy is for the others. We apply 60\%/20\%/20\% train/val/test random splits for Amazon and Coauthor benchmarks and follow the standard splits as the original papers for the rest of the benchmarks. We reproduce the results of Exphormer~\cite{shirzad_ExphormerSparseTransformers_2023} on (genius, arXiv-year, OGN-Proteins, questions, amazon-ratings, tolokers, minesweeper), Nodeformer~\cite{wu_NodeFormerScalableGraph_2022} and SGFormer~\cite{wu_SGFormerSimplifyingEmpowering_2023} on (genius and arXiv-year) with their released codes for a fair comparison.

For experiments on all benchmarks, the learning rate is set to $1\times 10^{-3}$. We adopt Adam \cite{kingma_AdamMethodStochastic_2015} as optimizer and set weight decay as $1\times 10^{-6}$. Early stopping regularization is employed, where we stop the training if there is no further reduction in the validation loss during 300 epochs. The maximum epoch number is set to 1,000. The detailed hyper-parameter settings on all benchmarks are reported in Tab.~\ref{tab:app-hyp}.

\begin{table}[t]
\caption{\textbf{Effectiveness study on peptides-struct (measured by mean absolute error).}}
\label{tab:app-squashing}
\begin{center}
\begin{small}
\begin{sc}
\begin{tabular}{lc} 
\toprule
          & Peptides-struct $\downarrow$  \\ 
\midrule
GCNII     & 34.71            \\
GCN       & 34.96            \\
GINE      & 35.47            \\
GatedGCN  & 33.57            \\
SAN       & 25.45            \\
PathNN    & 25.45            \\
Drew-GCN  & 25.36            \\
Exphormer & 24.81            \\
\textbf{\nsqure~(Ours)}     & 25.12            \\
\bottomrule
\end{tabular}
\end{sc}
\end{small}
\end{center}
\end{table}

\section{Additional Experimental Results}
\subsection{Effectiveness on Tackling Over-squashing}\label{ssec:app-squshing}
We further benchmark \nsqure~on Peptides-struct~\cite{dwivedi_LongRangeGraph_2022}, a graph regression benchmark involving long-range interactions. Baselines include traditional GNNs: GCN~\cite{kipf_SemiSupervisedClassificationGraph_2017}, GCNII~\cite{chen_SimpleDeepGraph_2020}, GINE~\cite{xu*_HowPowerfulAre_2019, hu*_StrategiesPretrainingGraph_2019}, and GatedGCN~\cite{bresson_ResidualGatedGraph_2018}; methods aiming to capture long-range features: SAN~\cite{kreuzer_RethinkingGraphTransformers_2021}, PathNN~\cite{sun_HomophilyStructureawarePath_2022}, Drew-GCN~\cite{gutteridge_DRewDynamicallyRewired_2023}, and Exphormer~\cite{shirzad_ExphormerSparseTransformers_2023}. As presented in Tab.~\ref{tab:app-squashing}, we can see that \nsqure~achieves competitive performance with Exphormer and surpasses the rest of baselines. This indicates the effectiveness of \nsqure~to encounter over-squashing.

\begin{figure*}[htb]
\centering
\vskip 0.05in
\subfigure{
      \rotatebox{90}{\scriptsize{~~~~~~~~Epoch=0}}
      \begin{minipage}[t]{0.18\linewidth}
      \centering
      \includegraphics[width=\textwidth]{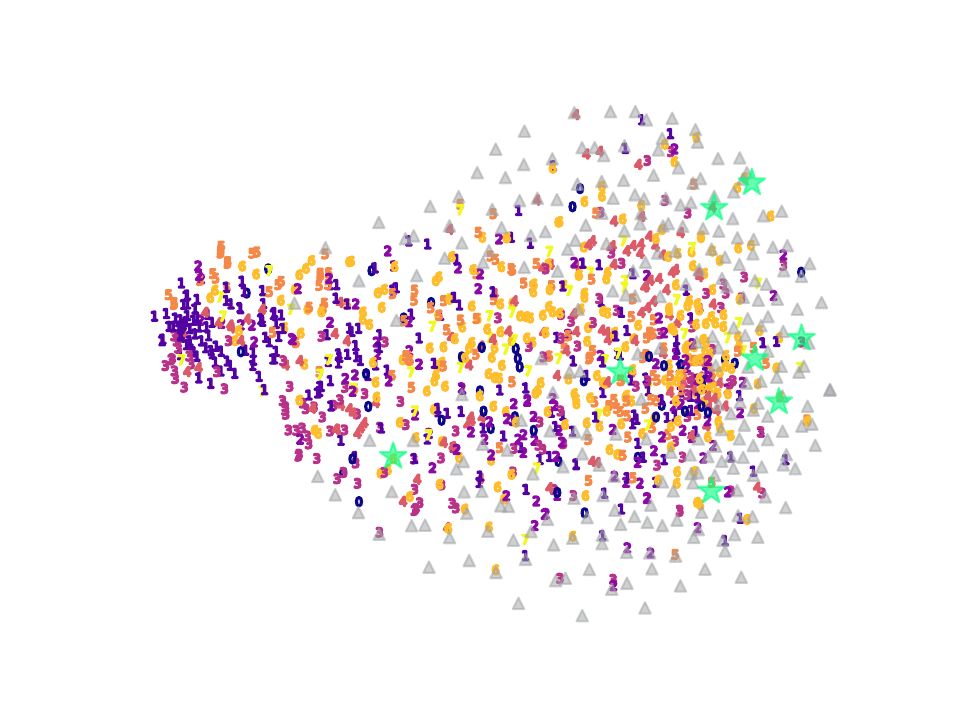}
      \end{minipage}}
 \subfigure{
      \begin{minipage}[t]{0.18\linewidth}
      \centering
      \includegraphics[width=\textwidth]{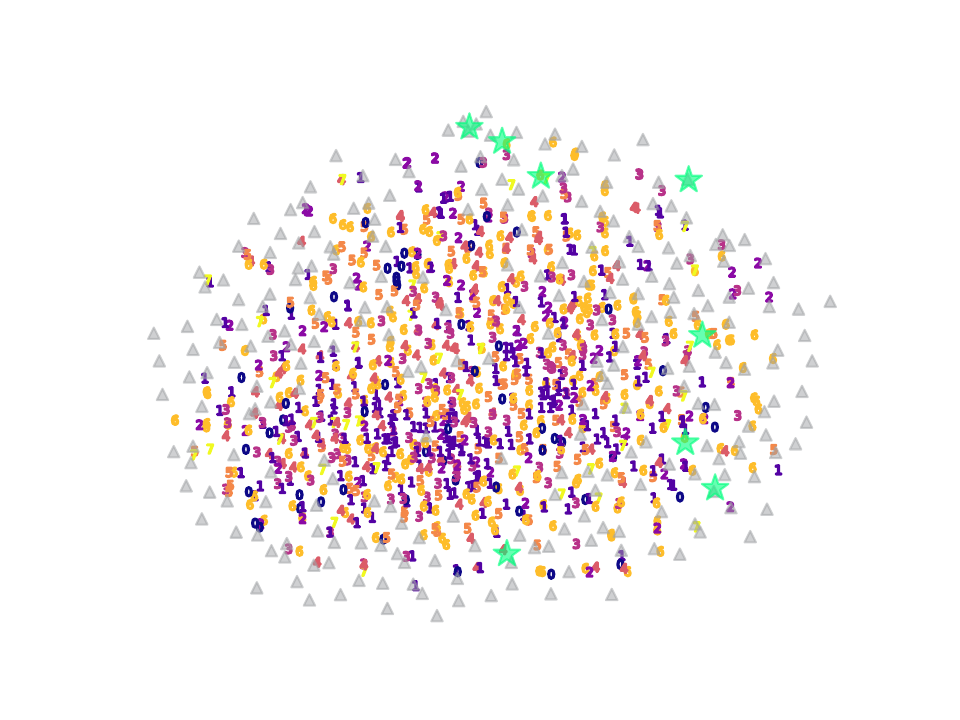}
      \end{minipage}}
 \subfigure{
      \begin{minipage}[t]{0.18\linewidth}
      \centering
      \includegraphics[width=\textwidth]{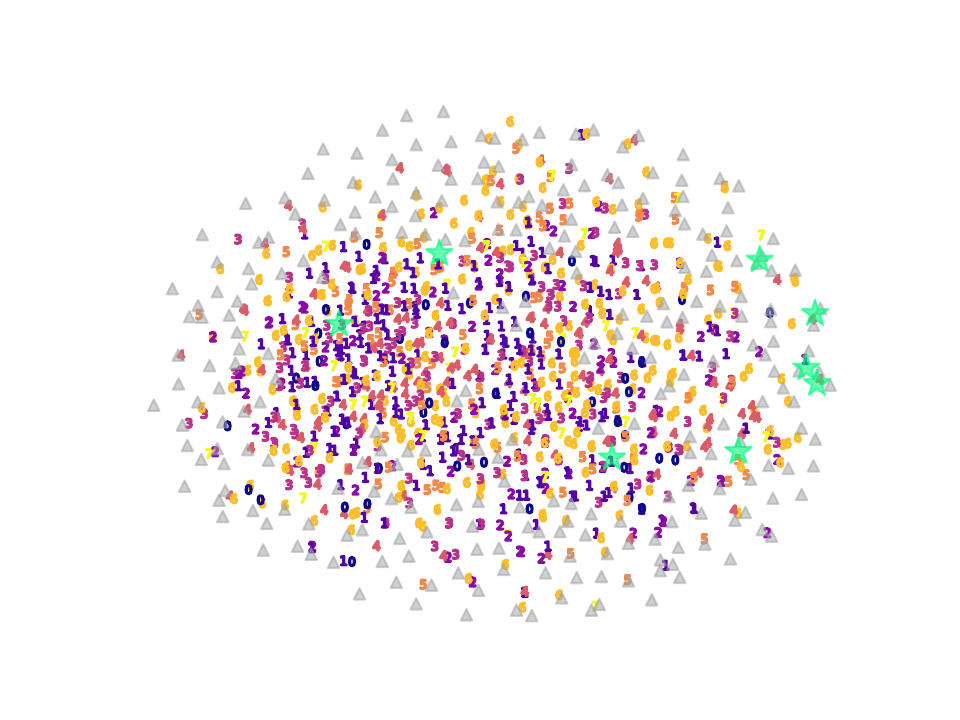}
      \end{minipage}}
\subfigure{
      \begin{minipage}[t]{0.18\linewidth}
      \centering
      \includegraphics[width=\textwidth]{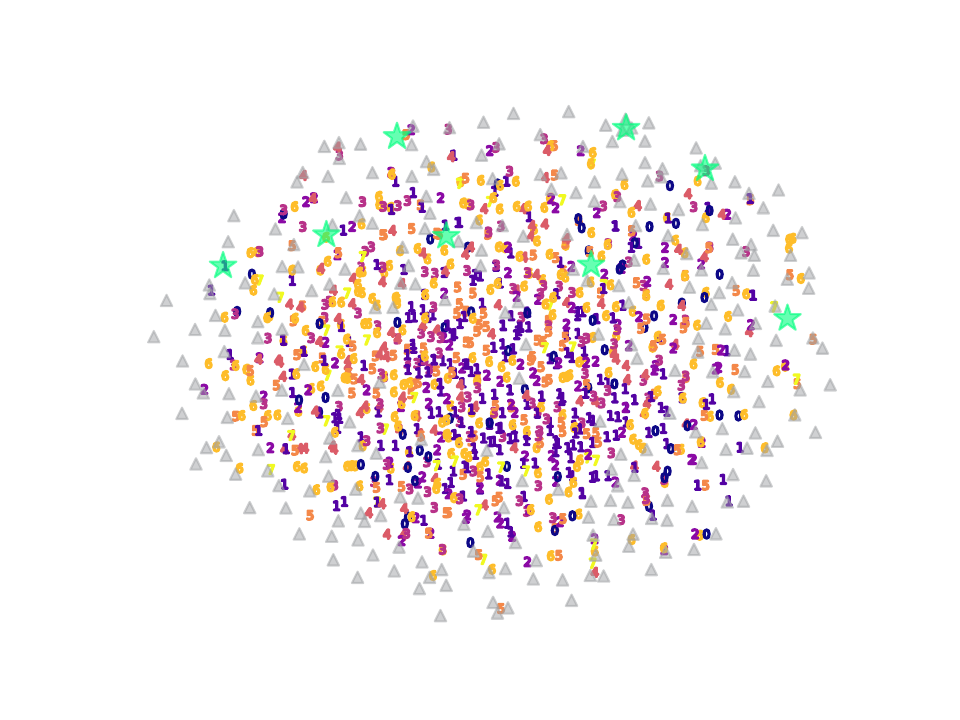}
      \end{minipage}}
 \subfigure{
      \begin{minipage}[t]{0.18\linewidth}
      \centering
      \includegraphics[width=\textwidth]{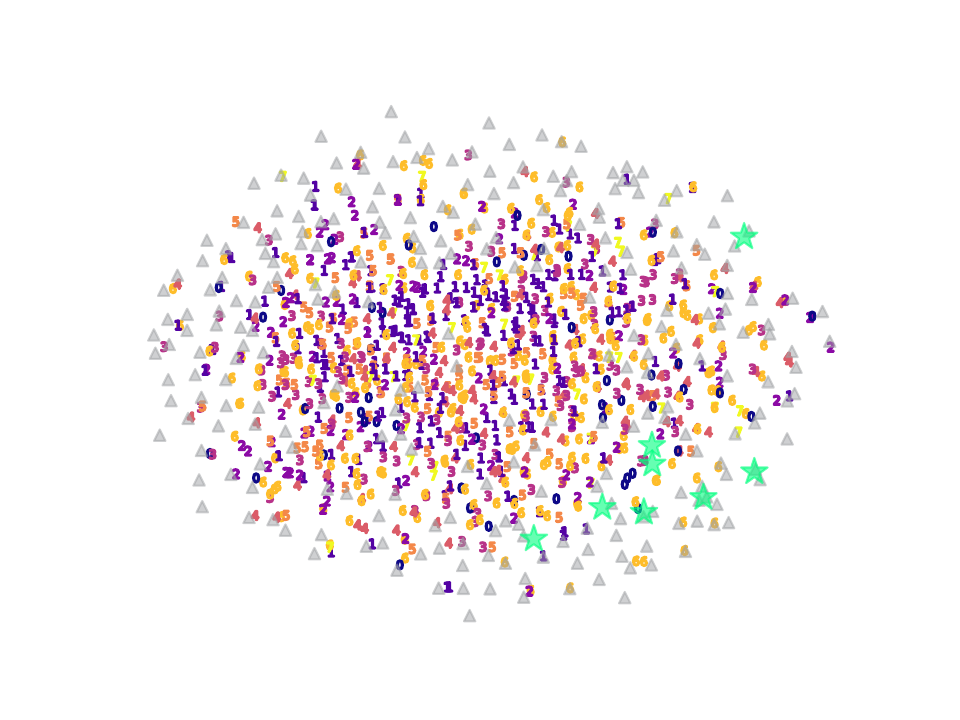}
      \end{minipage}}
\setcounter{subfigure}{0}
\subfigure{
      \rotatebox{90}{\scriptsize{~~~~~~~Epoch=20}}
      \begin{minipage}[t]{0.18\linewidth}
      \centering
      \includegraphics[width=\textwidth]{fig/distribution/020-0.pdf}
      \end{minipage}}
 \subfigure{
      \begin{minipage}[t]{0.18\linewidth}
      \centering
      \includegraphics[width=\textwidth]{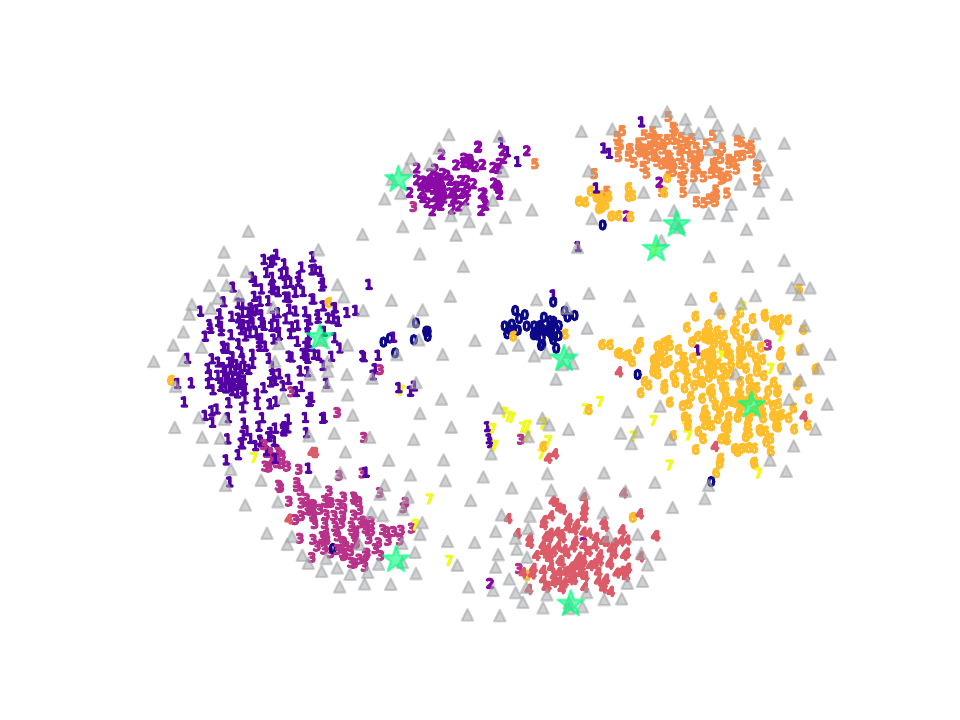}
      \end{minipage}}
 \subfigure{
      \begin{minipage}[t]{0.18\linewidth}
      \centering
      \includegraphics[width=\textwidth]{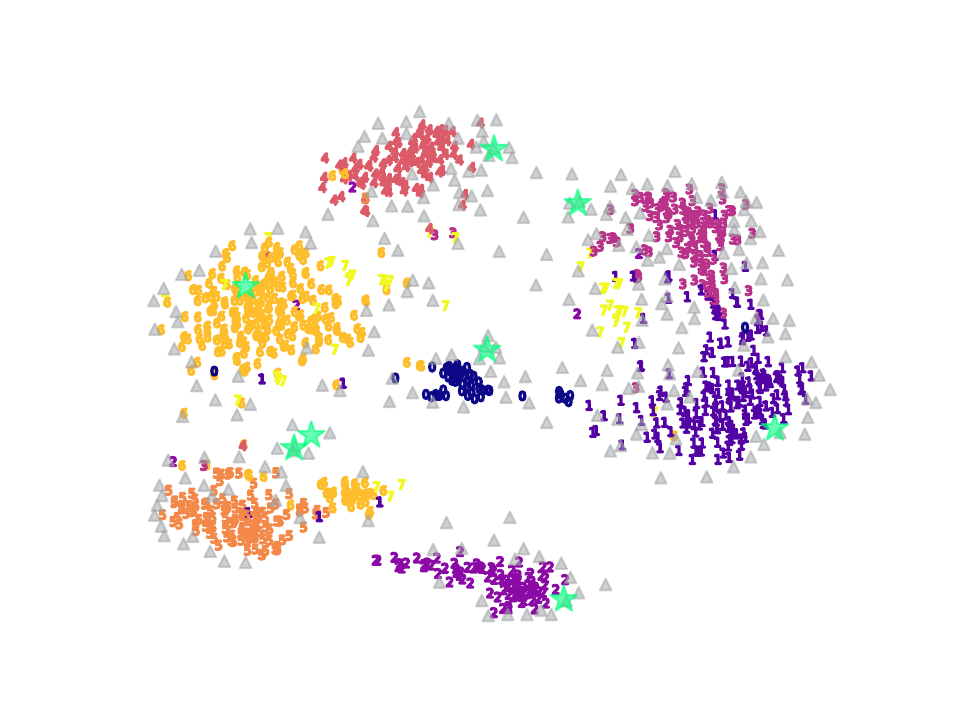}
      \end{minipage}}
\subfigure{
      \begin{minipage}[t]{0.18\linewidth}
      \centering
      \includegraphics[width=\textwidth]{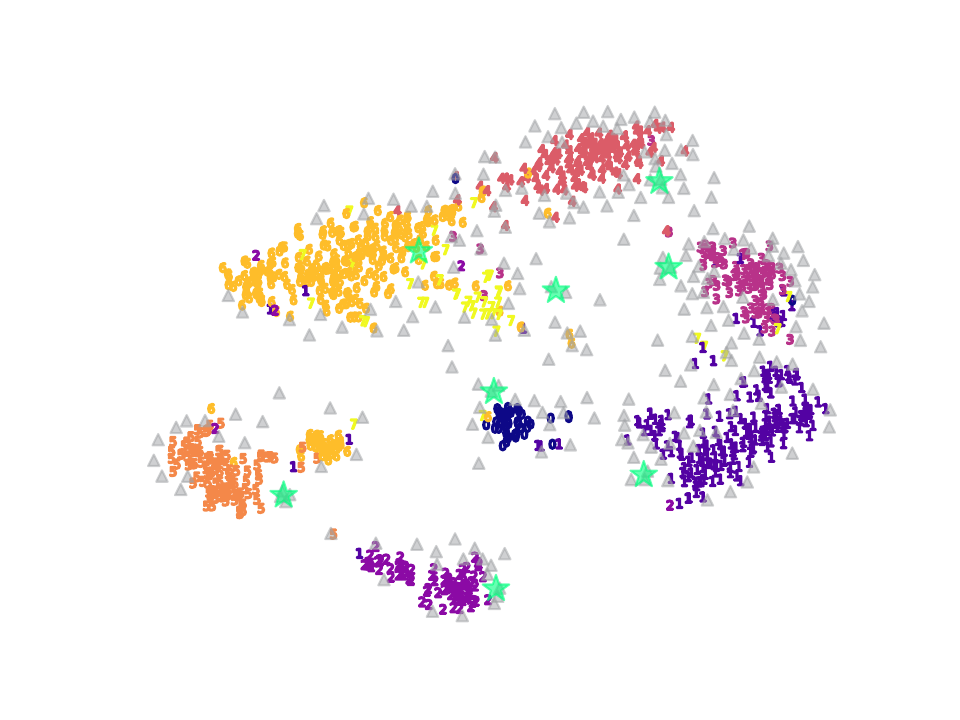}
      \end{minipage}}
 \subfigure{
      \begin{minipage}[t]{0.18\linewidth}
      \centering
      \includegraphics[width=\textwidth]{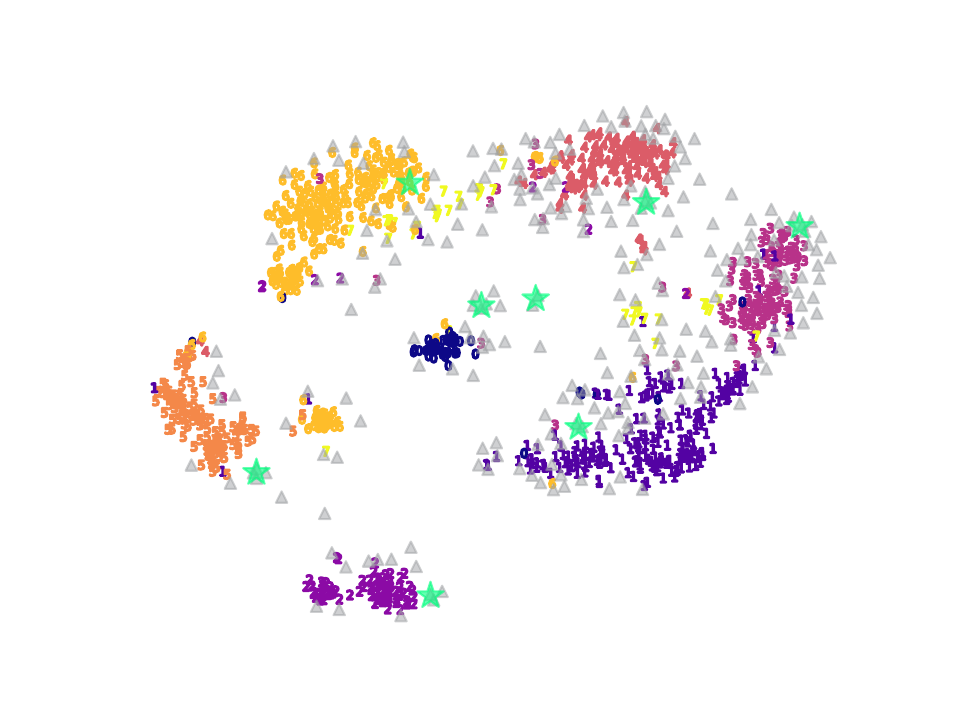}
      \end{minipage}}
\setcounter{subfigure}{0}
\subfigure[$l=0$]{
      \rotatebox{90}{\scriptsize{~~~~~~Epoch=500}}
      \begin{minipage}[t]{0.18\linewidth}
      \centering
      \includegraphics[width=\textwidth]{fig/distribution/500-0.pdf}
      \end{minipage}}
 \subfigure[$l=1$]{
      \begin{minipage}[t]{0.18\linewidth}
      \centering
      \includegraphics[width=\textwidth]{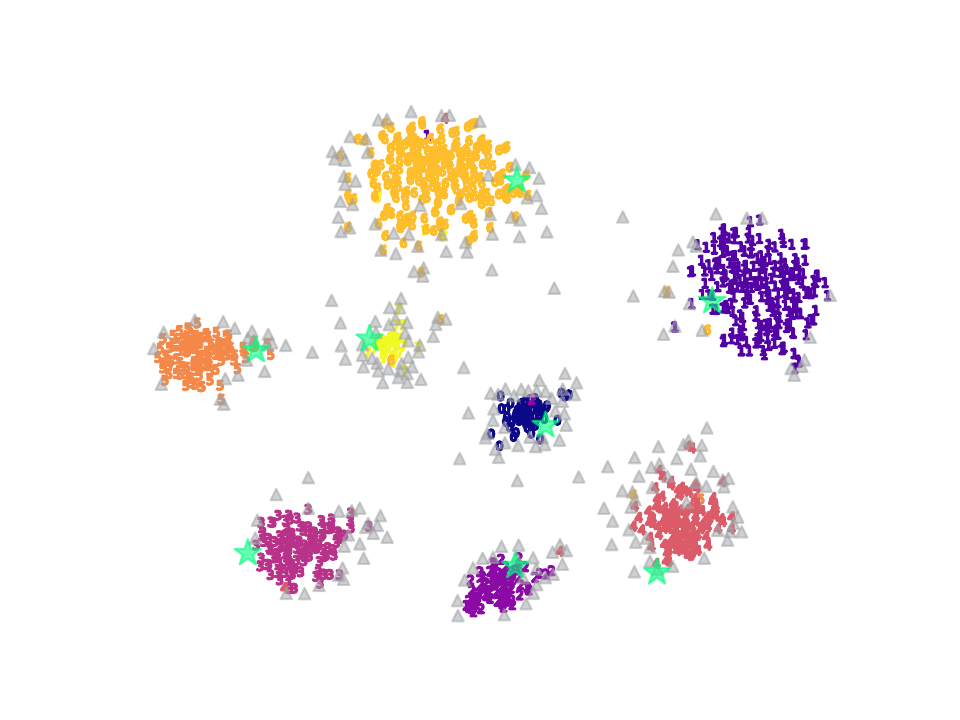}
      \end{minipage}}
 \subfigure[$l=2$]{
      \begin{minipage}[t]{0.18\linewidth}
      \centering
      \includegraphics[width=\textwidth]{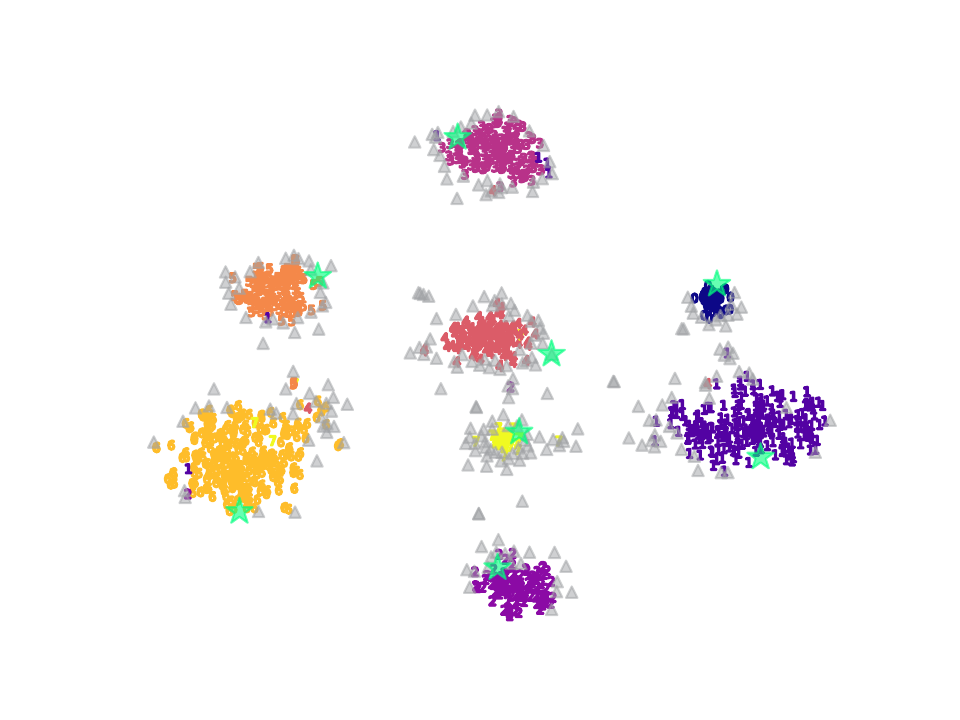}
      \end{minipage}}
\subfigure[$l=3$]{
      \begin{minipage}[t]{0.18\linewidth}
      \centering
      \includegraphics[width=\textwidth]{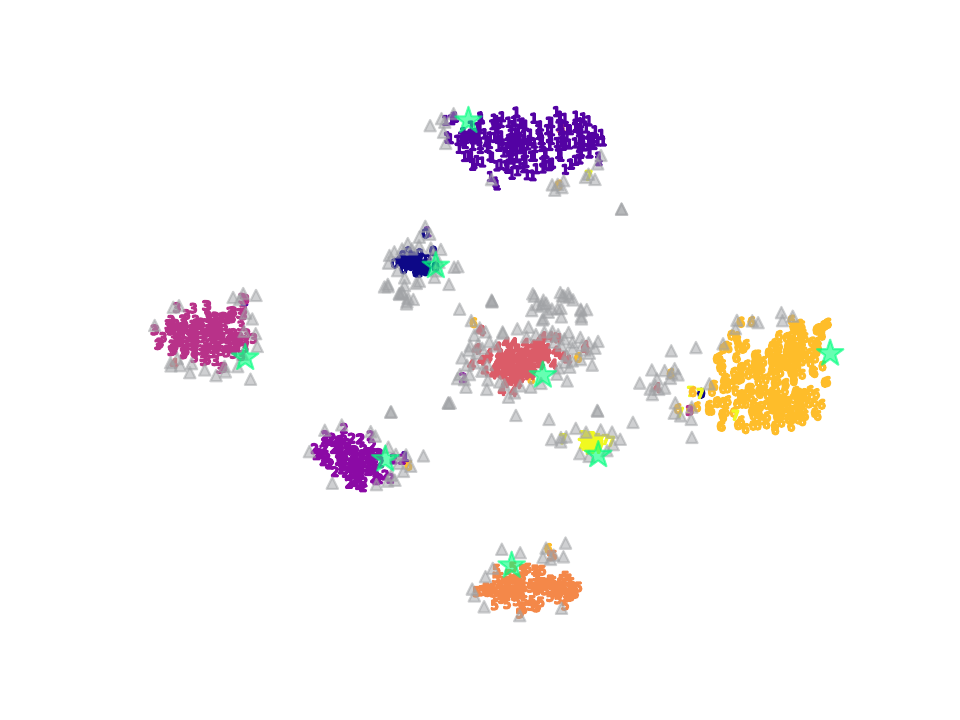}
      \end{minipage}}
 \subfigure[$l=4$]{
      \begin{minipage}[t]{0.18\linewidth}
      \centering
      \includegraphics[width=\textwidth]{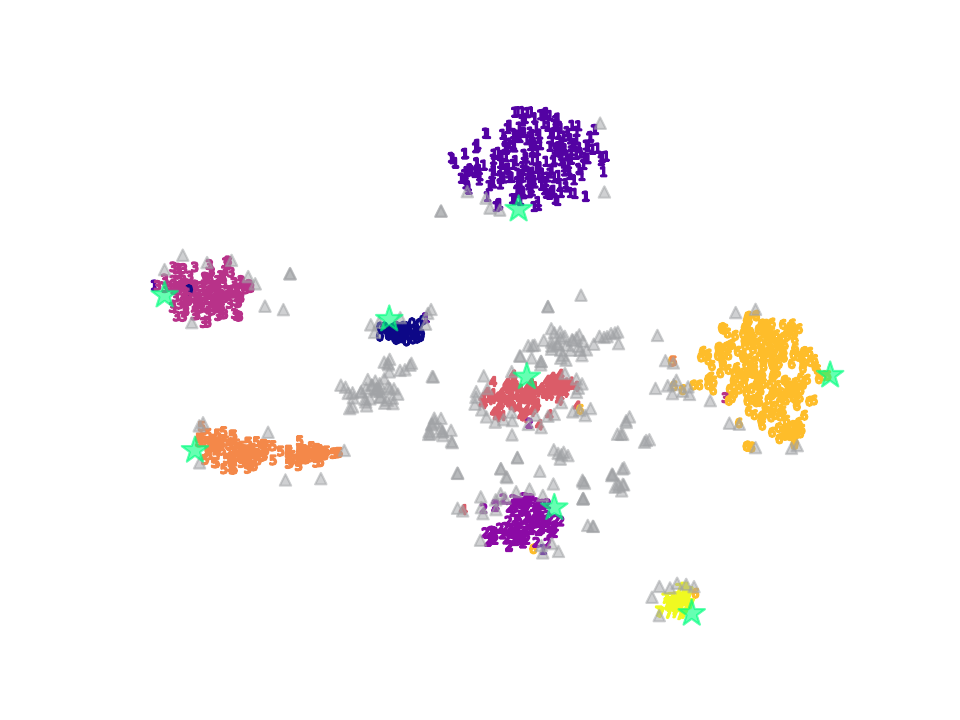}
      \end{minipage}}
\caption{\textbf{Full results on the distribution of embedded nodes.} The t-sne~\cite{t-SNE} results under different recursive steps across training are compared. Each row is from the same epoch. Each column depicts results from the same recursive step ($l$-th step). Input graph nodes with different labels are depicted as: \textbf{\textcolor[RGB]{22,17,139}{0}, \textcolor[RGB]{83,2,163}{1}, \textcolor[RGB]{139,10,165}{2}, \textcolor[RGB]{184,50,137}{3}, \textcolor[RGB]{219,92,104}{4}, \textcolor[RGB]{244,136,74}{5}, \textcolor[RGB]{254,189,42}{6}}. Pseudo nodes are depicted as \textbf{\textcolor[RGB]{197,198,199}{$\bigtriangleup$}}. Class nodes are depicted as \textbf{\textcolor[RGB]{76,255,164}{$\star$}}.}
\label{fig:app-distribution}
\end{figure*}

\subsection{Distribution of Embedded Nodes}\label{ssec:app-distribution}
The distribution of the embedded nodes through training is visualized in Fig.~\ref{fig:app-distribution}. We can see that pseudo nodes and graph nodes adjust their relative position actively in the state space. For the same recursive step through training, pseudo nodes are split into several groups. Each group is attracted toward a distinct graph node cluster. For the same epoch, the relative positions between graph node clusters and the attracted pseudo nodes also evolve through recursive steps. This indicates that \nsqure~optimizes the proximity between graph nodes and pseudo nodes recursively, constructing dynamic message-passing pathways.

\begin{figure*}[htb]
\centering
\subfigure{
      \rotatebox{90}{\scriptsize{~~~~~~~~~~~~$l=1$}}
      \begin{minipage}[t]{0.22\linewidth}
      \centering
      \includegraphics[width=\textwidth]{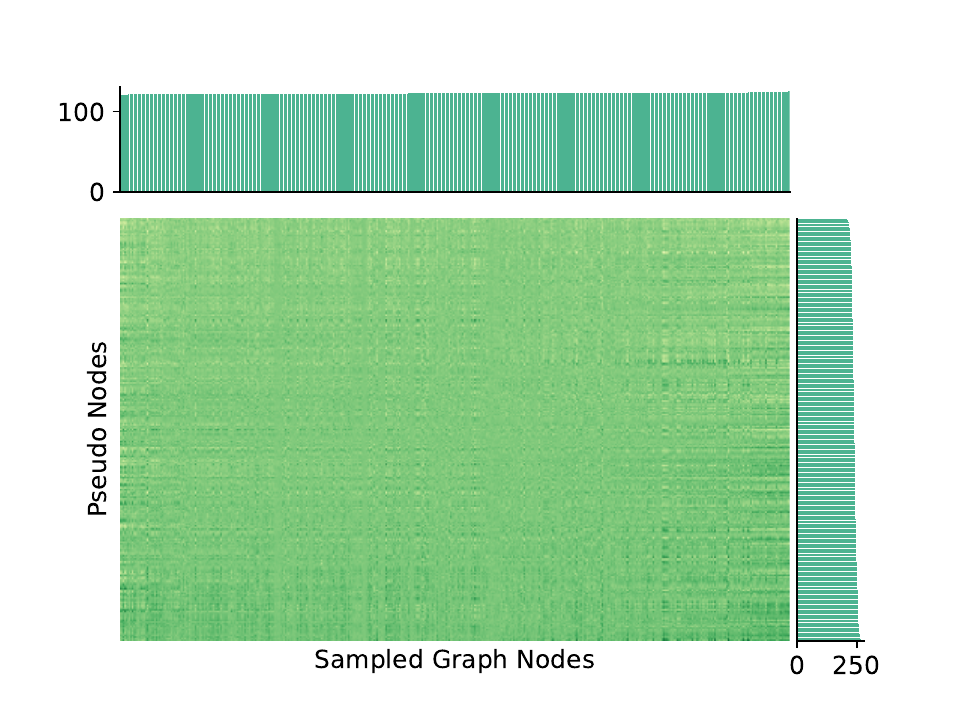}
      \end{minipage}}
 \subfigure{
      \begin{minipage}[t]{0.22\linewidth}
      \centering
      \includegraphics[width=\textwidth]{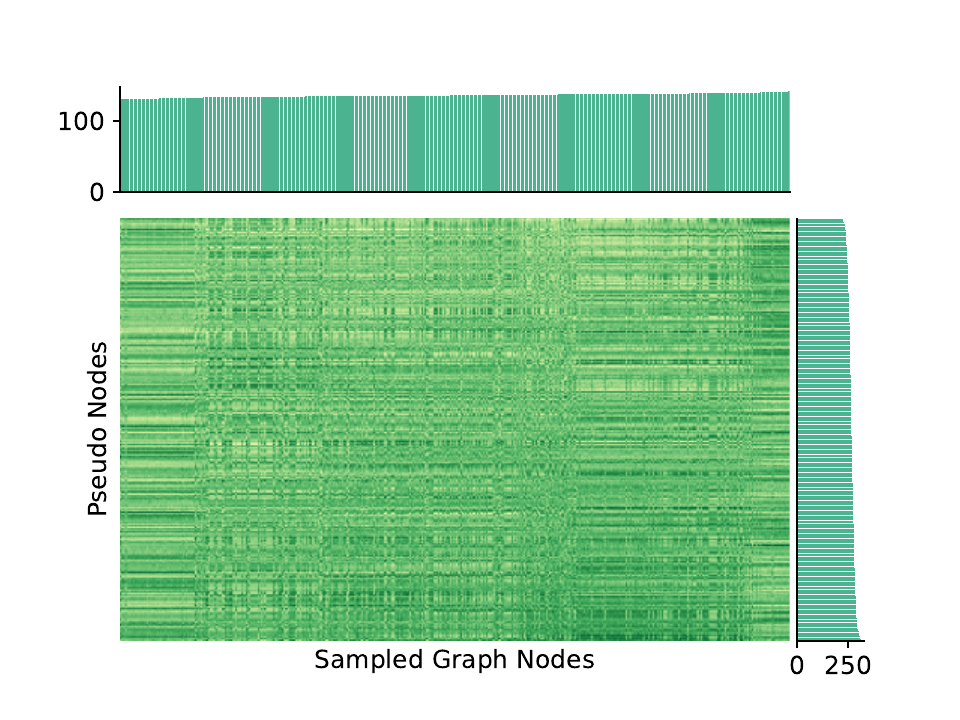}
      \end{minipage}}
 \subfigure{
      \begin{minipage}[t]{0.22\linewidth}
      \centering
      \includegraphics[width=\textwidth]{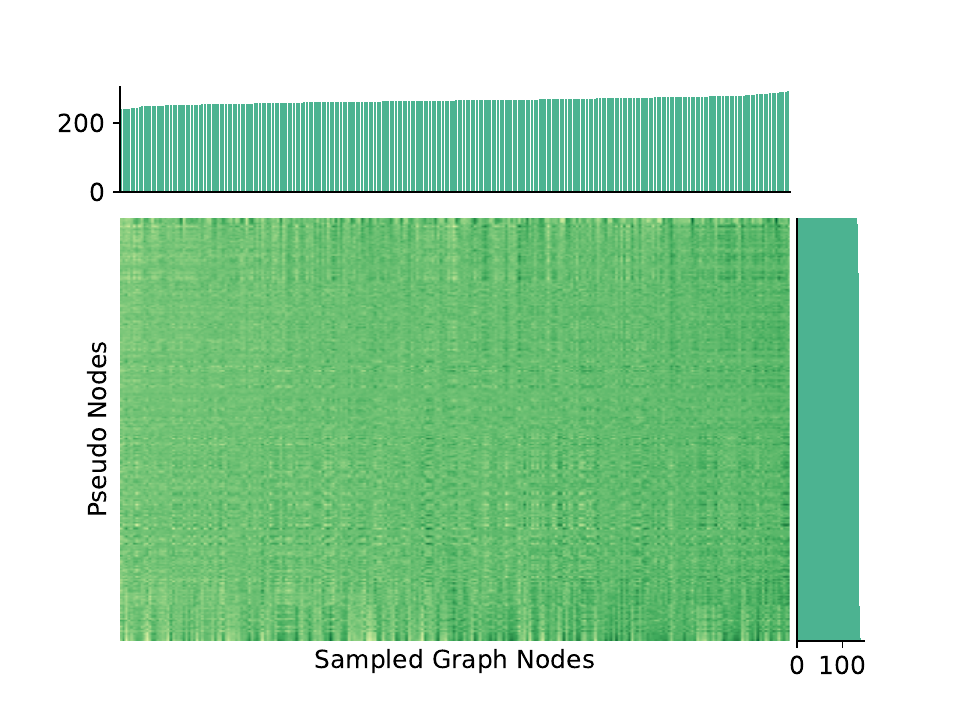}
      \end{minipage}}
\subfigure{
      \begin{minipage}[t]{0.22\linewidth}
      \centering
      \includegraphics[width=\textwidth]{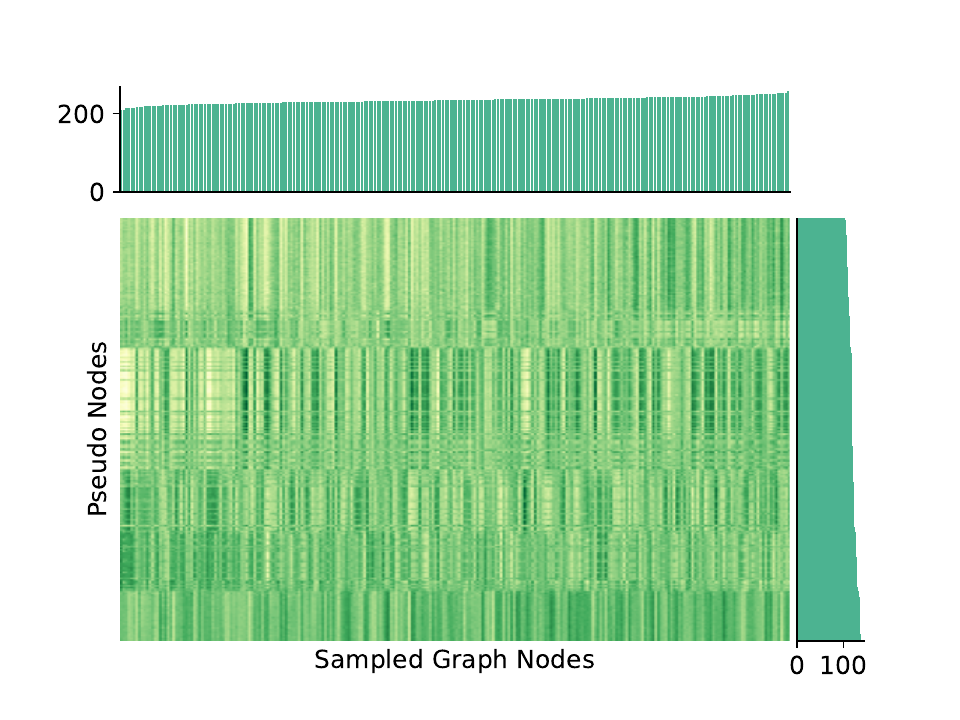}
      \end{minipage}}
\setcounter{subfigure}{0}
\subfigure{
      \rotatebox{90}{\scriptsize{~~~~~~~~~~~~$l=2$}}
      \begin{minipage}[t]{0.22\linewidth}
      \centering
      \includegraphics[width=\textwidth]{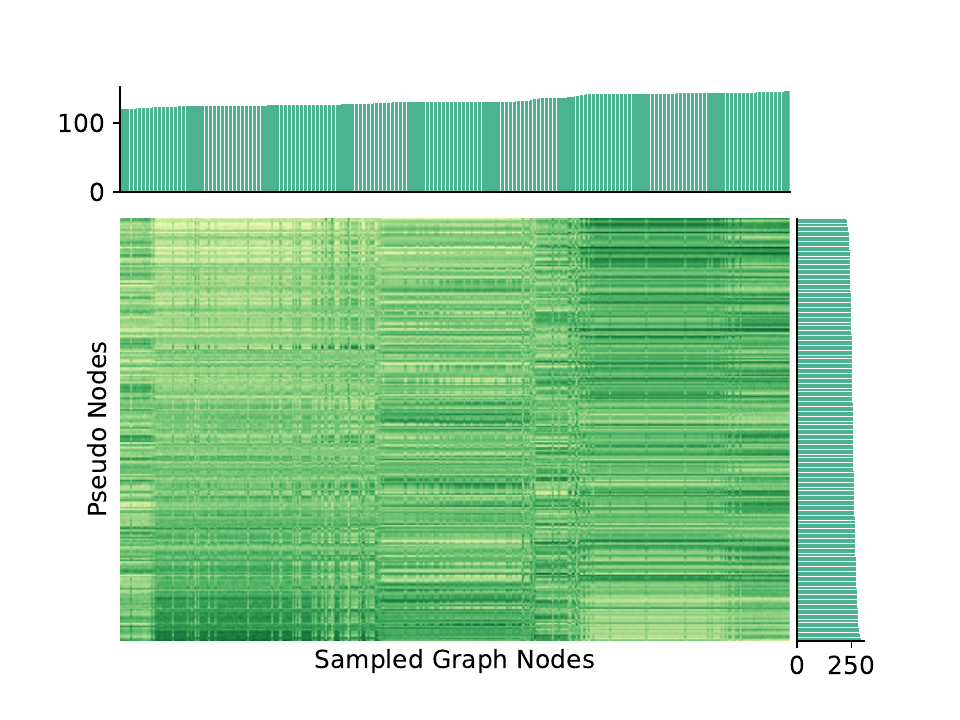}
      \end{minipage}}
\subfigure{
      \begin{minipage}[t]{0.22\linewidth}
      \centering
      \includegraphics[width=\textwidth]{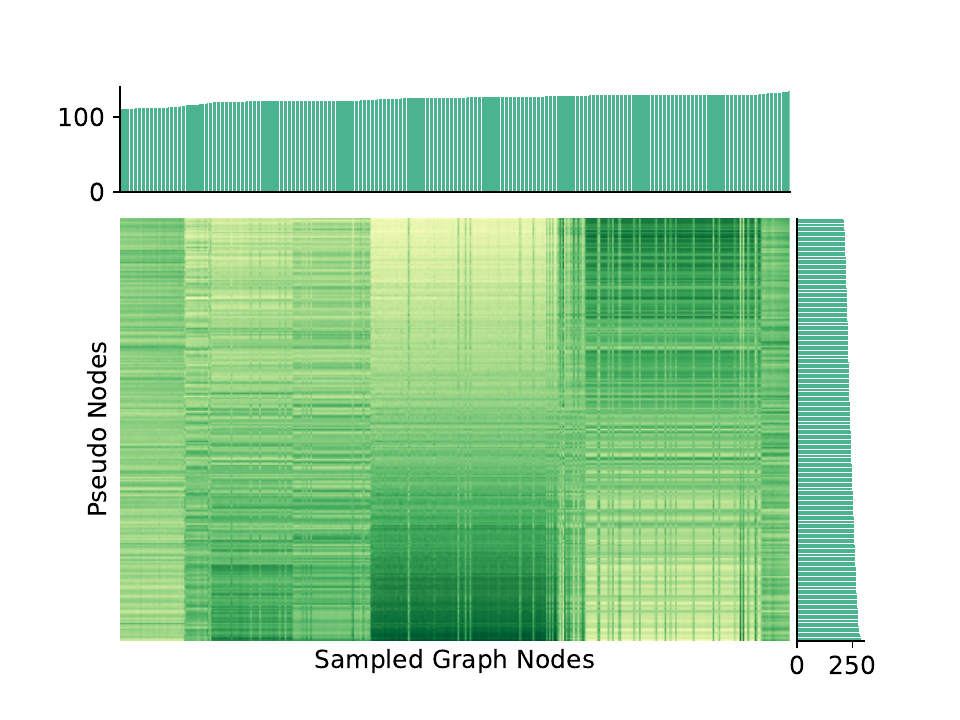}
      \end{minipage}}
\subfigure{
      \begin{minipage}[t]{0.22\linewidth}
      \centering
      \includegraphics[width=\textwidth]{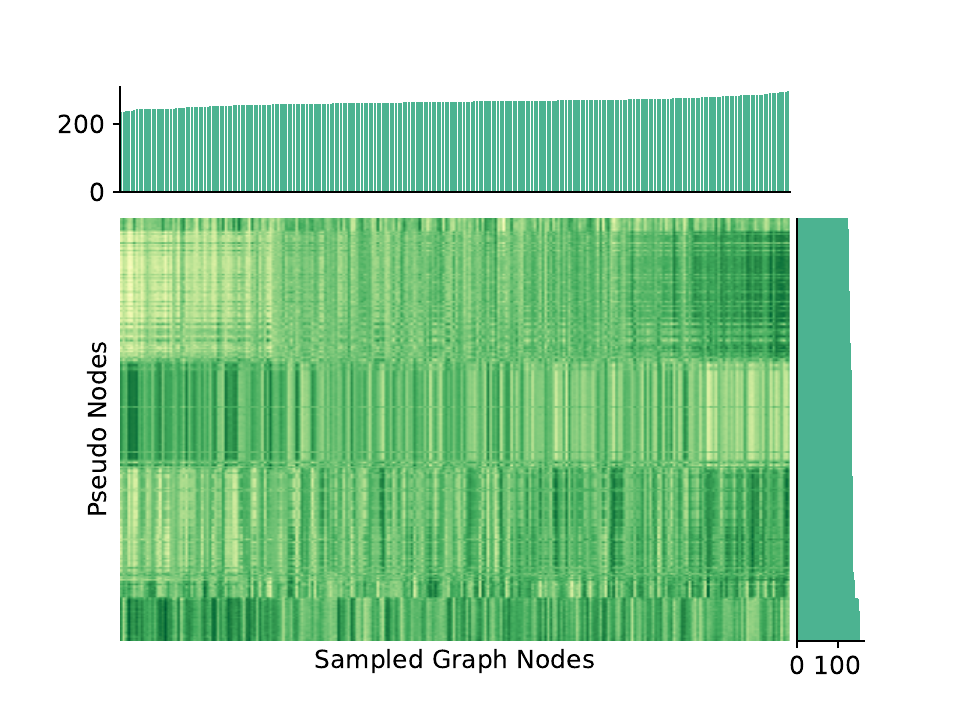}
      \end{minipage}}
\subfigure{
      \begin{minipage}[t]{0.22\linewidth}
      \centering
      \includegraphics[width=\textwidth]{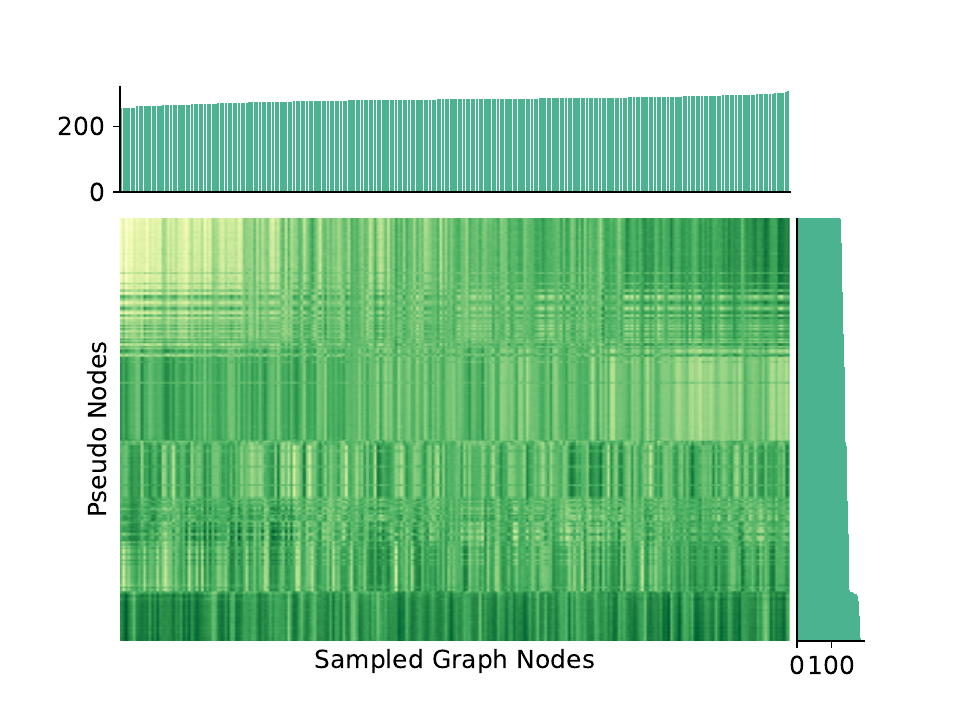}
      \end{minipage}}
\setcounter{subfigure}{0}
\subfigure{
      \rotatebox{90}{\scriptsize{~~~~~~~~~~~~$l=3$}}
      \begin{minipage}[t]{0.22\linewidth}
      \centering
      \includegraphics[width=\textwidth]{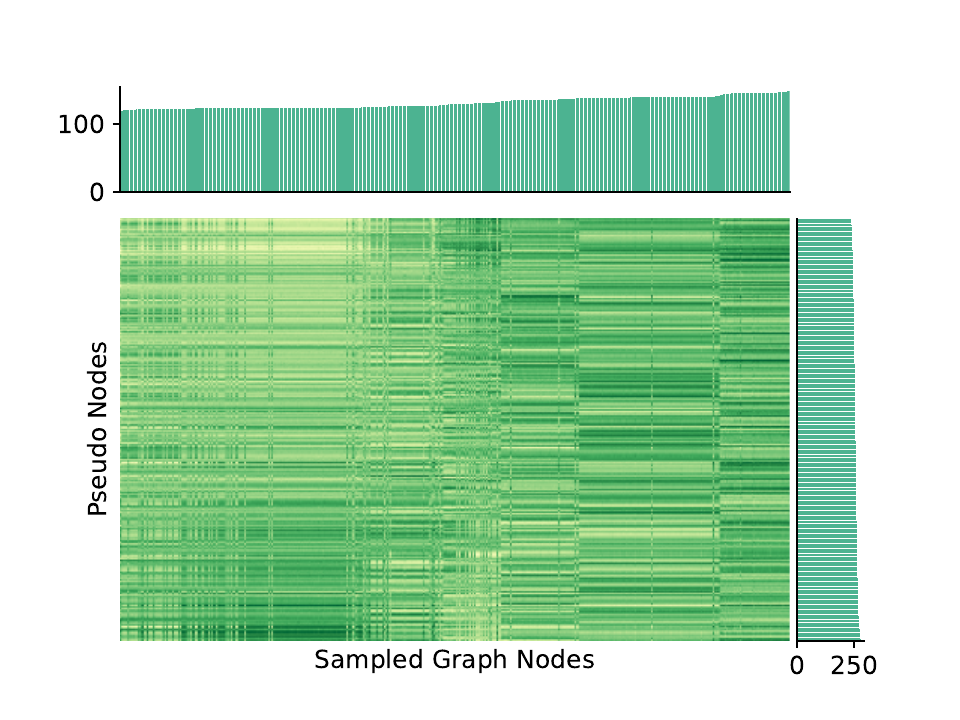}
      \end{minipage}}
\subfigure{
      \begin{minipage}[t]{0.22\linewidth}
      \centering
      \includegraphics[width=\textwidth]{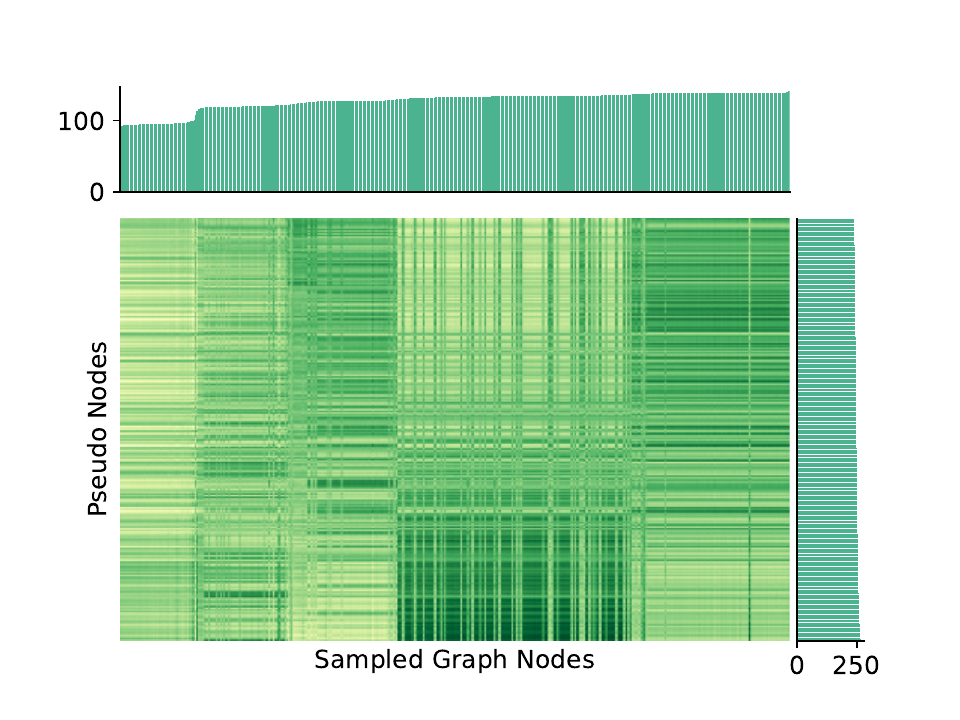}
      \end{minipage}}
\subfigure{
      \begin{minipage}[t]{0.22\linewidth}
      \centering
      \includegraphics[width=\textwidth]{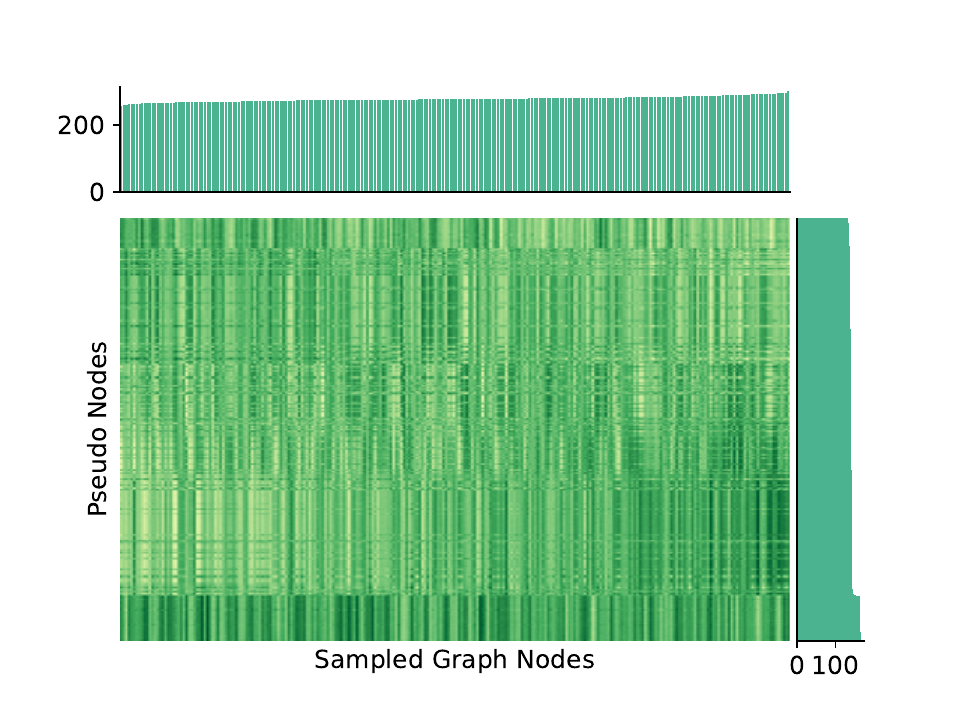}
      \end{minipage}}
\subfigure{
      \begin{minipage}[t]{0.22\linewidth}
      \centering
      \includegraphics[width=\textwidth]{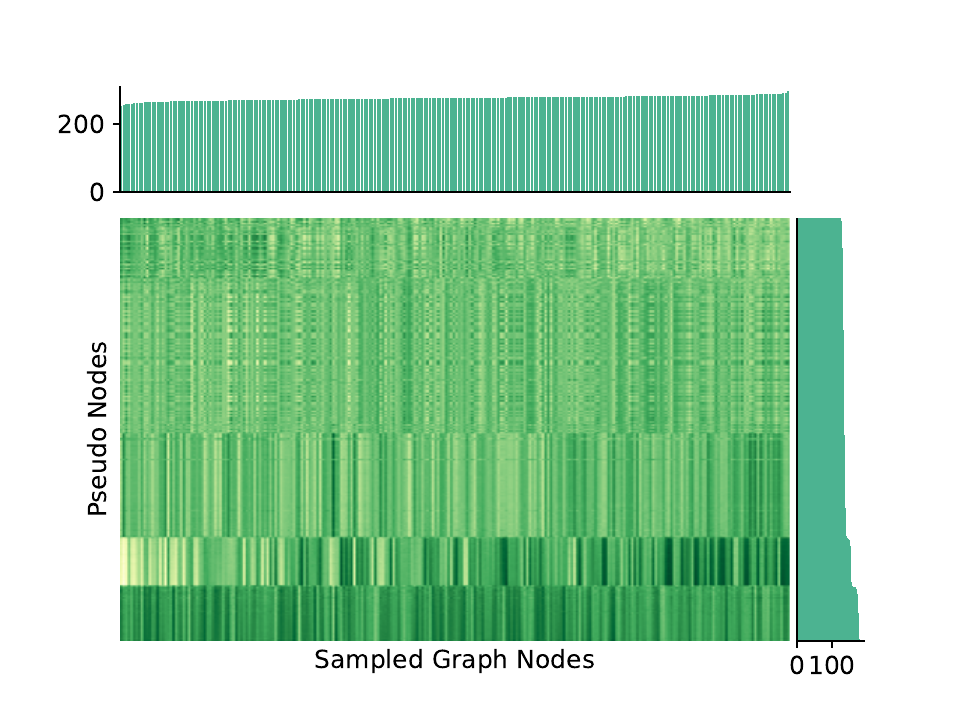}
      \end{minipage}}
\setcounter{subfigure}{0}
\subfigure[PNode. Adapt. $\mathrb{E}^\mathtt{np}$]{
      \rotatebox{90}{\scriptsize{~~~~~~~~~~~~$l=4$}}
      \begin{minipage}[t]{0.22\linewidth}
      \centering
      \includegraphics[width=\textwidth]{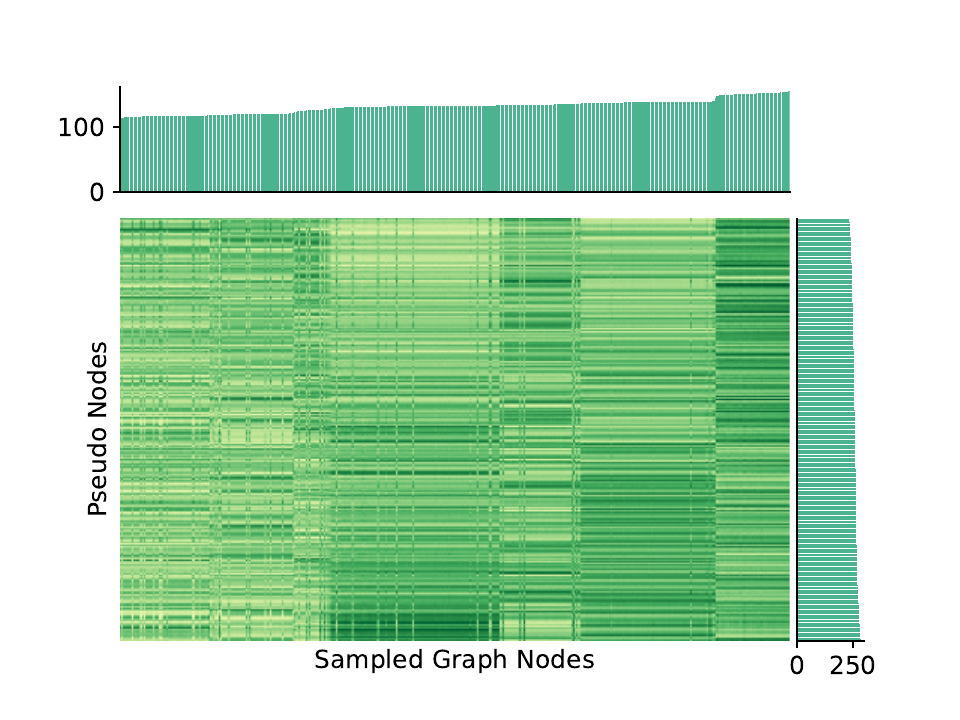}
      \end{minipage}}
\subfigure[PNode Adapt. $\mathrb{E}^{\mathtt{pn}\top}$]{
      \begin{minipage}[t]{0.22\linewidth}
      \centering
      \includegraphics[width=\textwidth]{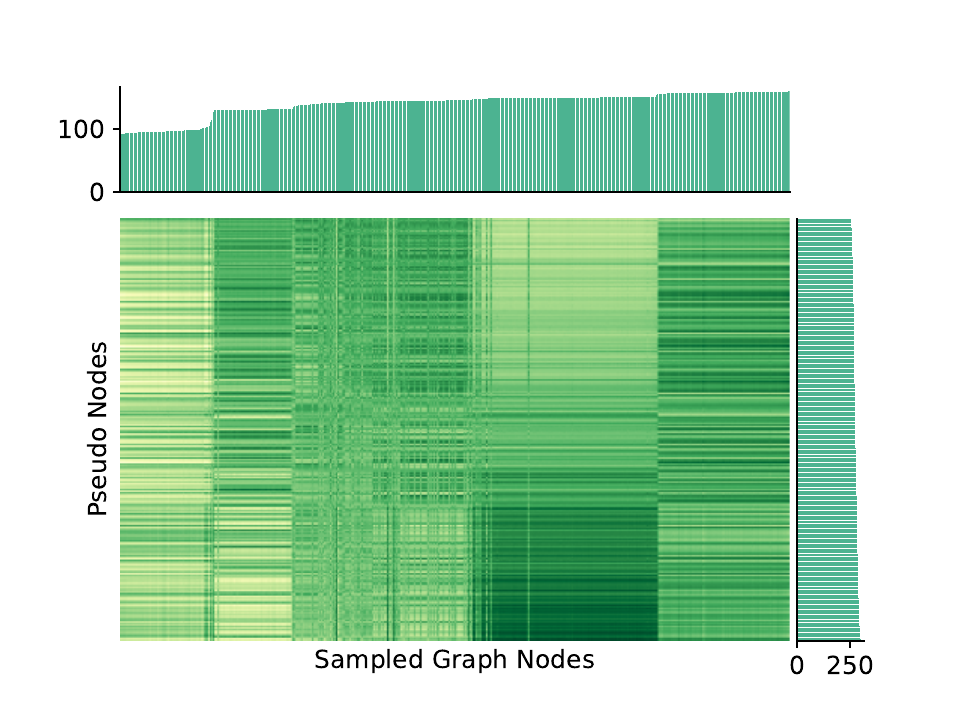}
      \end{minipage}}
\subfigure[Dy. MP. $\mathrb{E}^\mathtt{np}$]{
      \begin{minipage}[t]{0.22\linewidth}
      \centering
      \includegraphics[width=\textwidth]{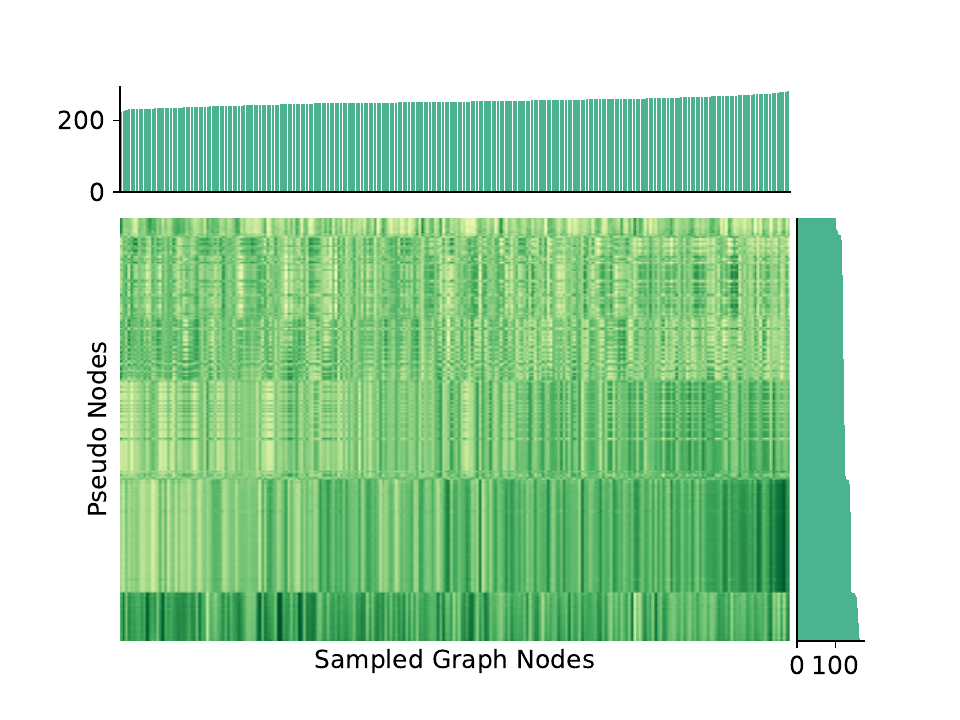}
      \end{minipage}}
\subfigure[Dy. MP. $\mathrb{E}^{\mathtt{pn}\top}$]{
      \begin{minipage}[t]{0.22\linewidth}
      \centering
      \includegraphics[width=\textwidth]{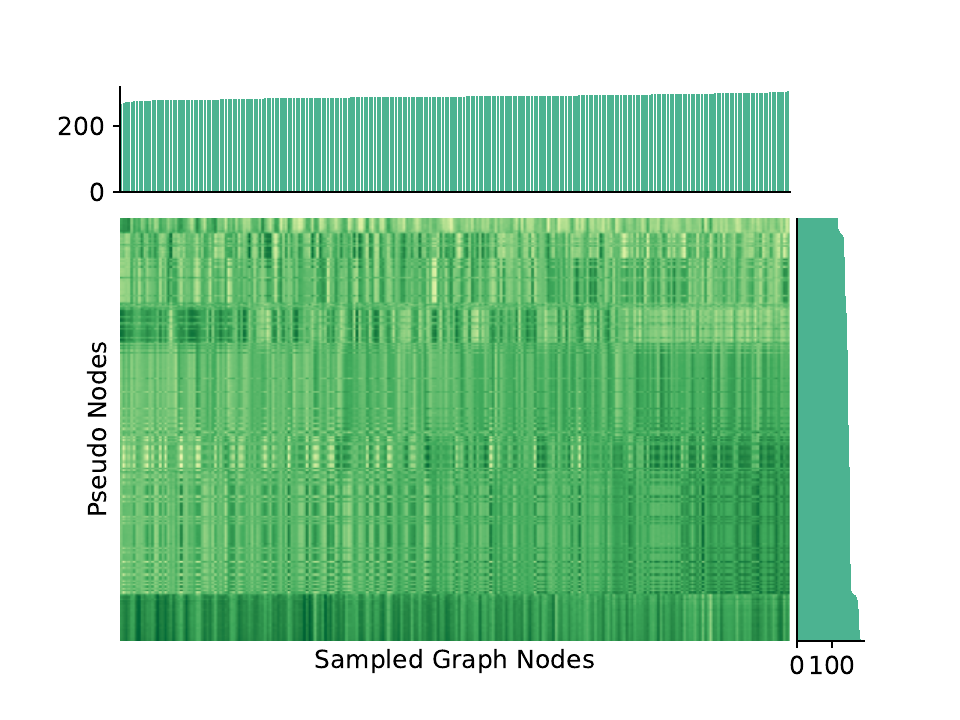}
      \end{minipage}}
\caption{\textbf{Full results on the message passing analysis.} The results under different recursive steps across training are compared. Each row is from the same recursive step ($l$-th step). PNode. Adpat. denotes the pseudo-node adaptation module in \nsqure. Dy. MP. denotes the dynamic message passing in \nsqure. The proximities between sampled graph nodes and pseudo nodes are depicted in the center of each sub-figure, where \textcolor[RGB]{0,96,51}{darker green} indicates higher proximity and \textcolor[RGB]{81,179,101}{brighter green} indicates lower proximity. Each column of the proximity matrix associates with a graph node while each row associates a pseudo node. The distribution on the top of each sub-figure denotes the sum of proximity for each graph node while the distribution on the right is for each pseudo node. The sampled graph nodes and pseudo nodes are ranked based on the sum of proximity.}
\label{fig:app-mp}
\end{figure*}

\subsection{Proximity for Message Passing}\label{ssec:app-mp}
To further analyze the message passing between graph nodes and pseudo nodes, we visualize their corresponding proximity on AmazonPhoto. In Fig.~\ref{fig:app-mp}, $1,000$ graph nodes are sampled randomly. Both graph nodes and pseudo nodes are ranked based on their intro-/outre-proximity summation. The intro-/outre-proximity summation indicates the message load a graph node takes or emits during the message passing. From Fig.~\ref{fig:app-mp}, we can see that the intro-/outre-proximity is distributed evenly across different graph nodes and pseudo nodes. This indicates that both graph nodes and pseudo nodes assume a balanced message load. For each proximity matrix, the proximity value varies between different pairs of graph nodes and pseudo nodes, thus constructing dynamic pathways instead of uniform pathways.

\subsection{Displacements Comparison between Single Shared Recurrent Layer and Multiple Recurrent Layers}\label{ssec:app-disp}
\begin{figure*}
    \begin{minipage}{\linewidth}
        \centering
        \subfigure[amazon-ratings]{
              \begin{minipage}[t]{0.45\linewidth}
              \centering
              \includegraphics[width=\textwidth]{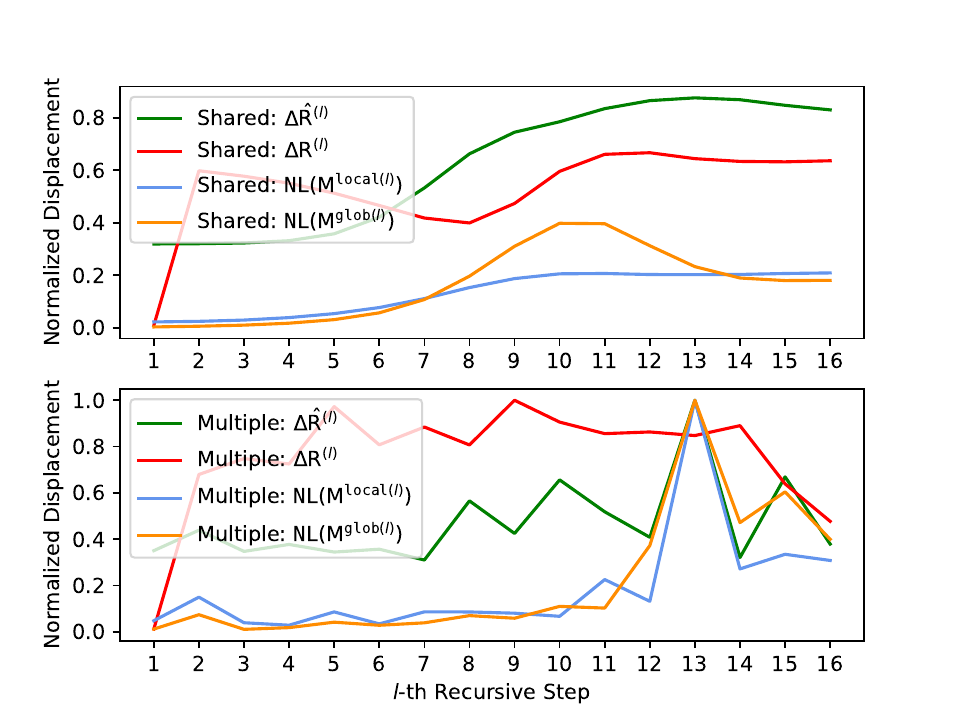}
              \end{minipage}}%
        \hfill
        \subfigure[AmazonPhoto]{
              \begin{minipage}[t]{0.45\linewidth}
              \centering
              \includegraphics[width=\textwidth]{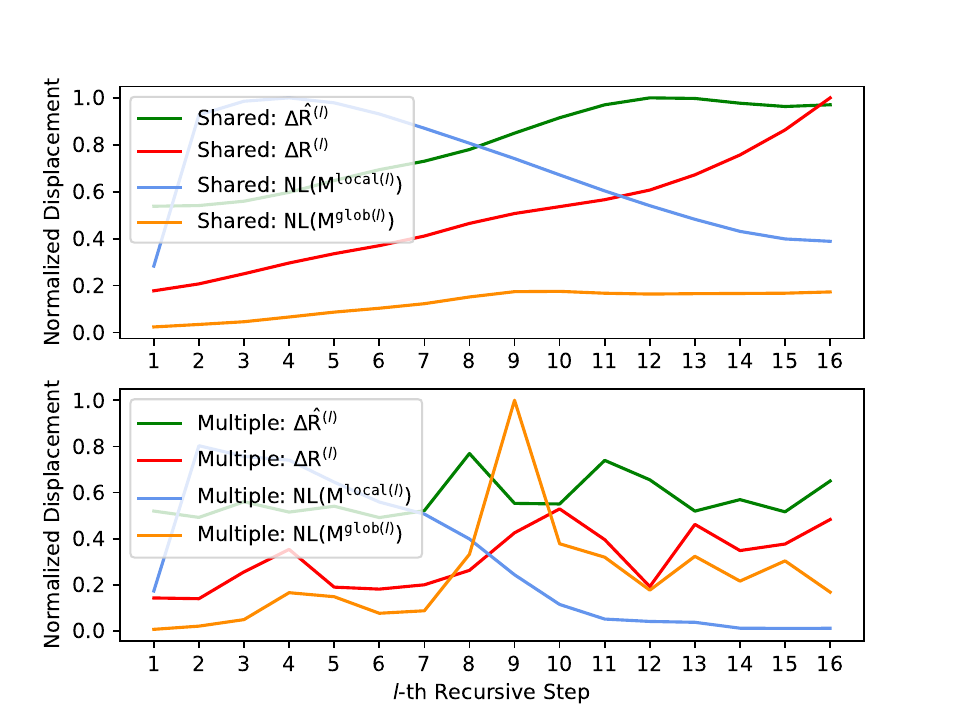}
              \end{minipage}}
        \vspace{-0.1in}
        \figcaption{\textbf{Displacement comparison between \nsqure~with a shared recurrent layer and \nsqure~with multiple recurrent layers.}}
        \label{fig:app-disp}
    \end{minipage}
\end{figure*}

To understand the difference in multi-step performance, where \nsqure~with multiple recurrent layers ($L_p=L$) achieves better performance against \nsqure~with shared parameters ($L_p=1$), we further analyze the displacements of the embedded nodes on AmazonPhoto and amazon-ratings. Fig.~\ref{fig:app-disp} depicts the comparison results.  Four displacement types are presented in the figure, including displacements of pseudo nodes in pseudo-node adaptation $\Delta\mathrb{\hat{R}}^{(l)}$ (Eq.~\ref{eq:adaptation}) and global message passing $\Delta\mathrb{R}^{(l)}$ (Eq.~\ref{eq:nsqure-global-mp}), displacements of graph nodes in local message passing $\mathtt{NL}(\mathrb{M}^{\mathtt{local}(l)})$ (Eq.~\ref{eq:nsqure-local-mp}) and global message passing $\mathtt{NL}(\mathrb{M}^{\mathtt{glob}(l)})$ (Eq.~\ref{eq:nsqure-global-mp}). We take the Frobenius norm of each displacement matrix and divide it by the larger results between \nsqure~with $L_p=1$ and \nsqure~with $L_p=L$ across recursive steps, \textit{e.g.}, $\mathtt{max}\left(\{\Delta\mathrb{R}^{(l)}_{L_p=1}, \Delta\mathrb{R}^{(l)}_{L_p=L}, l\in[1, L]\}\right)$. 

From Fig.~\ref{fig:app-disp}, we can see the consistency in the shape of the displacement curve between amazon-ratings and AmazonPhoto. The displacement curves of \nsqure~with $L_p=1$ are smooth, whereas \nsqure~with  $L_p=L$ demonstrates fluctuation. We attribute this difference to the distinct inertia characteristics exhibited by the embedded nodes. 
For the embedded nodes in \nsqure~with $L_p=1$, they generally possess greater inertia, and thus tend to maintain current dynamics through recursive steps. In contrast, the embedded nodes in \nsqure~with $L_p=L$ have smaller inertia and vary their dynamics. Therefore, \nsqure~with $L_p=L$ can adjust the displacements of the embedded nodes more flexibly according to different situations and gain better performance in situations requiring precise distribution optimization. A step-dependent parameter may further improve the performance of \nsqure~with $L_p=1$ on a larger number of recursive steps. We leave this for future exploration.

\subsection{Ablation on Proximity Measurement}\label{ssec:nq}
\begin{figure*}
    \centering
    \includegraphics[width=0.6\textwidth]{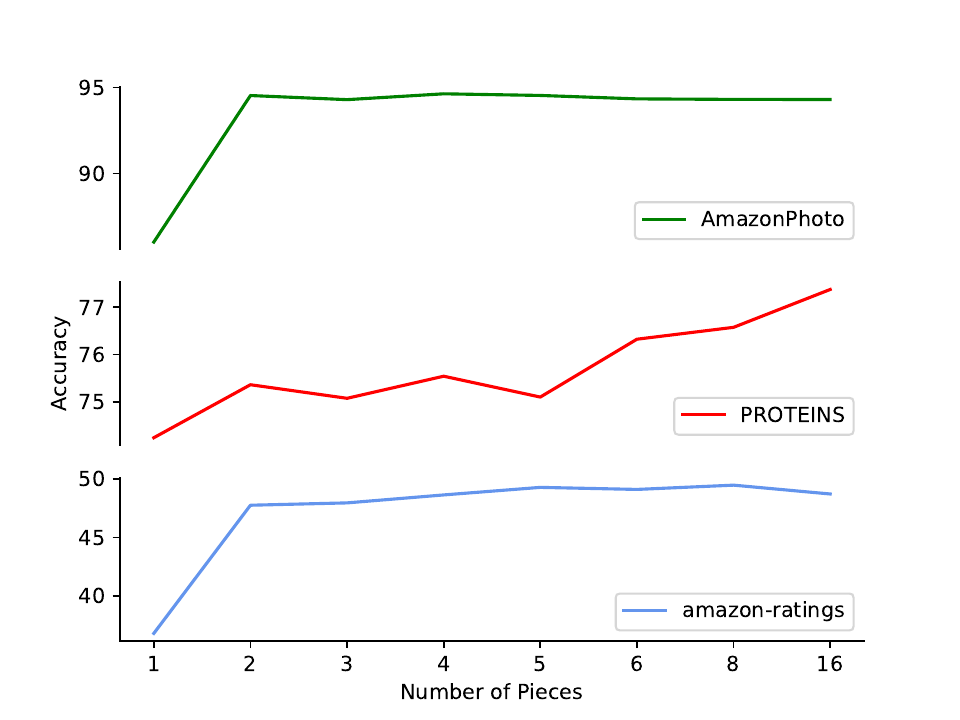}
    \figcaption{\textbf{Ablation studies on the number of pieces $k$.}}
    \label{fig:app-nq}
\end{figure*}

To approximate non-linear relations with low complexity, we employ piece-wise weighted inner products on the proximity measurement in Eq.~\ref{eq:w-innerproduct}. Each embedded node is divided into $k$ pieces. Ablation studies are conducted on $k$ with amazon-ratings, AmazonPhoto, and PROTEINS. As depicted in Fig.~\ref{fig:app-nq}, \nsqure~with multiple pieces outperforms a single piece on all three benchmarks and gains improvement with $k$ increasing. The optimal number of pieces $k$ is around $8$.

\subsection{Ablation on Messages}
In \nsqure, graph nodes and pseudo nodes perform message passing based on their current states in the common state space and pass on the learned messages to each other. States and messages take different roles in our proposed method. Specifically, states are the descriptive embeddings of graph nodes corresponding to specific tasks, and messages are employed for information exchange. For the message passing from graph nodes to pseudo nodes, the messages contain the features of an input graph. From pseudo nodes to graph nodes, the messages may contain the query information towards input graphs.
To evaluate the necessity of distinguishing states and messages, we conduct ablation studies on messages by directly passing graph node states in message passing. The results with or without specially distinguished messages are presented in Tab.~\ref{tab:app-message}. We can see that applying both states and messages achieves the best performance.

\begin{table}[t]
\caption{\textbf{Ablation studies on messages.}}
\label{tab:app-message}
\begin{center}
\begin{small}
\begin{sc}
\begin{tabular}{lccc} 
\toprule
                & w. messages     & w/o.~messages & w/o.-w.  \\ 
\midrule
questions       & \textbf{78.07} & 76.01        & -2.06  \\
amazon-ratings  & \textbf{50.25} & 47.79        & -2.46  \\
tolokers        & \textbf{86.25} & 85.05        & -1.20  \\
minesweeper     & \textbf{93.97} & 92.38        & -1.59  \\
CoauthorCS      & \textbf{94.44} & 92.39        & -2.05  \\
CoauthorPhysics & \textbf{97.56} & 96.27        & -1.29  \\
AmazonPhoto     & \textbf{95.75} & 93.80        & -1.95  \\
AmazonComputers & \textbf{92.51} & 90.73        & -1.78  \\
\bottomrule
\end{tabular}
\end{sc}
\end{small}
\end{center}
\end{table}
\begin{table}[t]
\caption{\textbf{Comparison between dense and pseudo-node-based message passing.}}
\label{tab:app-pseudo}
\begin{center}
\begin{small}
\begin{sc}
\begin{tabular}{lcc} 
\toprule
         & Dense & \nsqure  \\ 
\midrule
IMDB-B   & 77.94 & \textbf{79.95}          \\
IMDB-M   & 56.41 & \textbf{57.31}          \\
PROTEINS & 76.94 & \textbf{77.53}          \\
\bottomrule
\end{tabular}
\end{sc}
\end{small}
\end{center}
\end{table}

\subsection{Ablation on Pseudo Nodes}\label{ssec:app-pseudo}
In addition to investigating the number of pseudo nodes, we conduct ablation studies to examine the impact of including pseudo nodes within \nsqure. For cases without pseudo nodes, dense message passing is employed. However, as dense computation is not scalable, only small-scale benchmarks are utilized for this ablation study. The results are presented in Table~\ref{tab:app-pseudo}. Surprisingly, we find that \nsqure~with pseudo nodes outperforms dense computation on IMDB-B, IMDB-M, and PROTEINS. This can be attributed to pseudo nodes functioning as information filters in global message passing, facilitating the removal of redundant information while extracting discriminative features from inputs.

\section{Intuitions for the \nsqure~Implementation}
We provide more intuitions for the implementation in Section~\ref{sec:implementation}. \nsqure~embeds graph nodes and pseudo nodes into the common state space, employing a recurrent layer to parameterize the displacements of the embedded nodes. The recurrent layer includes pseudo-node adaptation and dynamic message passing. Pseudo-node adaptation employs $\mathtt{GlobMP}$ to generate query messages toward graph nodes. Dynamic message passing then extracts graph features through $\mathtt{LocalMP}$ and refines these extracted features at the pseudo-node level with $\mathtt{GlobMP}$.

As described by the pseudo-node adaptation in Eq.~\ref{eq:adaptation}, graph nodes first diffuse messages to pseudo nodes. Based on the information learned from these messages, pseudo nodes generate displacements to adjust their own representations, enabling better interactions with graph nodes. Subsequently, graph nodes collect responding messages from pseudo nodes, which serve as the query signals toward specific information of the input graphs.

During the dynamic message passing, Eq.~\ref{eq:nsqure-local-mp} presents the local message passing on input graphs. Graph nodes collect query messages from pseudo nodes, and process them together with their own generated messages and states through the $\mathtt{LocalMP}$ function. As a result, graph nodes learn the features of the input graphs and generate feedback messages to pseudo nodes accordingly. The generated messages containing the features of input graphs can also be leveraged to determine the displacements of graph nodes.

Eq.~\ref{eq:nsqure-global-mp} presents how graph nodes perform global message passing intermediately through pseudo nodes. Pseudo nodes receive the feedback messages and again generate their displacements. As feedback messages are aggregated at the pseudo-nodes level, more global information is incorporated, guiding the movements of graph nodes and their message update.

\section{Limitations}\label{sec:app-limitation}
\nsqure~employs shared parameters to update the distribution of pseudo nodes and graph nodes recursively. We only studied \nsqure~with a simple update mechanism. As discussed in the model analysis section, \nsqure~encounters performance degradation when the number of recursive steps increases. In addition, the learnable pseudo nodes can be regarded as parameters in GNNs, classified in line with neurons of GNNs. Under this interpretation of pseudo nodes, only a subset of neurons from GNNs are embedded in the common
state space within our framework. Comprehensively bridging all neurons and graph nodes remains an open research direction.

\section{Societal Impact}\label{sec:app-impact}
This paper proposed a novel method for constructing dynamic message-passing pathways by bridging graph nodes and pseudo nodes in a common state space. Our goal is to advance the field of graph representation learning. The proposed method remains independent of specific downstream applications. As graph data are ubiquitous in the real world, 
there are many potential applications of our work, including computational biology~\cite{zaidi_PretrainingDenoisingMolecular_2023}, intelligent transportation~\cite{rahmani_GraphNeuralNetworks_2023}, and algorithmic reasoning~\cite{diao_RelationalAttentionGeneralizing_2023a}. Regarding ethical considerations, we do not presently foresee evident issues or potential for adverse societal impacts.


\newpage
\section*{NeurIPS Paper Checklist}

\begin{enumerate}

\item {\bf Claims}
    \item[] Question: Do the main claims made in the abstract and introduction accurately reflect the paper's contributions and scope?
    \item[] Answer: \answerYes{} 
    \item[] Justification: The main claims made in the abstract and introduction accurately reflect the paper's contributions and scope.

\item {\bf Limitations}
    \item[] Question: Does the paper discuss the limitations of the work performed by the authors?
    \item[] Answer: \answerYes{} 
    \item[] Justification: This paper discusses the limitations of the work in Appendix~\ref{sec:app-limitation}.

\item {\bf Theory Assumptions and Proofs}
    \item[] Question: For each theoretical result, does the paper provide the full set of assumptions and a complete (and correct) proof?
    \item[] Answer: \answerNA{} 
    \item[] Justification: This paper does not provide theoretical results.

    \item {\bf Experimental Result Reproducibility}
    \item[] Question: Does the paper fully disclose all the information needed to reproduce the main experimental results of the paper to the extent that it affects the main claims and/or conclusions of the paper (regardless of whether the code and data are provided or not)?
    \item[] Answer: \answerYes{} 
    \item[] Justification: Detailed experimental setups are provided in Appendix~\ref{sec:app-exp-setup}. Code is provided at \url{https://github.com/sunjss/N2}.

\item {\bf Open access to data and code}
    \item[] Question: Does the paper provide open access to the data and code, with sufficient instructions to faithfully reproduce the main experimental results, as described in supplemental material?
    \item[] Answer: \answerYes{} 
    \item[] Justification: Code with instructions for reproduction is provided at \url{https://github.com/sunjss/N2}.

\item {\bf Experimental Setting/Details}
    \item[] Question: Does the paper specify all the training and test details (e.g., data splits, hyperparameters, how they were chosen, type of optimizer, etc.) necessary to understand the results?
    \item[] Answer: \answerYes{} 
    \item[] Justification: Detailed experimental setups are provided in Appendix~\ref{sec:app-exp-setup}.

\item {\bf Experiment Statistical Significance}
    \item[] Question: Does the paper report error bars suitably and correctly defined or other appropriate information about the statistical significance of the experiments?
    \item[] Answer: \answerYes{} 
    \item[] Justification: Please refer to Tab.~\ref{tab:graph-small}-\ref{tab:node-large}.

\item {\bf Experiments Compute Resources}
    \item[] Question: For each experiment, does the paper provide sufficient information on the computer resources (type of compute workers, memory, time of execution) needed to reproduce the experiments?
    \item[] Answer: \answerYes{} 
    \item[] Justification: Please refer to the beginning of Sec.~\ref{sec:exp} and Fig.~\ref{fig:time}.
    
\item {\bf Code Of Ethics}
    \item[] Question: Does the research conducted in the paper conform, in every respect, with the NeurIPS Code of Ethics \url{https://neurips.cc/public/EthicsGuidelines}?
    \item[] Answer: \answerYes{} 
    \item[] Justification: The research conducted in the paper conforms with the NeurIPS Code of Ethics in every respect.

\item {\bf Broader Impacts}
    \item[] Question: Does the paper discuss both potential positive societal impacts and negative societal impacts of the work performed?
    \item[] Answer: \answerYes{} 
    \item[] Justification: Please refer to Appendix~\ref{sec:app-impact}.
    
\item {\bf Safeguards}
    \item[] Question: Does the paper describe safeguards that have been put in place for responsible release of data or models that have a high risk for misuse (e.g., pretrained language models, image generators, or scraped datasets)?
    \item[] Answer: \answerNA{} 
    \item[] Justification: This paper poses no such risks.

\item {\bf Licenses for existing assets}
    \item[] Question: Are the creators or original owners of assets (e.g., code, data, models), used in the paper, properly credited and are the license and terms of use explicitly mentioned and properly respected?
    \item[] Answer: \answerYes{} 
    \item[] Justification: Please refer to Appendix~\ref{sec:app-exp-setup}.

\item {\bf New Assets}
    \item[] Question: Are new assets introduced in the paper well documented and is the documentation provided alongside the assets?
    \item[] Answer: \answerYes{} 
    \item[] Justification: Please refer to the supplementary materials and our open source code at \url{https://github.com/sunjss/N2}.

\item {\bf Crowdsourcing and Research with Human Subjects}
    \item[] Question: For crowdsourcing experiments and research with human subjects, does the paper include the full text of instructions given to participants and screenshots, if applicable, as well as details about compensation (if any)? 
    \item[] Answer: \answerNA{} 
    \item[] Justification: This paper does not involve crowdsourcing or research with human subjects.

\item {\bf Institutional Review Board (IRB) Approvals or Equivalent for Research with Human Subjects}
    \item[] Question: Does the paper describe potential risks incurred by study participants, whether such risks were disclosed to the subjects, and whether Institutional Review Board (IRB) approvals (or an equivalent approval/review based on the requirements of your country or institution) were obtained?
    \item[] Answer: \answerNA{} 
    \item[] Justification: This paper does not involve crowdsourcing or research with human subjects.

\end{enumerate}

\end{document}